\documentclass[10pt,twocolumn,letterpaper]{article}

\usepackage{iccv}
\usepackage{times}
\usepackage{epsfig}
\usepackage{graphicx}
\usepackage{amsmath}
\usepackage{amssymb}
\usepackage{bm}
\usepackage{algorithm}
\usepackage{algorithmicx}
\usepackage[noend]{algpseudocode}
\usepackage{cuted}
\usepackage{capt-of}

\usepackage{subcaption}
\usepackage{float}
\usepackage{multirow}
\usepackage{cuted}

\usepackage{overpic}
\usepackage{stmaryrd}
\usepackage{makecell}
\usepackage{wrapfig}

\makeatletter
\@namedef{ver@everyshi.sty}{}
\makeatother
\usepackage{tikz}

\usepackage{enumitem}
\usepackage{multirow}

\usepackage{xspace}
\usepackage{ifthen}

\usepackage{caption}
\usepackage{subcaption}
\usepackage{cuted}
\usepackage{capt-of}
\usepackage[marginal]{footmisc}
\usepackage[accsupp]{axessibility}

\usepackage{siunitx}
\usepackage{xargs}
\usepackage{microtype}

\usepackage{booktabs} 
\usepackage{tabularx}
\newcolumntype{Y}{>{\centering\arraybackslash}X}

\usepackage[pagebackref=true,breaklinks=true,colorlinks,bookmarks=false]{hyperref}

\iccvfinalcopy 

\definecolor{sig2023_color}{rgb}{0.835,.659,0.298}
\newcommand{\revised}[1]{\textcolor{black}{#1}}

\ificcvfinal\fi

\newcommand{\bA}{\mathbf{A}}
\newcommand{\bb}{\mathbf{b}}
\newcommand{\bc}{\mathbf{c}}
\newcommand{\bd}{\mathbf{d}}

\newcommand{\bff}{\mathbf{f}} 

\newcommand{\bs}{\mathbf{s}}

\newcommand{\bu}{\mathbf{u}}
\newcommand{\bv}{\mathbf{v}}

\newcommand{\bx}{\mathbf{x}}\newcommand{\bX}{\mathbf{X}}

\newcommand{\bgamma}{\boldsymbol{\gamma}}

\newcommand{\nE}{\mathbb{E}}

\newcommand{\nR}{\mathbb{R}}

\newcommand{\cP}{\mathcal{P}}

\newcommand{\figref}[1]{Fig.~\ref{#1}}
\newcommand{\secref}[1]{Section~\ref{#1}}
\newcommand{\eqnref}[1]{Eq.~\eqref{#1}}
\newcommand{\tabref}[1]{Table~\ref{#1}}

\DeclareMathOperator*{\argmin}{argmin~}

\makeatletter
\DeclareRobustCommand\onedot{\futurelet\@let@token\@onedot}
\def\@onedot{\ifx\@let@token.\else.\null\fi\xspace}
\def\eg{e.g\onedot} 
\def\ie{i.e\onedot} 
 
\def\etc{etc\onedot}

\def\etal{et~al\onedot}

\makeatother

\newcommand{\boldparagraph}[1]{\vspace{0.2cm}\noindent{\bf #1:} }

\begin{document}

\newcommand{\tikzcircle}[2][red,fill=red]{\tikz[baseline=-0.7ex]\draw[#1,radius=#2] (0,0) circle ;}%
\definecolor{gold}{RGB}{221, 196, 65}
\definecolor{silver}{RGB}{215, 215, 215}
\definecolor{bronze}{RGB}{126, 66, 5}

\newcommand{\gold}{\tikzcircle[gold,fill=gold]{2pt}}
\newcommand{\silve}{\tikzcircle[silver,fill=silver]{2pt}}
\newcommand{\bronze}{\tikzcircle[bronze,fill=bronze]{2pt}}

\def\emptySmall{\ \, - \ \,}
\def\emptyMed{\ \:\, - \ \:\,}
\def\emptyLarge{\ \ \, - \ \ \, }

\newcommand{\nothing}{\relax}

\title{Factor Fields: A Unified Framework for Neural Fields and Beyond}

\author{Anpei Chen$^{1,4}$
\and
Zexiang Xu$^2$
\and
Xinyue Wei$^3$ 
\and
Siyu Tang$^1$
\and
Hao Su$^{3}$ \qquad \qquad  Andreas Geiger$^{4,5}$\\
$^{1}$ ETH Zürich \qquad $^{2}$ Adobe Research \qquad $^{3}$ University of California, San Diego \\ $^{4}$ University of Tübingen \qquad $^{5}$ Tübingen AI Center 
}

\maketitle
\ificcvfinal\thispagestyle{empty}\fi

\begin{strip}\centering
  \begin{overpic}[width=\linewidth]{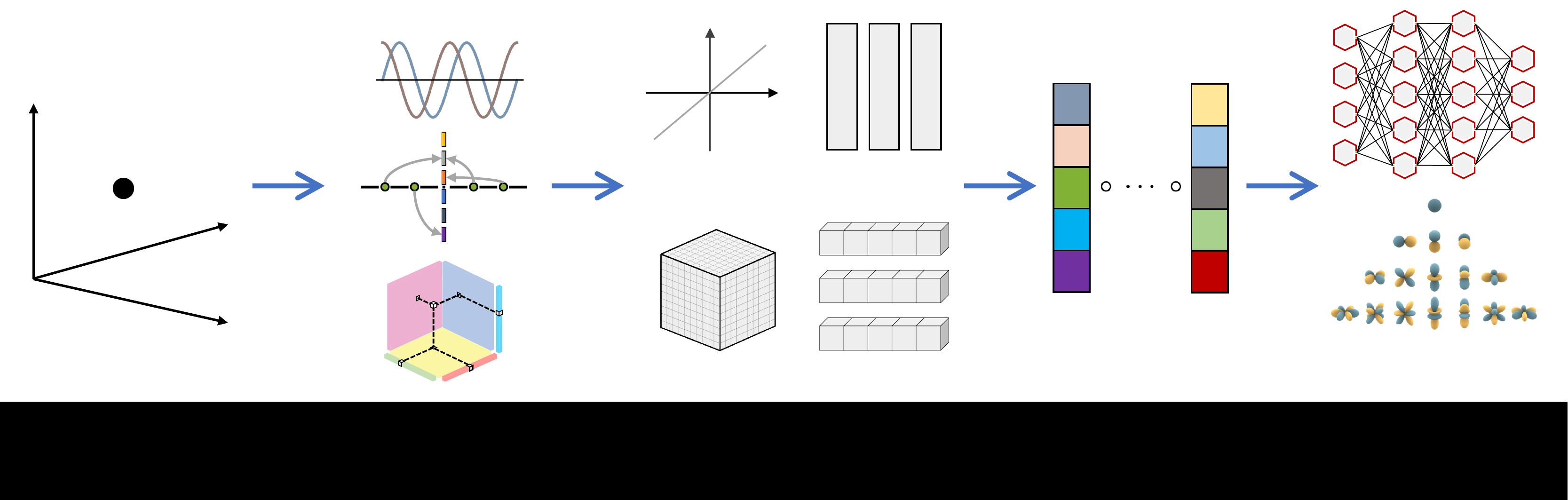}

    \put(5,23) { \small Point $\bx$ }
    
    \put(26.3,6.5) { $\cdots$ }
    \put(49,7.7) { $\cdots$ }
    \put(89.8,7.7) { $\cdots$ }

    \put(2,2.3) { \textcolor{white}{\large Input Domain }}
    
    \put(17.1,3.5) { \textcolor{white}{ Coordinate Transformation }}
    \put(26,1.1) { \textcolor{white}{$\bgamma(\bx)$}}
    
    \put(43,3.5) { \textcolor{white}{ Field Representation }}
    \put(48.3,1.1) { \textcolor{white}{$\bff(\bx)$}}
    
    \put(66,3.5) { \textcolor{white}{ Field Connector }}
    \put(71.5,1.1) { \textcolor{white}{\large $\circ$}}
    
    \put(86.9,3.5) { \textcolor{white}{ Projection }}
    \put(89.0,1.1) { \textcolor{white}{ $\cP(\bx)$}}
    
    \put(41.5,20.5) { \footnotesize Polynomial }
    \put(54,20.5) { \footnotesize MLPs }
    \put(43.6,7.5) { \footnotesize Grids }
    \put(53.3,7.5) { \footnotesize Vectors }
    
    \put(63.6,11) { \small $\bff_1(\bgamma_1(\bx))$ }
    \put(72.9,11) { \small $\bff_N(\bgamma_N(\bx))$ }
  \end{overpic}
\vspace{-5mm}
\captionof{figure}{\textbf{Factor Fields} is a framework  which unifies many recently proposed neural field representations (e.g., NeRF, EG3D, Instant-NGP, TensoRF) and allows for the creation of powerful new ones such as the proposed \textbf{Dictionary Factorization}. In particular, Factor Fields decomposes a signal into $N$ factors $\bff_1$ to $\bff_N$ (top-center), each of which is represented by one out of many different field representations operating on coordinate transformations $\bgamma_1$ to $\bgamma_N$. The resulting product field is passed to a projection function (\eg, MLP) which maps it to the signal's output.}
  \label{fig:teaser}
\end{strip}


\begin{abstract}
We present Factor Fields, a novel framework for modeling and representing signals. Factor Fields decomposes a signal into a product of factors, each represented by a classical or neural field representation which operates on transformed input coordinates. This decomposition results in a unified framework that accommodates several recent signal representations including NeRF, Plenoxels, EG3D, Instant-NGP, and TensoRF. Additionally, our framework allows for the creation of powerful new signal representations, such as the "Dictionary Field" (DiF) which is a second contribution of this paper. Our experiments show that DiF leads to improvements in approximation quality, compactness, and training time when compared to previous fast reconstruction methods. Experimentally, our representation achieves better image approximation quality on 2D image regression tasks, higher geometric quality when reconstructing 3D signed distance fields, and higher compactness for radiance field reconstruction tasks.
Furthermore, DiF enables generalization to unseen images/3D scenes by sharing bases across signals during training which greatly benefits use cases such as image regression from sparse observations and few-shot radiance field reconstruction. Our code is available at \href{https://apchenstu.github.io/FactorFields/}{\textcolor{sig2023_color}{https://apchenstu.github.io/FactorFields/}}.

\end{abstract}

\section{Introduction}

Effectively representing multi-dimensional digital content -- like 2D images or 3D geometry and appearance -- is critical for computer graphics and vision applications.
These digital signals are traditionally represented discretely as pixels, voxels, textures, or polygons.
Recently, significant headway has been made in developing advanced neural representations \cite{Mildenhall2020ECCV,Sun2022CVPR,Mueller2022TOG,Chen2022ECCV,Sitzmann2020NIPS}, which demonstrated superiority in modeling accuracy and efficiency over traditional representations for different image synthesis and scene reconstruction applications. 

In order to gain a better understanding of existing representations, make comparisons across their design principles, and create powerful new representations, we propose \emph{Factor Fields}, a novel mathematical framework that unifies many previous neural representations for multi-dimensional signals. This framework offers a simple formulation for modeling and representing signals.

Our framework decomposes a signal by factorizing it into multiple factor fields ($\bff_1,\dots,\bff_N$) operating on suitably chosen coordinate transformations ($\bgamma_1,\dots,\bgamma_N$) as illustrated in \figref{fig:teaser}. 
More specifically, each factor field decodes multi-channel features at any spatial location of a coordinate-transformed signal domain. 
The target signal is then regressed from the factor product via a learned projection function (\eg, MLP).

Our framework accommodates most previous neural representations. Many of them can be represented in our framework as a single factor with a domain transformation -- for example, the MLP network as a factor with a positional encoding transformation in NeRF \cite{Mildenhall2020ECCV}, the tabular encoding as a factor with a hash transformation in Instant-NGP \cite{Mueller2022TOG}, and the feature grid as a factor with identity transformation in DVGO \cite{Sun2022CVPR} and Plenoxels \cite{Keil2022CVPR}.
Recently, TensoRF \cite{Chen2022ECCV} introduced a tensor factorization-based representation, which can be seen as a representation of two Vector-Matrix or three CANDECOMP-PARAFAC decomposition factors with axis-aligned orthogonal 2D and 1D projections as transformations.
The potential of a multi-factor representation has been demonstrated by TensoRF, which has led to superior quality and efficiency on radiance field reconstruction and rendering, while being limited to orthogonal transformations. 

This motivates us to generalize previous classic neural representations via a single unified framework which enables easy and flexible combinations of previous neural fields and transformation functions, yielding novel representation designs.
As an example, we present \textit{Dictionary Field (DiF)}, a two-factor representation that is composed of (1) a basis function factor with periodic transformation to model the commonalities of patterns that are shared across the entire signal domain and (2) a coefficient field factor with identity transformation to express localized spatially-varying features in the signal.
The combination of both factors allows for an efficient representation of the global and local properties of the signal.
Note that, most previous single-factor representations can be seen as using only one of such functions -- either a basis function, like NeRF and Instant-NGP, or a coefficient function, like DVGO and Plenoxels.
In DiF, jointly modeling two factors (basis and coefficients) leads to superior quality over previous methods like Instant-NGP and enables compact and fast reconstruction, as we demonstrate on various downstream tasks.

As DiF is a member of our general Factor Fields family, we conduct a rich set of ablation experiments over the choice of basis/coefficient functions and basis transformations. 
We evaluate DiF against various variants and baselines on three classical signal representation tasks: 2D image regression, 3D SDF geometry reconstruction, and radiance field reconstruction for novel view synthesis.
We demonstrate that our factorized DiF representation is able to achieve state-of-the-art reconstruction results that are better or on par with previous methods, while achieving superior modeling efficiency.
For instance, compared to Instant-NGP our method leads to better reconstruction and rendering quality, while effectively \emph{halving} the total number of model parameters (capacity) for SDF and radiance field reconstruction, demonstrating its superior accuracy and efficiency.

Moreover, in contrast to recent neural representations that are designed for purely per-scene optimization, our factorized representation framework is able to learn basis functions across different scenes.
As shown in preliminary experiments, this enables learning across-scene bases from multiple 2D images or 3D radiance fields, leading to signal representations that generalize and hence improve reconstruction results from sparse observations such as in the few-shot radiance reconstruction setting. In summary,
\begin{itemize}
    \item We introduce a common framework \emph{Factor Fields} that encompasses many recent radiance field / signal representations and enables new models from the Factor Fields family.
    \item We propose \textit{DiF}, a new member of the Factor Fields family representation that factorizes a signal into coefficient and basis factors which allows for exploiting similar signatures spatially and across scales.
    \item Our model can be trained jointly on multiple signals, recovering general basis functions that allow for reconstructing parts of a signal from sparse or weak observations.  
    \item We present thorough experiments and ablation studies that demonstrate improved performance (accuracy, runtime, memory), and shed light on the performance improvements of DiF vs. other models in the Factor Fields family.
\end{itemize}

\section{Factor Fields}
\label{sec:factor_fields}

\begin{figure}
	\hspace{-0.8cm}
	\begin{subfigure}{0.25\linewidth}
		\centering
		\includegraphics[height=5.5cm]{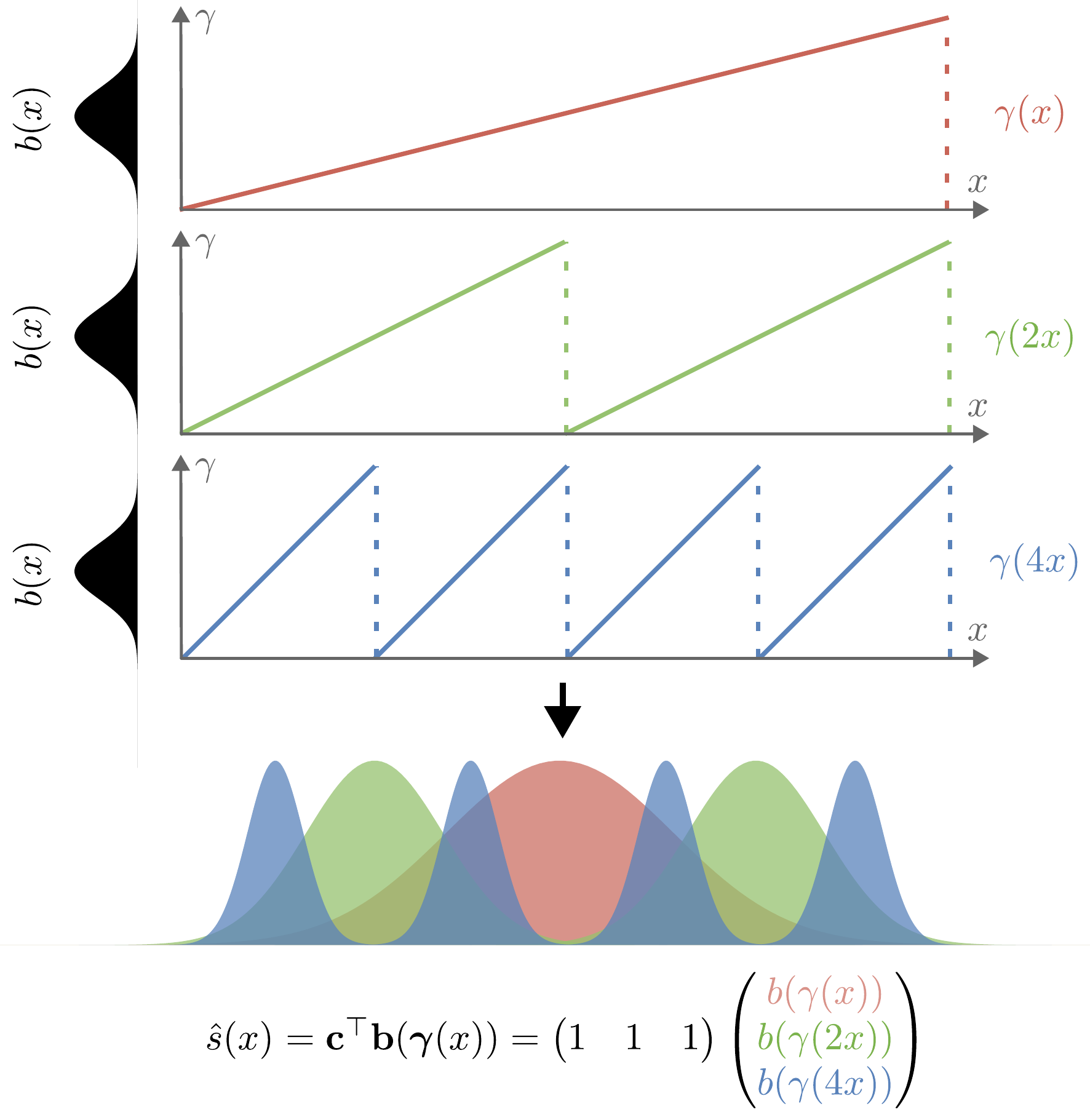}
            \label{fig:local_basis_1d}
	\end{subfigure}%
	\hspace{3.5cm}
	\begin{subfigure}{0.25\linewidth}
		\centering
		\includegraphics[height=5.5cm]{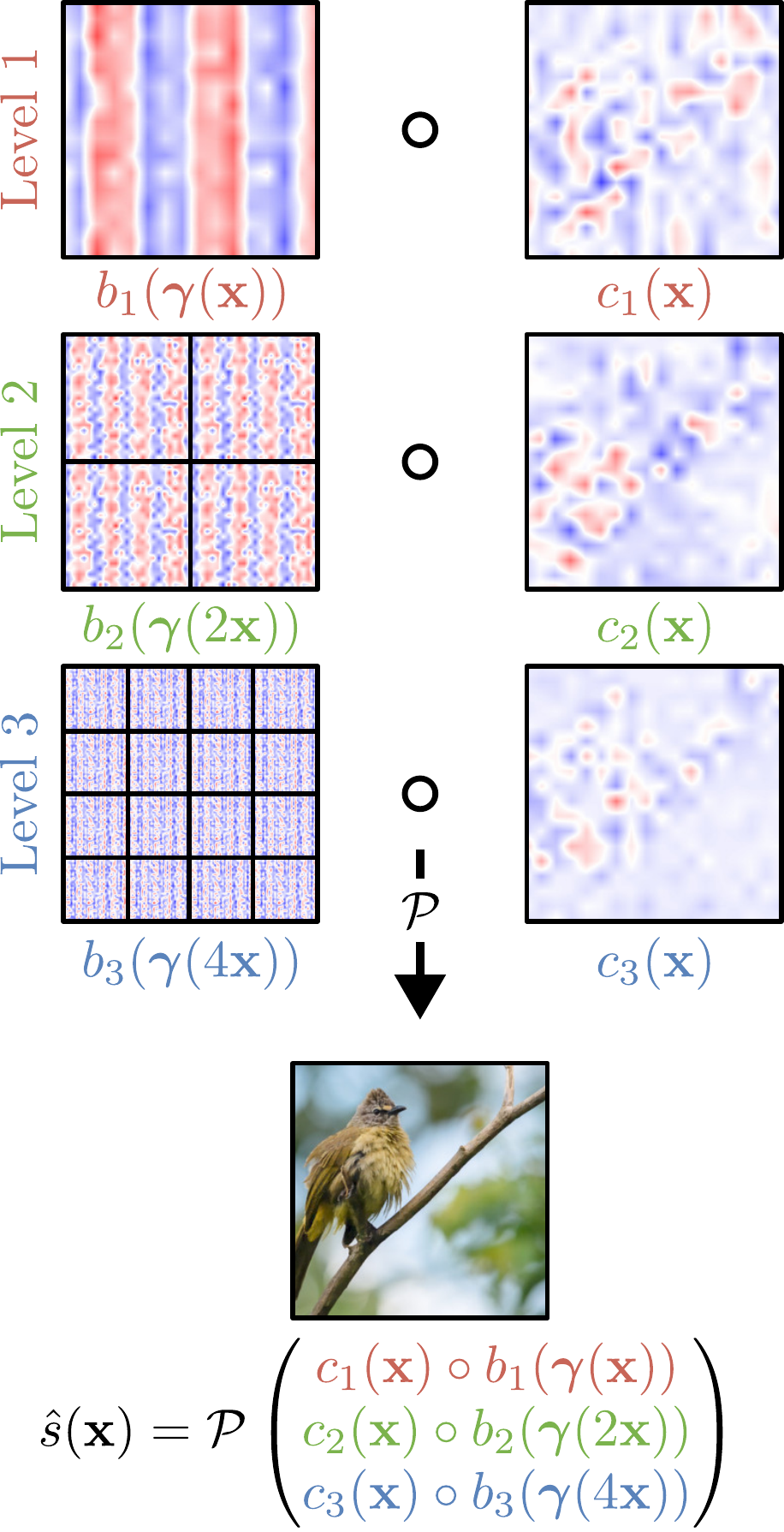}
            \label{fig:local_basis_2d}
	\end{subfigure}%
        \put(-177.0,-3) { \small (a) 1D Illustration }
        \put(-50.0,-3) { \small (b) 2D Example }
	\caption{\textbf{Local Basis.} (a) Choosing a (periodic) coordinate transformation $\bgamma(x)$ allows for applying the same basis function $b(x)$ at multiple spatial locations and scales. For clarity, we have chosen constant coefficients $\bc=\mathbf{1}$. (b) Composing multiple bases at different spatial resolutions with their respective coefficients yields a powerful representation for signal $s(\bx)$. In practice, we use multiple bases and coefficient fields at each resolution.}
	\label{fig:local_basis}
\end{figure}

We seek to compactly represent a continuous $Q$-dimensional signal $\bs:\nR^D \rightarrow \nR^Q$ on a $D$-dimensional domain. We assume that signals are not random, but structured and hence share similar signatures within the same signal (spatially and across different scales) as well as between different signals.
In the following, we develop our factor fields model step-by-step, starting from a standard basis expansion.

\boldparagraph{Dictionary Field (DiF)}
%
Let us first consider a 1D signal $s(\bx):\nR^D \rightarrow \nR$. Using basis expansion, we decompose $s(\bx)$ into a set of coefficients $\bc=(c_1, \dots, c_K)^\top$ with $c_k\in \nR$ and basis functions $\bb(\bx)=(b_1(\bx), \dots, b_K(\bx))^\top$ with $b_k:\nR^D \rightarrow \nR$:
\begin{equation}
\hat{s}(\bx) = \bc^\top \bb(\bx)
\label{eq:expansion_1}
\end{equation}
Note that we denote $s(\bx)$ as the true signal and $\hat{s}(\bx)$ as its approximation.

%
Representing the signal $s(\bx)$ using a global set of basis functions is inefficient as information cannot be shared spatially. We hence generalize the above formulation by (i) exploiting a spatially varying coefficient field $\bc(\bx)=(c_1(\bx), \dots, c_K(\bx))^\top$ with $c_k:\nR^D \rightarrow \nR$ and (ii) transforming the coordinates of the basis functions via a coordinate transformation function $\bgamma:\nR^D\rightarrow \nR^B$:
\begin{equation}
\hat{s}(\bx) = \bc(\bx)^\top \bb\left(\bgamma(\bx)\right)
\label{eq:expansion_2}
\end{equation}
When choosing $\bgamma$ to be a periodic function, this formulation allows us to apply the same basis at multiple spatial locations and optionally also at multiple different scales while varying the coefficients $\bc$, as illustrated in \figref{fig:local_basis}.
Note that in general $B$ does not need to match $D$, and hence the domain of the basis functions also changes accordingly: $b_k:\nR^B \rightarrow \nR$.
Further, we obtain \eqnref{eq:expansion_1} as a special case when setting $\bc(\bx)=\bc$ and $\bgamma(\bx) = \bx$.

%
So far, we have considered a 1D signal $s(\bx)$. However, many signals have more than a single dimension (\eg, 3 in the case of RGB images or 4 in the case of radiance fields). We generalize our model to $Q$-dimensional signals $\bff(\bx)$ by introducing a projection function $\cP:\nR^K\rightarrow \nR^Q$ and replacing the inner product with the element-wise/Hadamard product (denoted by $\circ$ in the following):
\begin{equation}
\hat{\bs}(\bx) = \cP\left(\bc(\bx) \circ \bb\left(\bgamma(\bx)\right)\right)
\label{eq:expansion_3}
\end{equation}
We refer to \eqnref{eq:expansion_3} as \textbf{Dictionary Field (DiF)}.
Note that in contrast to the scalar product $\bc^\top \bb$ in \eqnref{eq:expansion_2}, the output of $\bc \circ \bb$ is a $K$-dimensional vector which comprises the individual coefficient-basis products as input to the projection function $\cP$ which itself can be either linear or non-linear. In the linear case, we have $\cP(\bx) = \bA\bx$ with $\bA\in\nR^{Q\times K}$. Moreover, note that for $Q=1$ and $\bA = (1,\dots,1)$ we recover \eqnref{eq:expansion_2} as a special case. The projection operator $\cP$ can also be utilized to model the volumetric rendering operation when reconstructing a 3D radiance field from 2D image observations as discussed in \secref{sec:projection}.

\boldparagraph{Factor Fields}
To allow for more than 2 factors, we generalize \eqnref{eq:expansion_3} to our full Factor Fields framework  by replacing the coefficients $\bc(\bx)$ and basis $\bb(\bx)$ with a set of factor fields $\{\bff_i(\bx)\}_{i=1}^N$:
\begin{equation}
\hat{\bs}(\bx) = \cP\left(\prod_{i=1}^N \bff_i\left(\bgamma_i(\bx)\right)\right)
\label{eq:expansion_4}
\end{equation}
Here, $\prod$ denotes the element-wise product of a sequence of factors.
Note that in this general form, each factor $\bff_i:\nR^{F_i}\rightarrow\nR^K$ may be equipped with its own coordinate transformation $\bgamma_i:\nR^D\rightarrow\nR^{F_i}$.

We obtain \textbf{DiF} in \eqnref{eq:expansion_3} as a special case of our \textbf{Factor Fields} framework in \eqnref{eq:expansion_4} by setting $\bff_1(\bx) = \bc(\bx)$, $\bgamma_1(\bx) = \bx$, $\bff_2(\bx)=\bb(\bx)$ and $\bgamma_2(\bx) = \bgamma(\bx)$ with $N=2$.
Besides DiF, the Factor Fields framework generalizes many recently proposed radiance field representations in one unified model as we will discuss in \secref{sec:common_framework}.  


%
In our formulation, $\{\bgamma_i\}$ are considered deterministic functions while $\cP$ and $\{\bff_i\}$ are parametric mappings (\eg, polynomials, multi-layer perceptrons or 3D feature grids) whose parameters (collectively named $\theta$ below) are optimized.
The parameters $\theta$ can be optimized either for a single signal or jointly for multiple signals.
When optimizing for multiple signals jointly, we share the parameters of the projection function and basis factors (but not the parameters of the coefficient factors) across signals.

\subsection{Factor Fields $\bff_i$}

For modeling the factor fields $\bff_i:\nR^{F_i} \rightarrow \nR^K$, we consider various different representations in our Factor Fields framework as illustrated in \figref{fig:teaser} (bottom-left). In particular, we consider polynomials, MLPs, 2D and 3D feature grids and 1D feature vectors.

MLPs have been proposed as signal representations in Occupancy Networks \cite{Mescheder2019CVPR}, DeepSDF \cite{Park2019CVPR} and NeRF \cite{Mildenhall2020ECCV}. While MLPs excel in compactness and induce a useful smoothness bias, they are slow to evaluate and hence increase training and inference time. To address this, DVGO \cite{Sun2022CVPR} proposes a 3D voxel grid representation for radiance fields. While voxel grids are fast to optimize, they increase memory significantly and do not easily scale to higher dimensions. To better capture the sparsity in the signal, Instant-NGP \cite{Mueller2022TOG} proposes a hash function in combination with 1D feature vectors instead of a dense voxel grid, and TensoRF \cite{Chen2022ECCV} decomposes the signal into matrix and vector products.
Our \textit{Factor Fields} framework allows any of the above representations to model any factor $\bff_i$. As we will see in \secref{sec:common_framework}, many existing models are hence special cases of our framework.

\subsection{Coordinate Transformations $\bgamma_i$}

The coordinates input to each factor field $\bff_i$ are transformed by a coordinate transformation function $\bgamma_i:\nR^D\rightarrow \nR^{F_i}$. 

\boldparagraph{Coefficients}
When the factor field $\bff_i$ represents coefficients, we use the identity $\bgamma_i(\bx) = \bx$ for the corresponding coordinate transformation since coefficients vary freely over the signal domain.

\begin{figure}
  \centering
  \small\sffamily
  \input{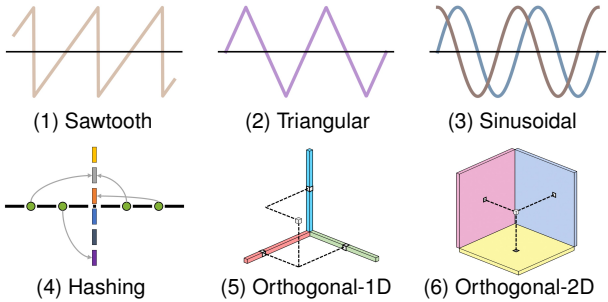}
  \vspace{-1mm}
  \caption{\label{fig:trans_func}%
    \textbf{Coordinate Transformations.} We show various periodic (top) and non-periodic (bottom) coordinate transformations $\bgamma$ used in our framework.
  }
  \vspace{-2mm}
\end{figure}

\boldparagraph{Local Basis}
The coordinate transformation $\bgamma_i$ enables the application of the same basis function $\bff_i$ at \textit{multiple locations} as illustrated in \figref{fig:local_basis}. In this paper, we consider sawtooth, triangular, sinusoidal (as in NeRF \cite{Mildenhall2020ECCV}), hashing (as in Instant-NGP \cite{Mueller2022TOG}) and orthogonal (as in TensoRF \cite{Chen2022ECCV}) transformations in our framework, see \figref{fig:trans_func}.

\boldparagraph{Multi-scale Basis}
The coordinate transformation $\bgamma_i$ also allows for applying the same basis $\bff_i$ at \textit{multiple spatial resolutions} of the signal by transforming the coordinates $\bx$ with (periodic) transformations of different frequencies as illustrated in \figref{fig:local_basis}. This is crucial as signals typically carry both high and low frequencies, and we seek to exploit our basis representation across the full spectrum to model fine details of the signal as well as smooth signal components.

Specifically, we model the target signal with a set of multi-scale (field of view) basis functions. We arrange the basis into $L$ levels where each level covers a different scale. Let $[\bu,\bv]$ denote the bounding box of the signal along one dimension. The corresponding scale is given by $(\bv-\bu)/f_l$ where $f_l$ is the frequency at level $l$. A large scale basis (\eg, level $1$) has a low frequency and covers a large region of the target signal while a small scale basis (\eg, level $L$) has a large frequency $f_L$ covering a small region of the target signal.

We implement our multi-scale representation (PR) by multiplying the scene coordinate $\bx$ with the level frequency $f_l$ before feeding it to the coordinate transformation function $\bgamma_i$ and then concatenating the results across the different scale $l=1,\dots,L$:
\begin{equation}
\bgamma^{\text{PR}}_i(\bx) = \left(\bgamma_i(\bx\, f_1),\dots,\bgamma_i(\bx\, f_L)\right)
\end{equation}
Here, $\bgamma_i$ is any of the coordinate transformations in \figref{fig:trans_func}, and $\bgamma_{\text{PR}}$ is the final coordinate transform of our multi-scale representation.

As illustrated in \figref{fig:local_basis} (b), when considering one coefficient factor $\bff_1(\bx)=\bc(\bx)$ and one basis factor $\bff_2(\bx)=\bb(\bx)$ with coordinate transformation $\bgamma^{\text{PR}}_2(\bx)$ results in the target signal $\bs(\bx)$ being decomposed as the product of spatial varying coefficient maps and multi-level basis maps which comprise repeated local basis functions. 


\subsection{Projection $\cP$}
\label{sec:projection}

To represent multi-dimensional signals, we introduced a projection function $\cP:\nR^K\rightarrow \nR^Q$ that maps from the $K$-dimensional Hadamard product $\prod_i \bff_i$ to the $Q$-dimensional target signal. We distinguish two cases in our framework: The case where direct observations from the target signal are available (\eg, pixels of an RGB image) and the indirect case where observations are projections of the target signal (\eg, pixels rendered from a radiance field).

\boldparagraph{Direct Observations}
In the simplest case, the projection function realizes a learnable linear mapping $\cP(\bx) = \bA\bx$ with parameters $\bA\in\nR^{Q\times K}$ to map the $K$-dimensional Hadamard product $\prod_i \bff_i$ to the $Q$-dimensional signal. However, a more flexible model is attained if $\cP$ is represented by a shallow non-linear multi-layer perceptron (MLP) which is the default setting in all of our experiments.

\boldparagraph{Indirect Observations}
In some cases, we only have access to \textit{indirect} observations of the signal. For example, when optimizing neural radiance fields, we typically only observe 2D images instead of the 4D signal (density and radiance). In this case, extend $\cP$ to also include the differentiable volumetric rendering process. More concretely, we first apply a multi-layer perceptron to map the view direction $\bd\in\nR^3$ and the product features $\prod_i \bff_i$ at a particular location $\bx\in\nR^3$ to a color value $\bc\in\nR^3$ and a volume density $\sigma\in\nR$. Next, we follow Mildenhall \etal \cite{Mildenhall2020ECCV} and approximate the intractable volumetric projection integral using numerical integration.
More formally, let $\{(\bc_i,\sigma_i)\}_{i=1}^N$ denote the color and volume density values of $N$ random samples along a camera ray.
%
%
The RGB color value $\bc$ at the corresponding pixel is obtained using alpha composition
\begin{equation}
\bc_r = \sum_{i=1}^N \, T_i \, \alpha_i \, \bc_i \hspace{0.5cm}
T_i = \prod_{j=1}^{i-1}\left(1-\alpha_j\right)\hspace{0.5cm}
\alpha_i = 1-\exp\left(-\sigma_i\delta_i\right)
\label{eq:alphacomp}
\end{equation}
where $T_i$ and $\alpha_i$ denote the transmittance and alpha value of sample $i$ and $\delta_i=\left\Vert\bx_{i+1}-\bx_{i}\right\Vert_2$ is the distance between neighboring samples. The \textit{composition} of the learned MLP and the volume rendering function in \eqnref{eq:alphacomp} constitute the projection function $\cP$.


\subsection{Space Contraction}
\label{sec:space_contraction}

We normalize the input coordinates $\bx\in\nR^D$ to $[0,1]$ before passing them to the coordinate transformations $\bgamma_i(\bx)$ by applying a simple space contraction function to $\bx$. We distinguish two settings:

For \textbf{bounded signals} with $D$-dimensional bounding box $[\bu,\bv]$ (where $\bu,\bv\in\nR^D$), we utilize a simple linear mapping to normalize all coordinates to the range $[0,1]$:
\begin{equation}
\text{contract}(\bx) = \frac{\bx-\bu}{\bv-\bu}
\end{equation}
For \textbf{unbounded signals} (\eg, an outdoor radiance field), we adopt Mip-NeRF 360's \cite{Barron2022CVPR} space contraction function\footnote{In our implementation, we slightly modify \eqnref{eq:mipnerf360} to map coordinates to a unit ball centered at 0.5 which avoids negative coordinates when indexing feature grids.}:
\begin{equation}
\text{contract}(\bx) = 
\begin{cases}
\bx & {\Vert\bx\Vert}_2\leq 1 \\
\left(2-\frac{1}{{\Vert\bx\Vert}_2}\right)
\left(\frac{\bx}{{\Vert\bx\Vert}_2}\right) & {\Vert\bx\Vert}_2 > 1
\end{cases}
\label{eq:mipnerf360}
\end{equation}
%

\subsection{Optimization}

Given samples $\{(\bx,\bs(\bx))\}$ from the signal, we minimize
\begin{equation}
\argmin_\theta \nE_{\bx} \left[ {\Vert \bs(\bx) - \hat{\bs}_\theta(\bx) \Vert}_2 + \Psi(\theta) \right]
\end{equation}
where $\Psi(\theta)$ is a regularizer on the model parameters. We optimize this objective using stochastic gradient descent.


\boldparagraph{Sparsity Regularization}
While using the $\ell_0$ norm for sparse coefficients is desirable, this leads to a difficult optimization problem. Instead, we use a simpler strategy which we found to work surprisingly well. We regularize our objective by randomly dropping a subset of the $K$ features of our model by setting them to zero with probability $\mu$. This forces the signal to be represented with random combinations of features at every iteration, encouraging sparsity and preventing co-adaptation of features. We implement this dropout regularizer using a random binary vector $\bm$ which we multiply element-wise with the factor product: $\bm \circ \prod_i \bff_i$.

\boldparagraph{Initialization}
During all our experiments, we initialize the basis factors using the discrete cosine transform (DCT), while initializing the parameters of the coefficient factors and projection MLP randomly. We experimentally found this to improve the quality of the solution as illustrated in our ablation study in \tabref{tab:factor_number} to \tabref{tab:concatenate}.

\boldparagraph{Multiple Signals}
When optimizing for multiple signals jointly, we share the parameters of the projection function and basis factors (but not the parameters of the coefficient factors) across signals.
As evidenced by our experiments in \secref{sec:experiments_generalization}, sharing bases across different signals while encouraging sparse coefficients improves generalization and enables reconstruction from sparse observations.

\section{Factor Fields As A Common Framework}
\label{sec:common_framework}
Advanced neural representations have emerged as a promising 
replacement for traditional representations and been applied to improve the reconstruction quality and efficiency in various graphics and vision applications, 
such as novel view synthesis \cite{Zhou2018SIGGRAPH, Lombardi2019SIGGRAPH, Thies2019TOG, Aliev2019ARXIV, Mildenhall2020ECCV, Liu2020NEURIPS, Chen2022ECCV, xu2022CVPR,verbin2021ref}, generative models \cite{Schwarz2020NEURIPS, Chan2021CVPR, Chan2022CVPR, Gao2022AINIPS}, 3D surface reconstruction \cite{Niemeyer2019ARXIV, Yariv2021NEURIPS, Wang2021NEURIPSa, Yu2022NEURIPS, Kobayashi2022NIPS}, image processing \cite{Chen2021CVPRa}, graphics asset modeling \cite{Rainer2019CGF,Kuznetsov2021SIGGRAPH,Zhu2021SIGGRAPH}, inverse rendering \cite{Bi2020ECCV,Bi2020ARXIV,Zhang2021CVPR,Boss2021CVPR,Boss2021NEURIPS,zhang2022modeling}, dynamic scene modeling \cite{Pumarola2021CVPR,Li2020ARXIV,park2021nerfies,Li2021ARXIV}, and scene understanding \cite{Peng2022ARXIV, Wallingford2022ARXIV} amongst others.

Inspired by classical factorization and learning techniques, like sparse coding \cite{Wright2009TPAMI, Yang2010TIP, Heide2015CVPR} and principal component analysis (PCA) \cite{Rubinstein2008eCSD, Mairal2009ICPS}, we propose a novel neural factorization-based framework for neural representations. 
Our Factor Fields framework unifies many recently published neural representations and enables the instantiation of new models in the Factor Fields family, which, as we will see, exhibit desirable properties in terms of approximation quality, model compactness, optimization time and generalization capabilities.
In this section, we will discuss the relationship to prior work in more detail. A systematic performance comparison of the various Factor Field model instantiations is provided in \secref{sec:designs}.

\boldparagraph{Occupancy Networks, IMNet and DeepSDF} \cite{Mescheder2019CVPR, Chen2019CVPR,  Park2019CVPR} represent the surface implicitly as the continuous decision boundary of an MLP classifier or by regressing a signed distance value. The vanilla MLP representation provides a continuous implicit $3$D mapping, allowing for the extraction of $3$D meshes at any resolution. This setting corresponds to our Factor Fields model when using a single factor (i.e., $N=1$) with $\bgamma_1(\bx)=\bx$, $\bff_1(\bx)=\bx$, $\cP(\bx)=\text{MLP}(\bx)$, thus $\hat{\bs}(\bx) = \text{MLP}(\bx)$. While this representation is able to generate high-quality meshes, it fails to model high-frequency signals, such as images due to the implicit smoothness bias of MLPs.

\boldparagraph{NeRF} \cite{Mildenhall2020ECCV} proposes to represent a radiance field via an MLP in Fourier space by encoding the spatial coordinates with a set of sinusoidal functions. This corresponds to our Factor Fields setting when using a single factor with $\bgamma_1(\bx)=\left(\sin(\bx f_1), \cos(\bx f_1), \dots, \sin(\bx f_L), \cos(\bx f_L) \right)$, $\bff_1(\bx)=\bx$ and $\cP(\bx)=\text{MLP}(\bx)$. 
Here, the coordinate transformation $\bgamma_1(\bx)$ is a sinusoidal mapping as shown in \figref{fig:trans_func} (3), which enables high frequencies.


\boldparagraph{Plenoxels} \cite{Keil2022CVPR} use sparse voxel grids to represent $3$D scenes, allowing for direct optimization without neural networks, resulting in fast training. This corresponds to our Factor Fields framework when setting $N=1$, $\bgamma_1(\bx)=\bx$, $\bff_1(\bx)=\text{3D-Grid}(\bx)$, and $\cP(\bx)=\bx$ for the density field while $\cP(\bx)=\text{SH}(\bx)$ (Spherical Harmonics) for the radiance field. In related work, DVGO \cite{Sun2022CVPR} proposes a similar design, but replaces the sparse $3$D grids with dense grids and uses a tiny MLP as the projection function $\cP$.
While dense grid modeling is simple and leads to fast feature queries, it requires high spatial resolution (and hence large memory) to represent fine details. Moreover, optimization is more easily affected by local minima compared to MLP representations that benefit from their inductive smoothness bias.



\boldparagraph{ConvONet and EG3D} \cite{Peng2020ECCV, Chan2022CVPR} use a tri-plane representation to model $3$D scenes by applying an orthogonal coordinate transformation to spatial points within a bounded scene, and then representing each point as the concatenation of features queried from a set of 2D feature maps. This representation allows for aggregating $3$D features using only $2$D convolution, which significantly reduces memory footprint compared to standard $3$D grids. The setting can be viewed as an instance of our Factor Fields framework, with $N=1$, $\bgamma_1(\bx)=\text{Orthogonal-2D}(\bx)$, $\bff_1(\bx)=\text{2D-Maps}(\bx)$ and $\cP(\bx)=\text{MLP}(\bx)$. However, while the axis-aligned transformation allows for dimension reduction and feature sharing  along the axis, it can be challenging to handle complicated structures due to the axis-aligned bias of this representation.

\boldparagraph{Instant-NGP} \cite{Mueller2022TOG} exploits a multi-level hash grid to efficiently model internal features of target signals by hashing spatial locations to $1$D feature vectors. This approach corresponds to our Factor Fields framework when using $N=1$, $\bgamma_1(\bx)=\text{Hashing}(\bx)$, $\bff_1(\bx)=\text{Vectors}(\bx)$ and $\cP(\bx)=\text{MLP}(\bx)$ with $L=16$ scales.
However, the multi-level hash mappings can result in dense collisions at fine scales, and the one-to-many mapping forces the model to distribute its capacity bias towards densely observed regions and noise in areas with fewer observations. 
The concurrent work VQAD \cite{Takikawa2022SIGGRAPH} introduces a hierarchical vector-quantized auto-decoder (VQ-AD) representation that learns an index table as the coordinate transformation function which allows for higher compression rates.

\boldparagraph{TensoRF} \cite{Chen2022ECCV} factorizes the radiance fields into the products of vectors and matrices (TensoRF-VM) or multiple vectors (TensoRF-CP), achieving efficient feature queries at low memory footprint. This setting instantiates our Factor Fields framework for $N=2$, $\bgamma_1(\bx)=\text{Orthogonal-1D}(\bx)$, $\bff_1(\bx)=\text{Vectors}(\bx)$, $\bgamma_2(\bx)=\text{Orthogonal-2D}(\bx)$, $\bff_2(\bx)=\text{2D-Maps}(\bx)$ for VM decomposition and $N=3$, $\bgamma_i(\bx)=\text{Orthogonal-1D}(\bx)$, $\bff_i(\bx)=\text{Vectors}(\bx)$ for CP decomposition. Moreover, TensoRF uses both SH and MLP models for the projection $\cP$.
Similar to ConvONet and EG3D, TensoRF is sensitive to the orientation of the coordinate system due to the use of an orthogonal coordinate transformation function. 
Note that, with the exception of TensoRF \cite{Chen2022ECCV}, all of the above representations factorize the signal using a single factor field, that is $N=1$. As we will show in \tabref{tab:factor_number} to \tabref{tab:field_representation}, using multiple factor fields (\ie, $N>1$) provides stronger model capacity.

\boldparagraph{ArXiv Preprints}
The field of neural representation learning is advancing fast and many novel representations have been published as preprints on ArXiv recently. We now briefly discuss the most related ones and their relationship to our work:
\textit{Phasorial Embedding Fields (PREF)} \cite{Huang2022ARXIV} proposes to represent a target signal with a set of phasor volumes and then transforms them into the spatial domain with an inverse fast Fourier Transform (iFFT) for compact representation and efficient scene editing. This method shares a similar idea with DVGO and extends the projection function $\cP$ in our formulation with an iFFT function.
\textit{Tensor4D} \cite{Shao2022ARXIV} extends the triplane representation to 4D human reconstruction by using $3$ triplane Factor Fields and indexing plane features with $3$ orthogonal coordinate transformations (i.e., $\text{Orthogonal-2D}(\bx)=(xy,xt,yt)$, $\text{Orthogonal-2D}(\bx)=(xz,xt,zt)$, $\text{Orthogonal-2D}(\bx)=(yz,yt,zt)$). 
\textit{NeRFPlayer} \cite{Song2022ARXIV} represents dynamic scenes via deformation, newness and decomposition fields, using multiple factors similar to TensoRF. It further extends the features by a time dimension.
\textit{D-TensoRF} \cite{Jang2022ARXIV} reconstructs dynamic scenes using matrix-matrix factorization, similar to the VM factorization of TensoRF, but replacing $\bgamma_1(\bx)=\text{Orthogonal-1D}(\bx)$ and $\bff_1(\bx)=\text{Vectors}(\bx)$ with $\bgamma_1(\bx)=\text{Orthogonal-2D}(\bx)$ and $\bff_1(\bx)=\text{2D-Maps}(\bx)$. 
\textit{Quantized Fourier Features (QFF)} \cite{Lee2022ARXIV} factorizes internal features into bins of Fourier features, corresponding to our Factor Fields framework when using $N=1$, $\bgamma_1(\bx)=\text{sinusoidal}(\bx)$, $\bff_1(\bx)=\text{2D-Maps}(\bx)$, and $\cP(\bx)=\text{MLP}(\bx)$ for the $2$D signal representation.

\boldparagraph{Dictionary Field (DiF)}
Besides existing representations, our Factor Fields framework enables the design of novel representations with desirable properties.
As an example, we now discuss the DiF representation which we have already introduced in \eqnref{eq:expansion_3}.
%
DiF offers implicit regularization, compactness and fast optimization while also generalizing across multiple signals. 
The central idea behind the DiF representation is to decompose the target signals into two fields: a global field (\ie, the basis) and a local field (\ie, the coefficient). The global field promotes structured signal signatures shared across spatial locations and scales as well as between signals, while the local field allows for spatially varying content.
More formally, DiF factorizes the target signal into coefficient fields $\bff_1(\bx)=\bc(\bx)$ and basis functions $\bff_2(\bx)=\bb(\bx)$ which differ primarily by their respective coordinate transformation: We choose the identity mapping $\bgamma_1(\bx) = \bx$ for $\bc(\bx)$ and a periodic coordinate transformation $\bgamma_2(\bx)$ for $\bb(\bx)$, see \figref{fig:trans_func} (top). 
As representation of the two factor fields $\bff_1$ and $\bff_2$ we may choose any of the ones illustrated in \figref{fig:teaser} (bottom-left). To facilitate comparison with previous representations, we use the sawtooth function as the basis coordinate transformation $\bgamma_2$ and uniform grids (\ie, 2D Maps for 2D signals and 3D Grids for 3D signals) as the representation of the coefficient fields $\bff_1$ and basis functions $\bff_2$ for most of our experiments. Besides, we also systematically ablate the number of factors, the number of levels, the coordinate transformation and the field representation in \secref{sec:designs}.

\section{Experiments}
\label{sec:experiments}
We now present extensive evaluations of our Factor Fields framework and DiF representation. 
We first briefly discuss our implementation and hyperparameter configuration. We then compare the performance of DiF with previously proposed representations on both per-signal reconstruction (optimization) and across-signal generalization tasks. At the end of this section, we examine the properties of our Factor Fields framework by varying the number of factors $N$, level number $L$, different types of transformation functions $\bgamma_i$ field representation $\bff_i$, and field connector $\circ$.

\subsection{Implementation}
\label{sec:implementation}
We implement our Factor Fields framework using vanilla PyTorch without customized CUDA kernels. Performance is evaluated on a single RTX 6000 GPU using the Adam optimizer \cite{Kingma2015ICLR} with a learning rate of $0.02$.

We instantiate \revised{DiF} using $L=6$ levels with frequencies (linearly increasing) $f_l \in [2., 3.2, 4.4, 5.6, 6.8, 8.]$, and \revised{feature channels} $K = [4,4,4,2,2,2]\revised{^\top}\cdot2$\revised{$^\eta$}, where \revised{$\eta$} controls the number of \revised{feature channels}. We use $\revised{\eta} = 3$ for our 2D experiments and \revised{$\eta$}$ = \revised{0}$ for our 3D experiments. 
%
The model parameters $\theta$ are distributed across 3 model components: coefficients $\theta_\bc$, basis $\theta_\bb$, and projection function $\theta_{\cP}$.
The size of each component can vary greatly depending on the chosen representation. 

In the following experiments, we refer to the default model setting as ``DiF-Grid'', which implements the coefficients $\bc$ and bases $\bb$ with learnable tensor grids, $\cP(\bx)=\text{MLP}(\bx)$, and $\bgamma(\bx)=\text{Sawtooth}(\bx)$, where $\text{Sawtooth}(\bx) = \bx \ \text{\revised{mod}}\ 1.0$. In the DiF-Grid setting, the total number of optimizable parameters is mainly determined by the resolution of the coefficient and basis grids $M_{\bc}^l$, $M_{\bb}^l$:
\begin{equation}
\begin{split}
    |\theta| &= |\theta_{\cP}| + |\theta_{\bc}| + |\theta_{\bb}|
      = |\theta_{\cP}| + \sum_{l=1}^L {M_{\bc}^l}^D + \, K_l \cdot {M_{\bb}^l}^D
    \label{eqn:model_size}
\end{split}
\end{equation}
We implement the basis grid using linearly increasing resolutions $M^l_\bb\in [32,128]\revised{^T} \cdot \frac{\revised{min(\bv-\bu)}}{1024}$ with interval $[32,128]$ and scene bounding box $[u,v]$. This leads to increased resolution for modeling higher-resolution signals in our experiments. 
We use the same coefficient grid resolution $M_\bc^l$ across all $L$ levels 
for query efficiency and to lower per-signal memory footprint.

In the following, the different model variants of our Dictionary factorization are labeled "DiF-xx", where "xx" indicates the differences from the default setting ``DiF-Grid''. For example, "-MLP-B" refers to using an MLP basis representation, and "SL" stands for single level. 


\subsection{Single Signals}
\label{sec:per-signal}

We first evaluate the accuracy and efficiency of our DiF-Grid representation on various multi-dimensional signals, comparing it to several recent neural signal representations. Towards this goal, we consider three popular benchmark tasks for evaluating neural representations: 2D image regression, 3D Signed Distance Field (SDF) reconstruction and Radiance Field Reconstruction / Novel View Synthesis. We evaluate each method's ability to approximate high-frequency patterns, interpolation quality, compactness, and robustness to ambiguities and sparse observations. 

\begin{figure*}
  \small\sffamily
  
\setlength{\tabcolsep}{1.0pt}%
\renewcommand{\arraystretch}{1.1}%
\begin{tabular}{cccc}\centering

        \scriptsize  Summer Day &
	\scriptsize  Albert &
	\scriptsize  Pluto &
	\scriptsize  Girl with a Pearl Earring \\
 
	\makecell{\includegraphics[height=0.218\linewidth]{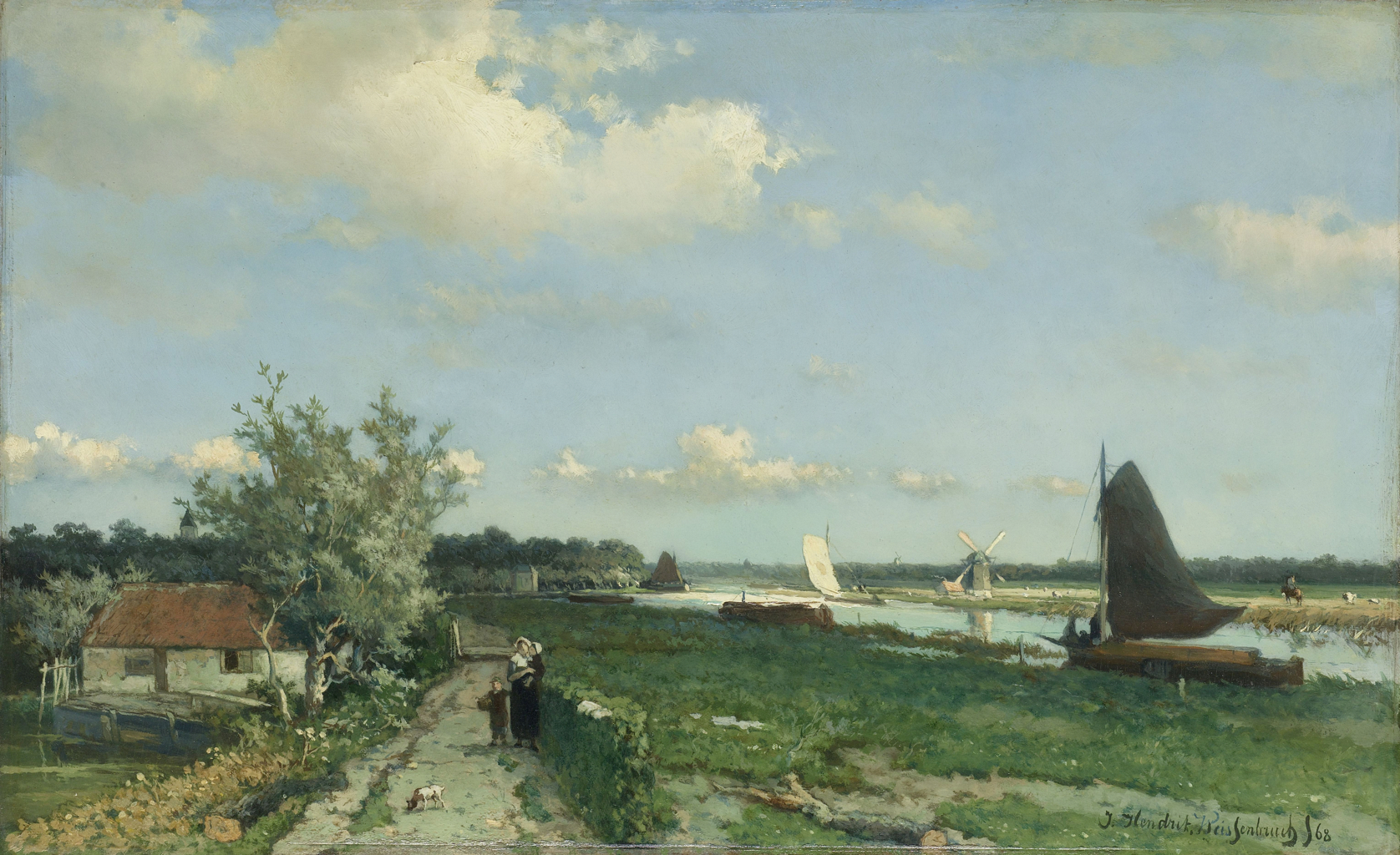}} &
	\makecell{\includegraphics[height=0.218\linewidth]{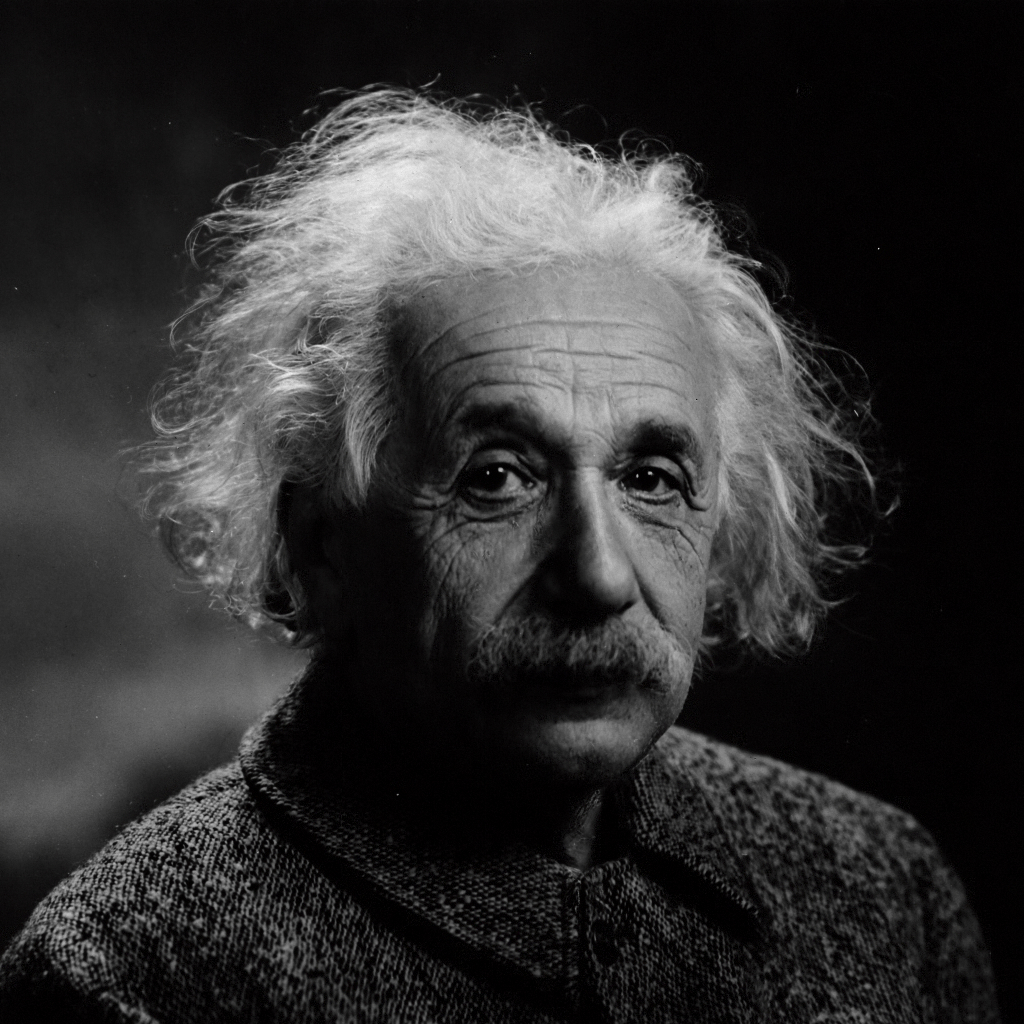}} &
	\makecell{\includegraphics[height=0.218\linewidth]{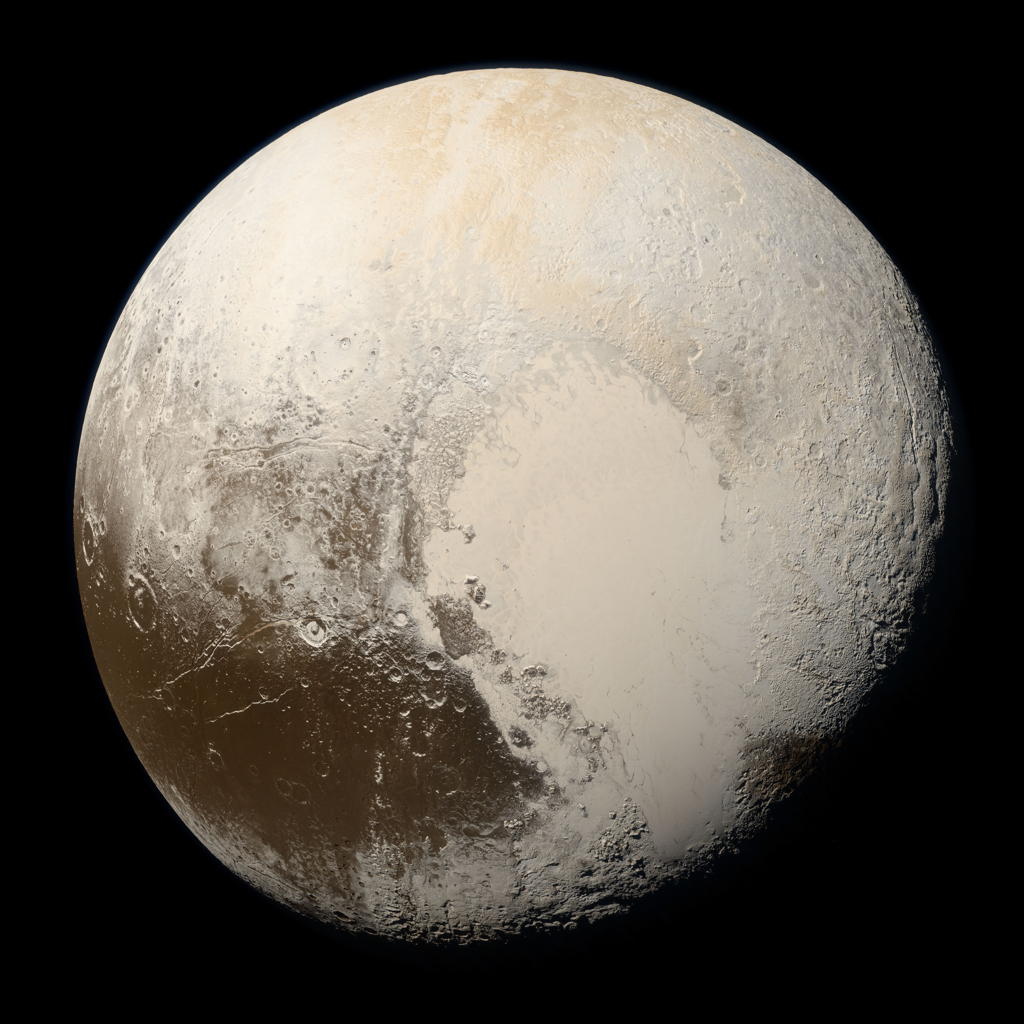}} &
	\makecell{\includegraphics[height=0.218\linewidth]{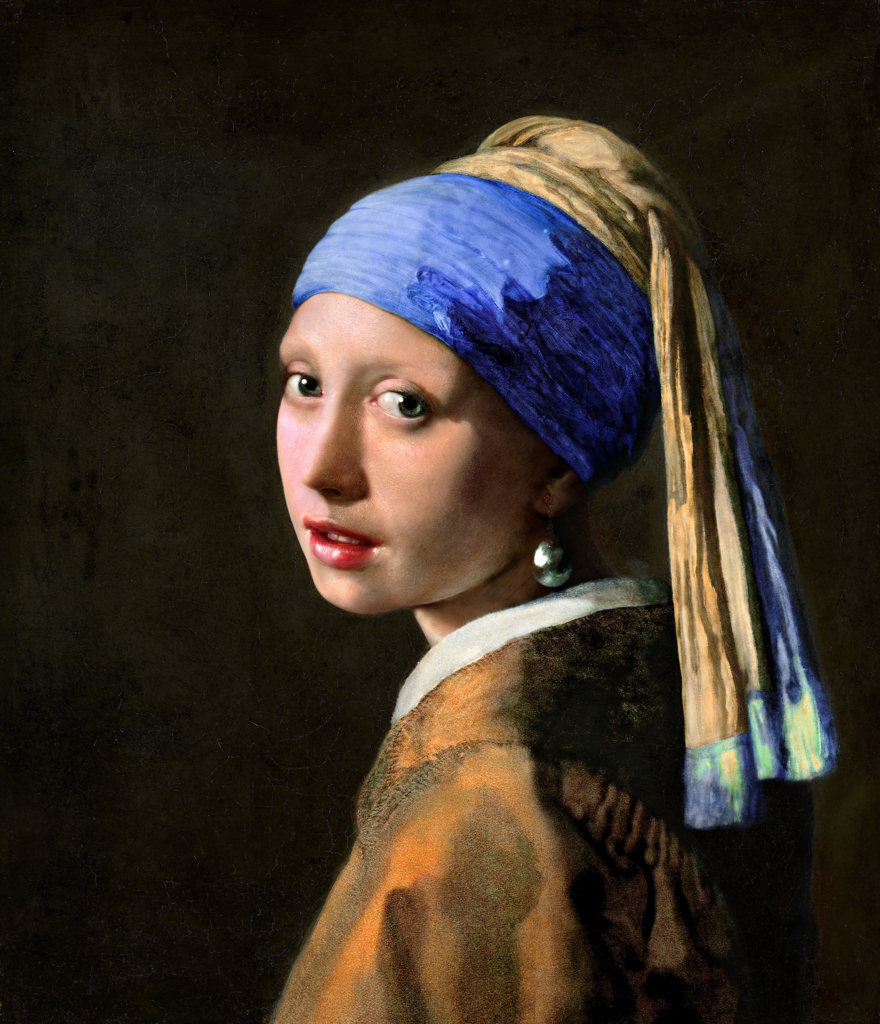}} \\[-1.2mm]
	\scriptsize $6114\times3734\times3$ (resolution) / \SI{35.42}{\mega\nothing} (params)& 
	\scriptsize $1024\times1024\times4$ / \SI{1.36}{\mega\nothing}&
	\scriptsize $8000\times8000\times3$ / \SI{58.60}{\mega\nothing}&
	\scriptsize $8000\times9302\times3$ / \SI{68.52}{\mega\nothing} \\[-1.2mm]
	\scriptsize   $0$:$46$ vs. $4$:$13$ (mm:ss) / $42.4$ vs. $49.0$ dB (PSNR) &
	\scriptsize   $0$:$30$ vs. $1$:$13$ /  $51.0$ vs. $62.7$ &
	\scriptsize  $0$:$50$ vs. $5$:$32$ /  $44.30$ vs. $46.19$ &
	\scriptsize  $1$:$05$ vs. $6$:$06$ / $37.4$ vs. $38.9$ 

\end{tabular}

\vspace{1mm}
\setlength{\tabcolsep}{5pt}%
\begin{tabular}{cccc}

    
    \makecell{\includegraphics[height=0.165\linewidth,width=0.23\linewidth]{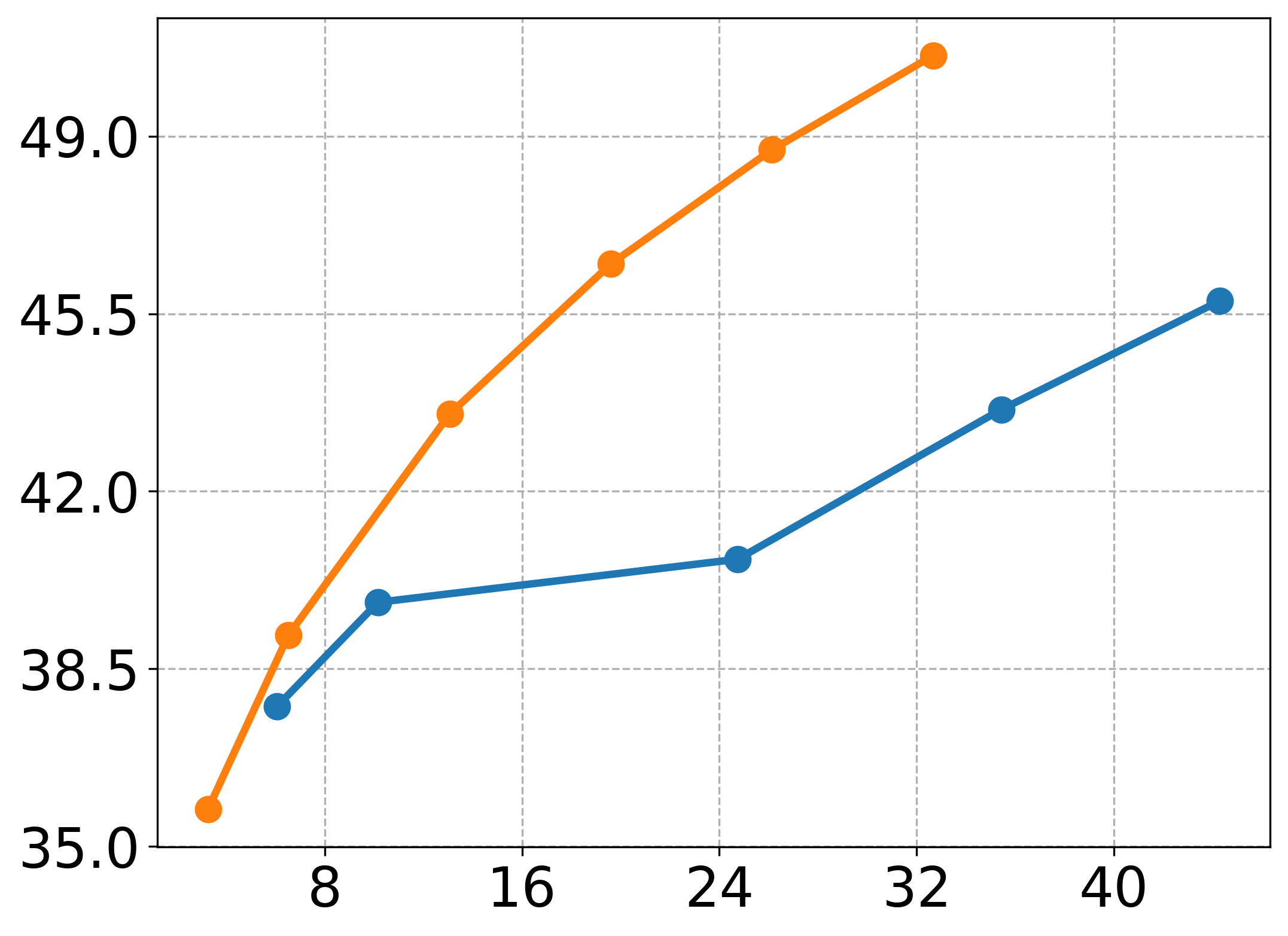}} &
	\makecell{\includegraphics[height=0.165\linewidth,width=0.23\linewidth]{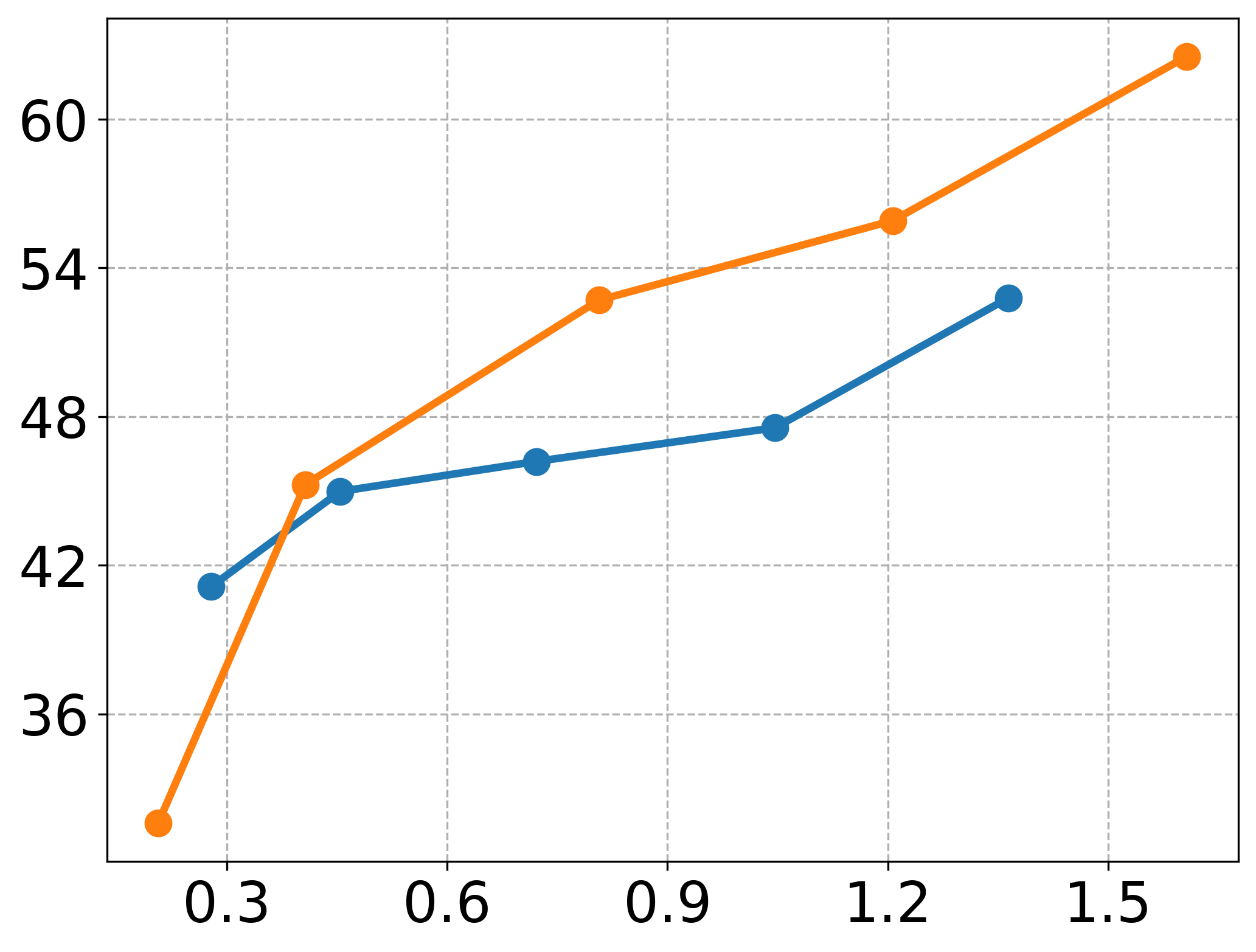}} &
	\makecell{\includegraphics[height=0.165\linewidth,width=0.23\linewidth]{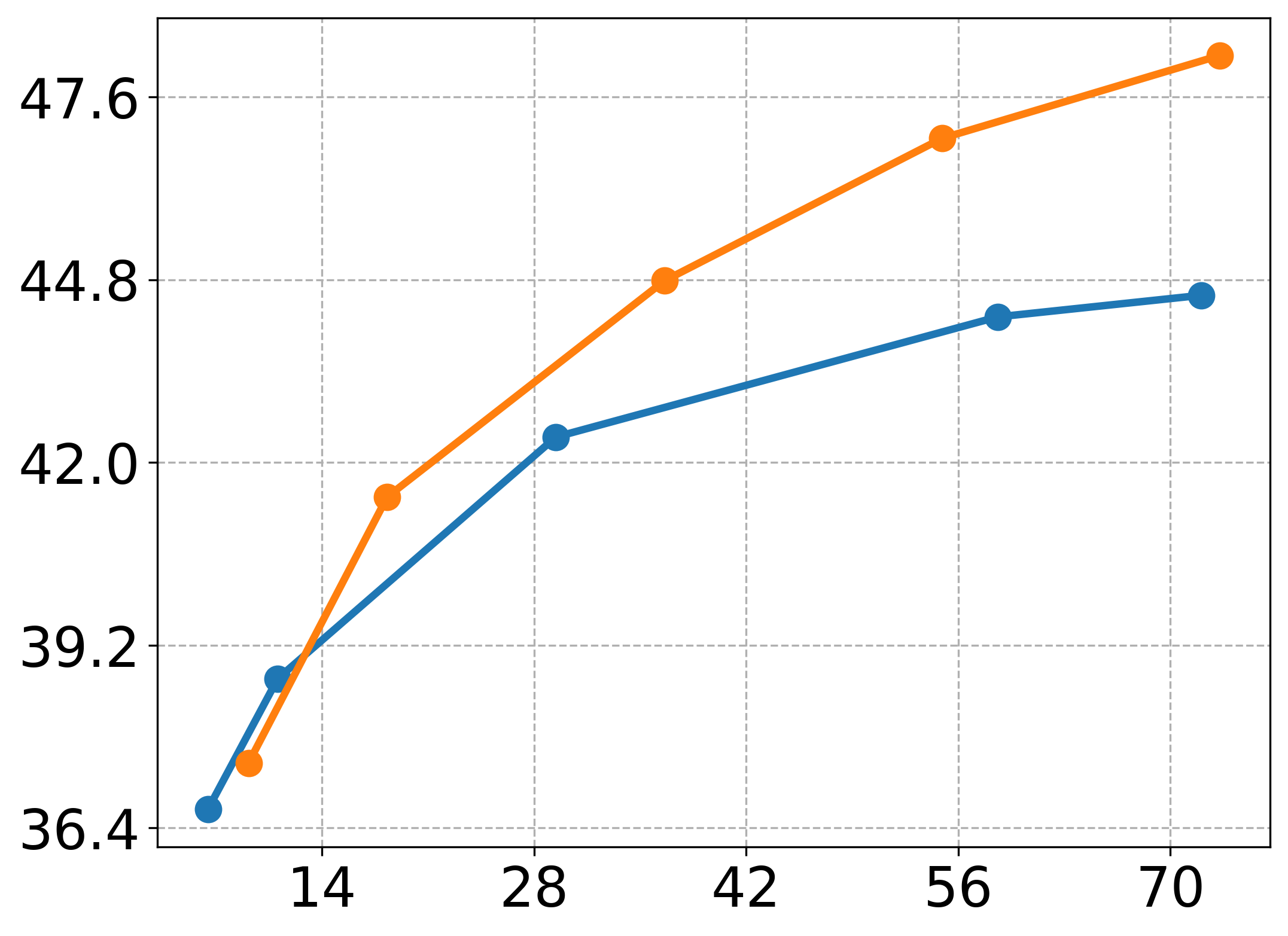}} &
	\makecell{\includegraphics[height=0.165\linewidth,width=0.23\linewidth]{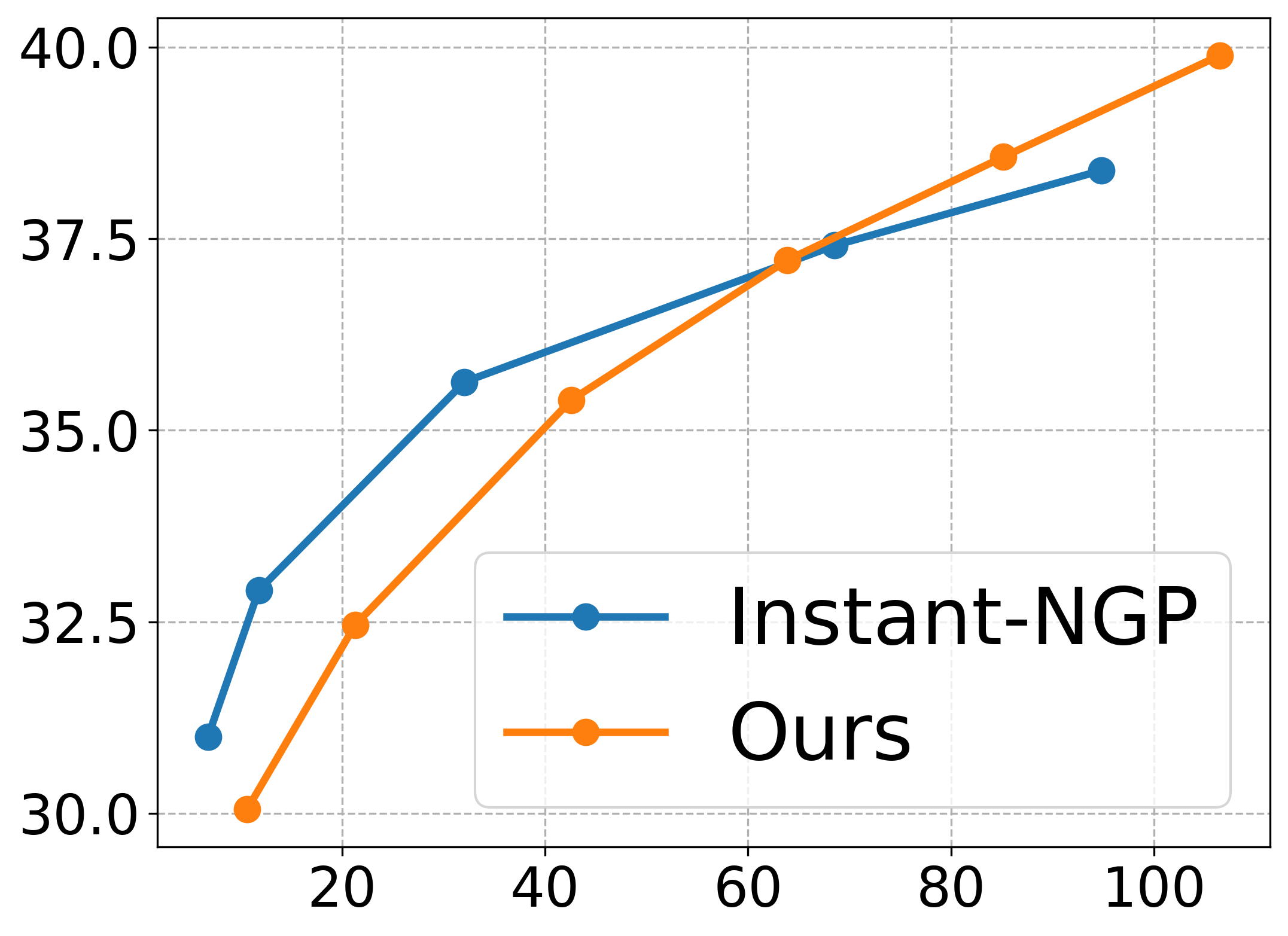}} 

    \put(-67.5,-44) { \small  \revised{params (\SI{}{\mega\nothing)}}}
    \put(-193,-44) { \small \revised{params (\SI{}{\mega\nothing)}}}
    \put(-324,-44) { \small \revised{params (\SI{}{\mega\nothing)}}}
    \put(-441,-44) { \small \revised{params (\SI{}{\mega\nothing)}}}

    \put(-472,35) { \footnotesize PSNR (dB)}
    \put(-352,35) { \footnotesize \revised{PSNR (dB)}}
    \put(-223,35) { \footnotesize \revised{PSNR (dB)}}
    \put(-99,35)  { \footnotesize \revised{PSNR (dB)}}

\end{tabular}
  \vspace{-1mm}
  \caption{\label{fig:img_results}%
    \textbf{2D Image Regression.} This figure shows images represented using our  \revised{DiF-Grid} model. The respective image resolutions and numbers of model parameters are shown below each image. Moreover, we also report a comparison to Instant-NGP (first number) in terms of optimization time and PSNR metrics (Instant-NGP vs Ours) at the bottom using the same number of model parameters. Note that our method achieves better reconstruction quality on all images when using the same model size. While optimization is slower than Instant-NGP, we use a vanilla PyTorch implementation without customized CUDA kernels. ``Summer Day'' credit goes to Johan Hendrik Weissenbruch and \href{https://www.rijksmuseum.nl/en/collection/SK-A-3005}{rijksmuseum}. ``Albert'' credit goes to Orren Jack Turner. ``Pluto'' credit goes to \href{https://solarsystem.nasa.gov/resources/933/true-colors-of-pluto/}{NASA}. ``Girl With a Pearl Earring'' renovation \copyright Koorosh Orooj \href{http://profoundism.com/free_licenses.html}{(CC BY-SA 4.0)}.
  }
\end{figure*}

\boldparagraph{2D Image Regression}
\label{sec:image_regression}
In this task, we directly regress RGB pixel colors from pixel coordinates. 
We evaluate our DiF-Grid on fitting four complex high-resolution images, where the total number of pixels ranges from \SI{4}{\mega\nothing} to \SI{213}{\mega\nothing}.
In \figref{fig:img_results}, we show the reconstructed images with the corresponding model size, optimization time, and image PSNRs, and compare them to Instant-NGP \cite{Mueller2022TOG}, a state-of-the-art neural representation that supports image regression and has shown superior quality over prior art including Fourier Feature Networks \cite{Tancik2020NEURIPS} and SIREN \cite{Sitzmann2020NIPS}.
Compared to Instant-NGP, our model consistently achieves higher PSNR on all images when using the same model size, demonstrating the superior accuracy and efficiency of our model.
On the other hand, while Instant-NGP achieves faster optimization owing to its highly optimized CUDA-based framework, 
our model, implemented in pure PyTorch, leads to comparably fast training while relying on a vanilla PyTorch implementation without custom CUDA kernels which simplifies future extensions.


\boldparagraph{Signed-Distance Field Reconstruction}
Signed Distance Function (SDF), as a classic geometry representation, describes a set of continuous iso-surfaces, where a 3D surface is represented as the zero level-set of the function. 
We evaluate our DiF-Grid on modeling several challenging object SDFs that contain rich geometric details and compare with previous state-of-the-art neural representations, including Fourier Feature Networks~\cite{Tancik2020NEURIPS}, SIREN~\cite{Sitzmann2020NIPS}, and Instant-NGP~\cite{Mueller2022TOG}. 
To allow for fair comparisons in terms of the training set and convergence, we use the same training points for all methods by pre-sampling \SI{8}{\mega\nothing} SDF points from the target meshes for training, with $80\%$ points near the surface and the remaining $20\%$ points uniformly distributed inside the unit volume. 
Following the evaluation setting of Instant-NGP, we randomly sample \SI{16}{\mega\nothing} points for evaluation and calculate the geometric IOU metric based on the SDF sign
%
%
\begin{equation}
    gIoU = \frac{\sum (s(\bX)>0) \cap (\hat{s}(\bX)>0)}{\sum (s(\bX)>0) \cup (\hat{s}(\bX)>0)}
    \label{eqn:iou}
\end{equation}
where $\bX$ is the evaluation point set, $s(\bX)$ are the ground truth SDF values, and $\hat{s}(\bX)$ are the predicted SDF values.

\begin{figure*}
  \small\sffamily
  
\setlength{\tabcolsep}{1.8pt}%
\renewcommand{\arraystretch}{1.1}%
\hspace*{-4.0mm}\begin{tabular}{c@{\hskip 2.0mm}ccccc@{\hskip 2.0mm}c}
	\makecell{DiF (ours)} &
	\makecell{SIREN} &
	\makecell{Frequency} &
	\makecell{Instant-NGP} &
	\makecell{DiF-Grid} &
	\makecell{Reference} &
	\makecell{DiF (ours)}
	\\[0.5mm]

	\makecell{\includegraphics[width=0.20\linewidth]{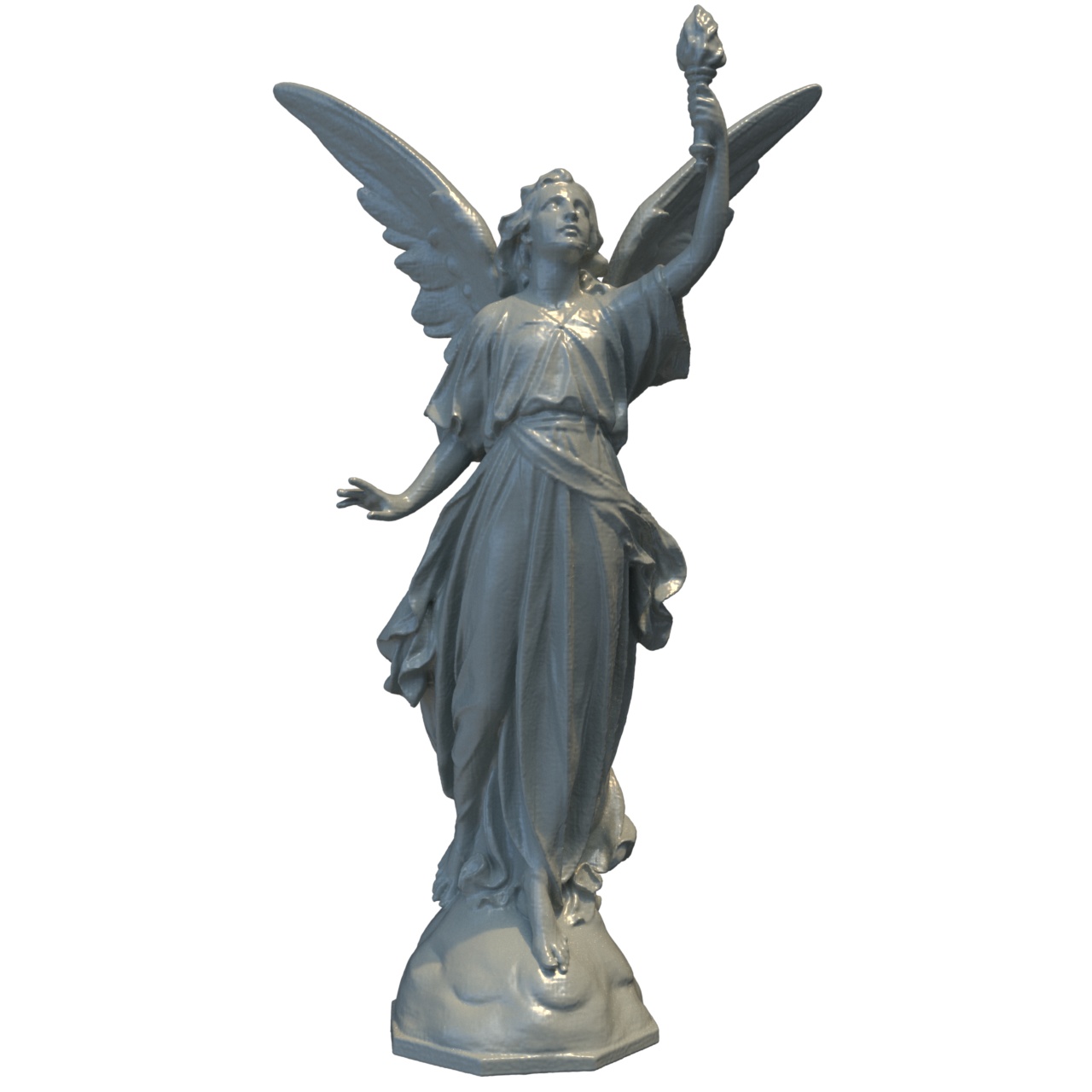}}&
	\makecell{\frame{\includegraphics[width=0.12\linewidth]{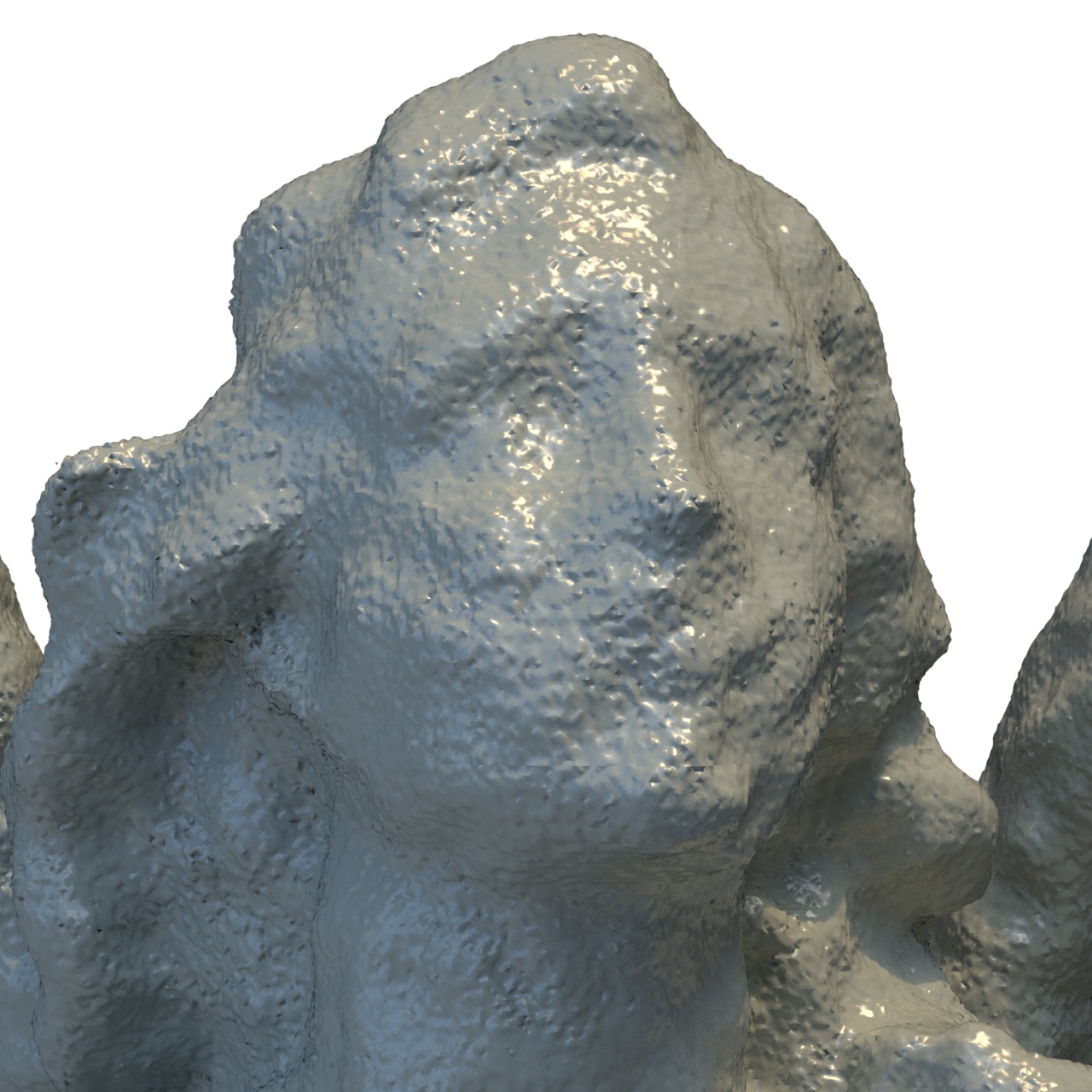}}\\[-0.5mm]\frame{\includegraphics[width=0.12\linewidth]{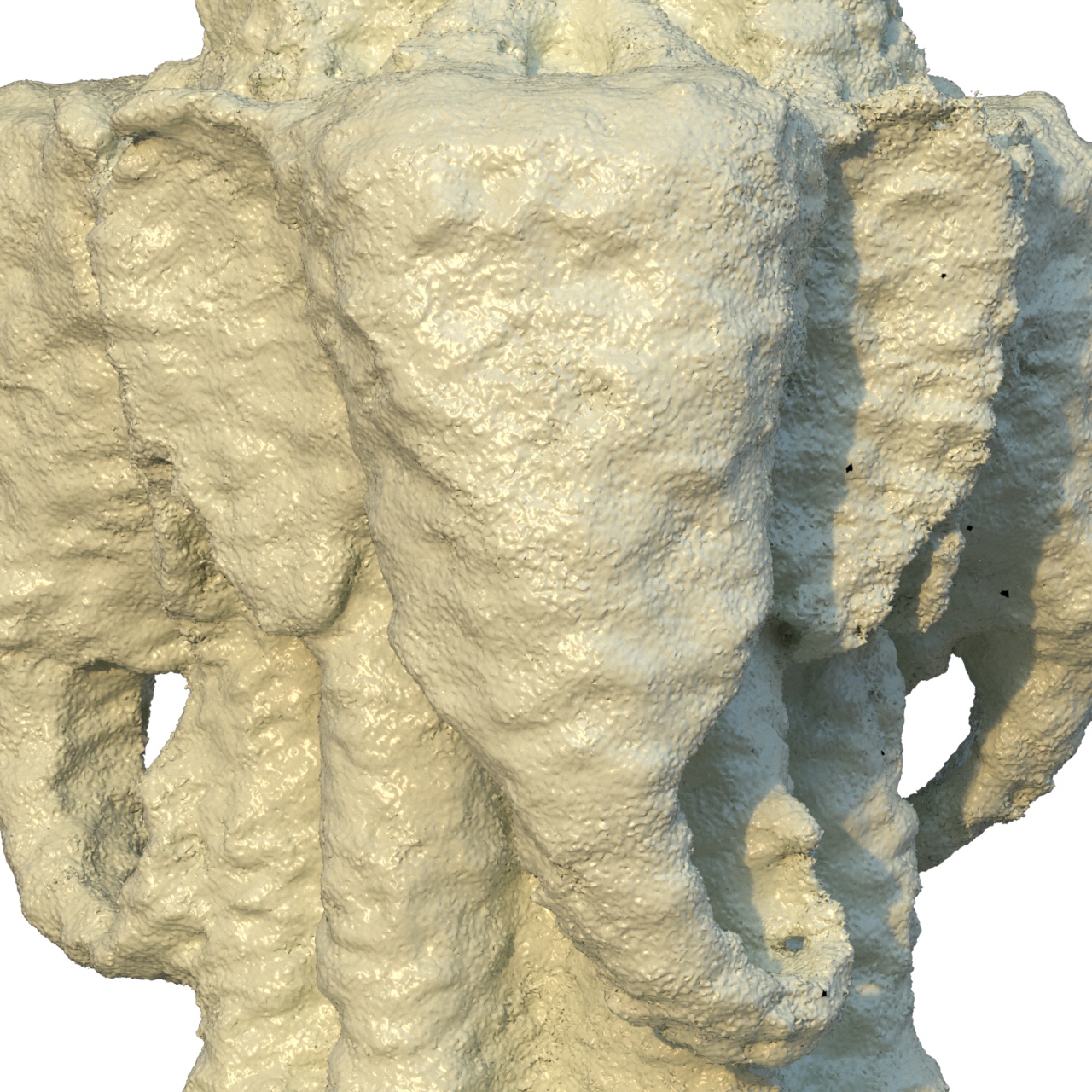}}} &
        \makecell{\frame{\includegraphics[width=0.12\linewidth]{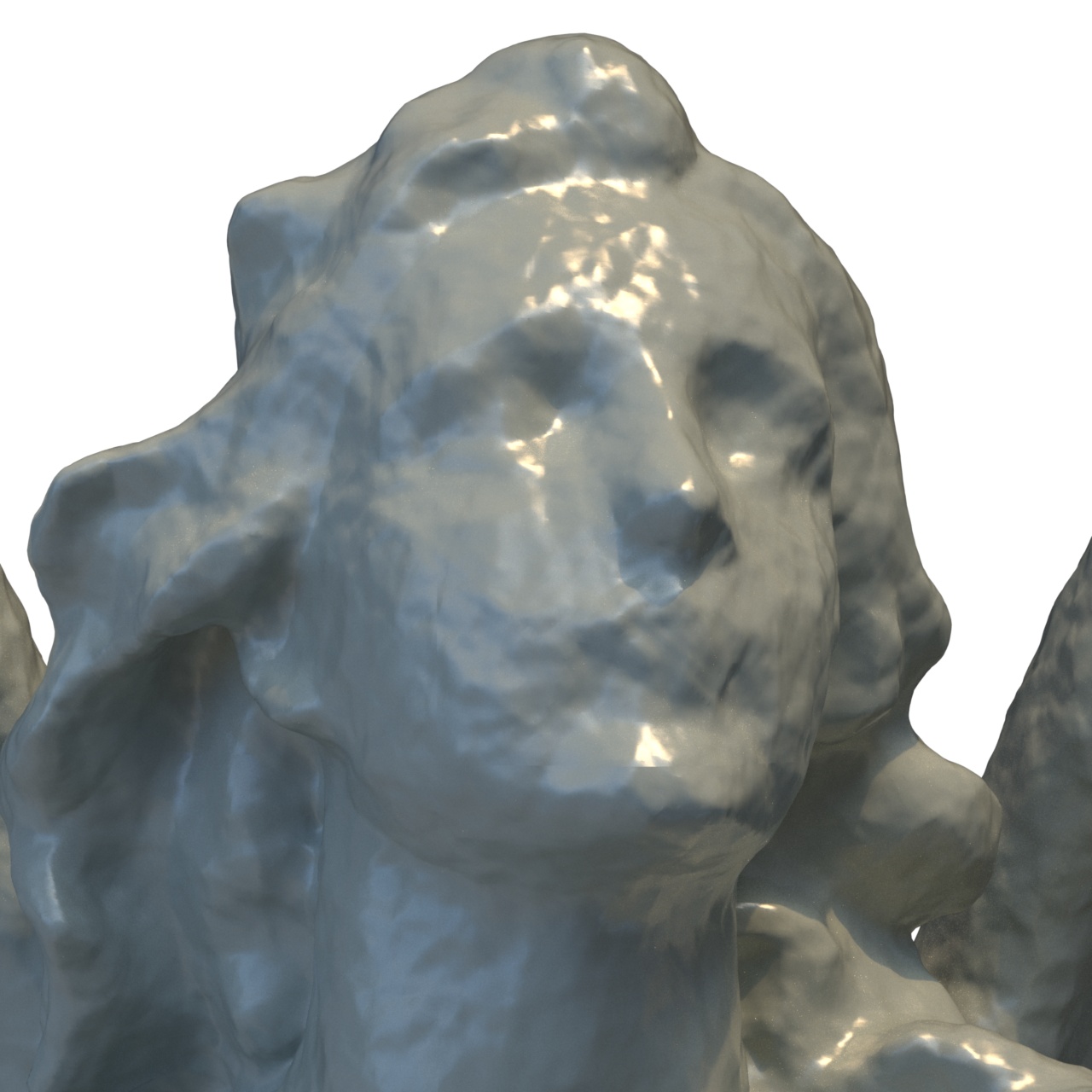}}\\[-0.5mm]\frame{\includegraphics[width=0.12\linewidth]{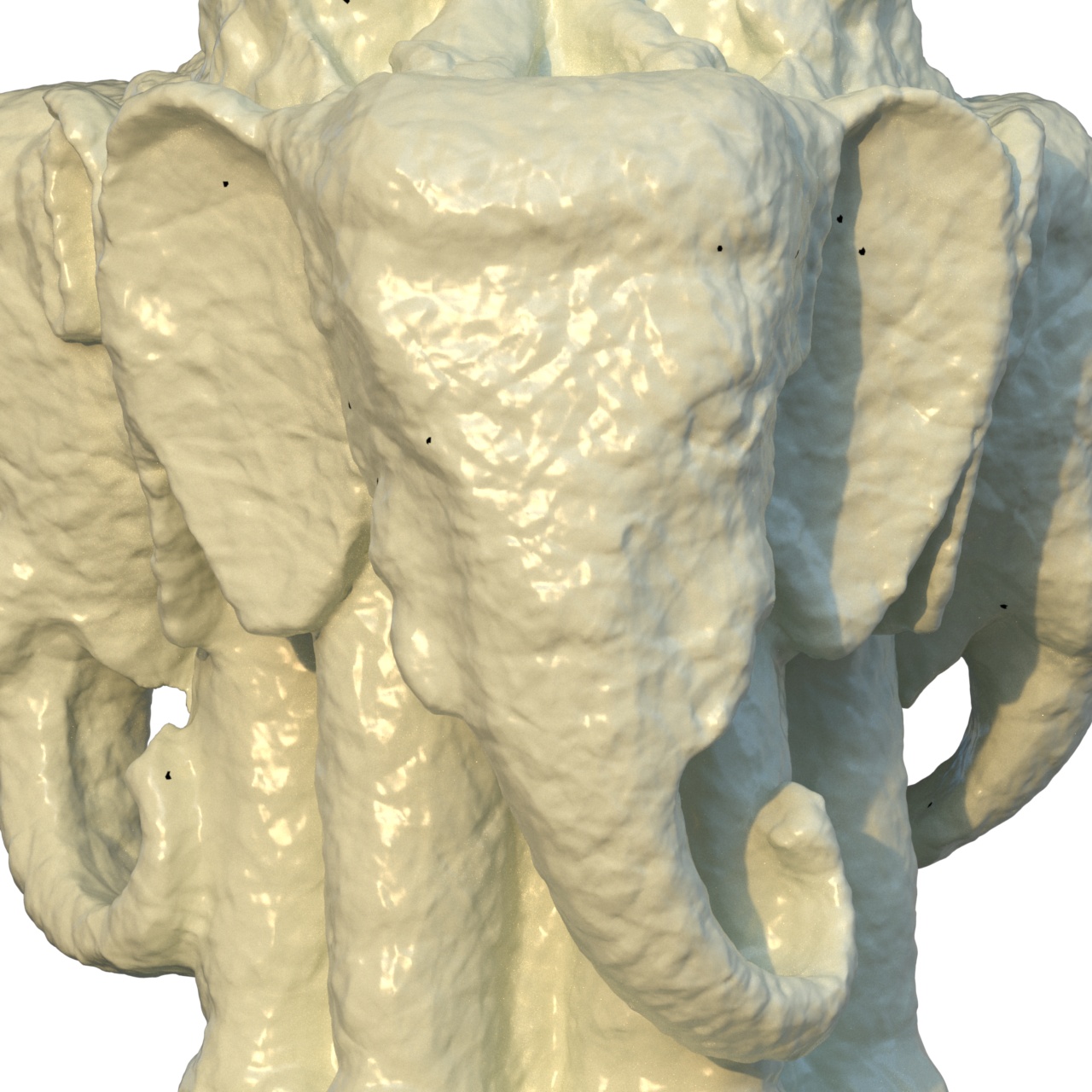}}} &
        \makecell{\frame{\includegraphics[width=0.12\linewidth]{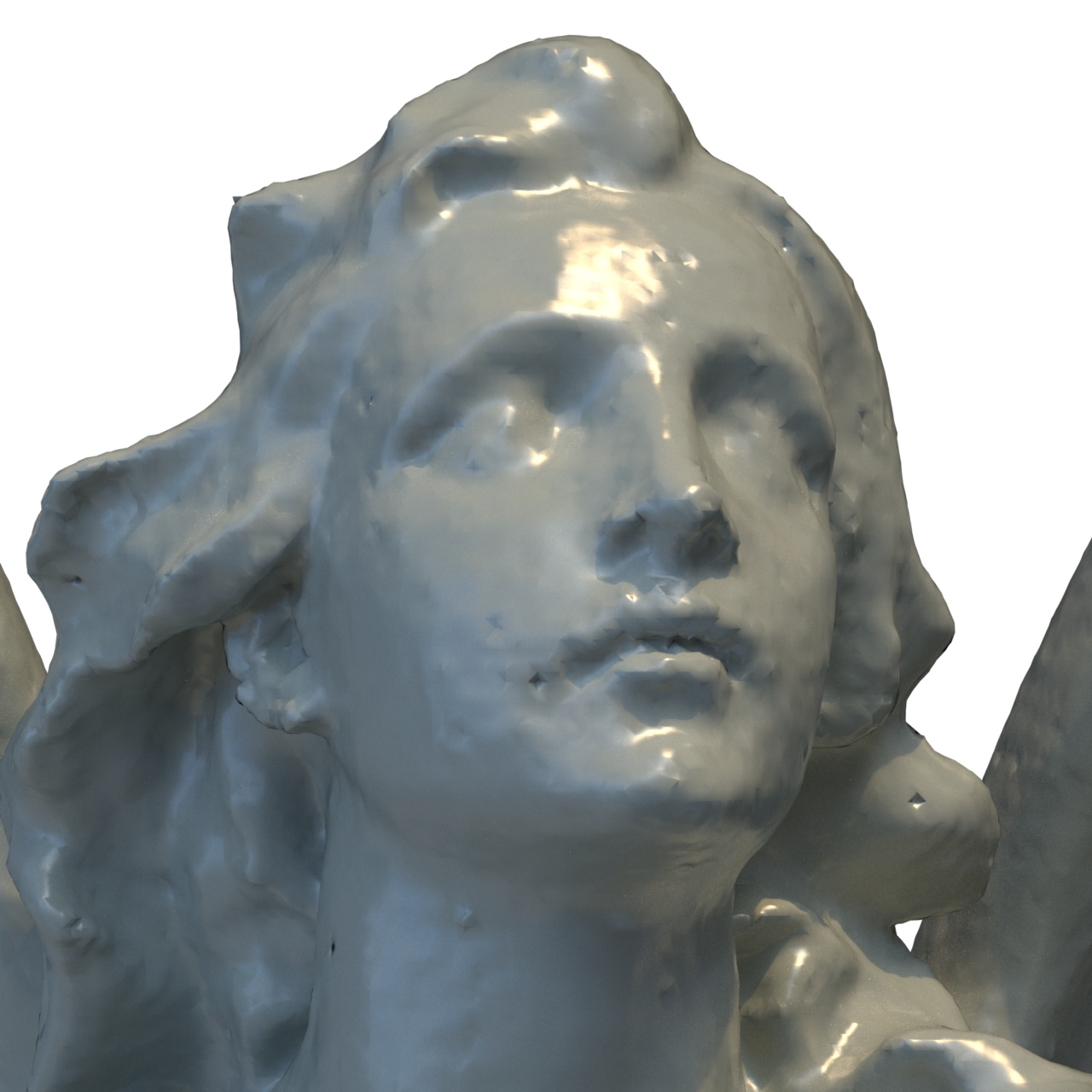}}\\[-0.5mm]\frame{\includegraphics[width=0.12\linewidth]{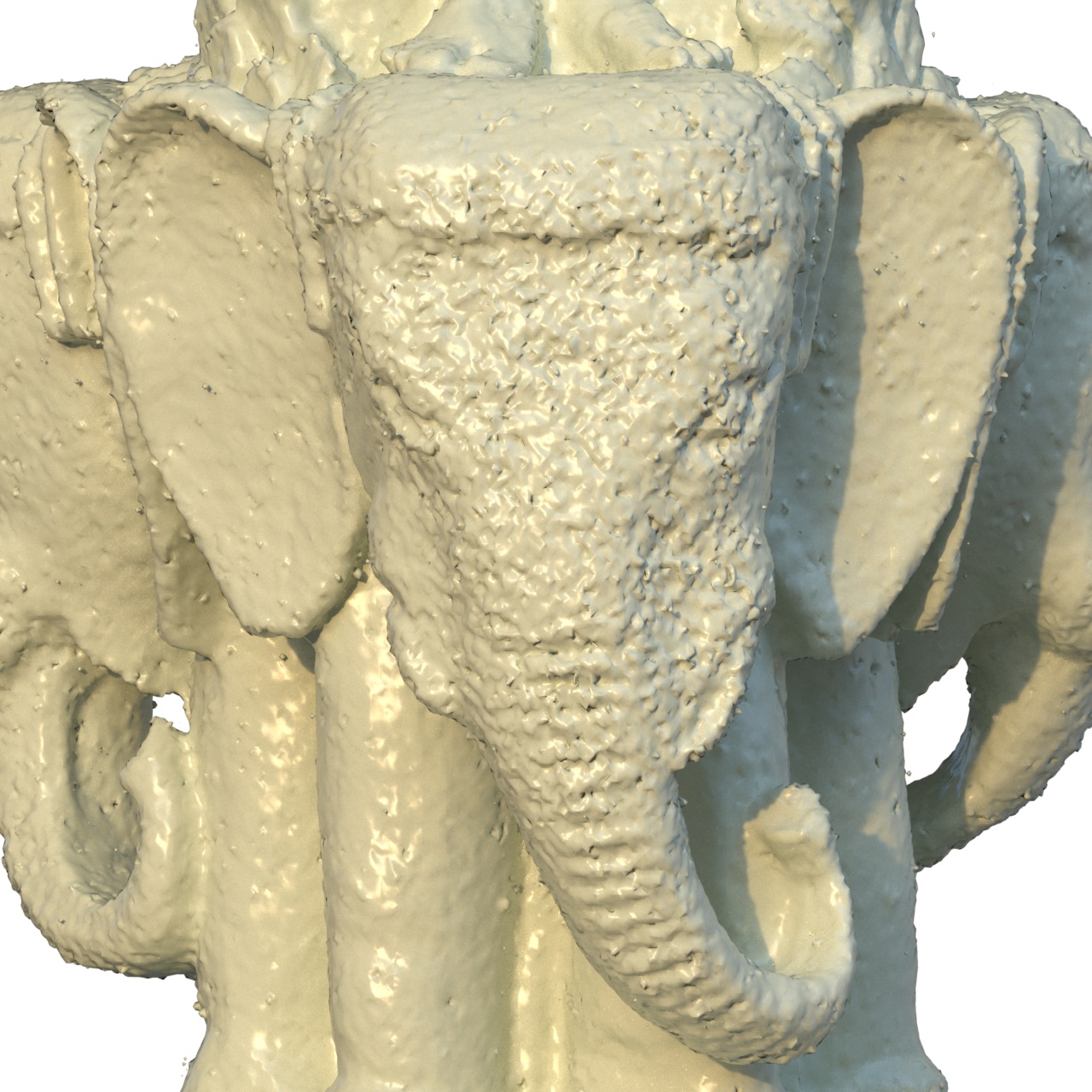}}} &
	\makecell{\frame{\includegraphics[width=0.12\linewidth]{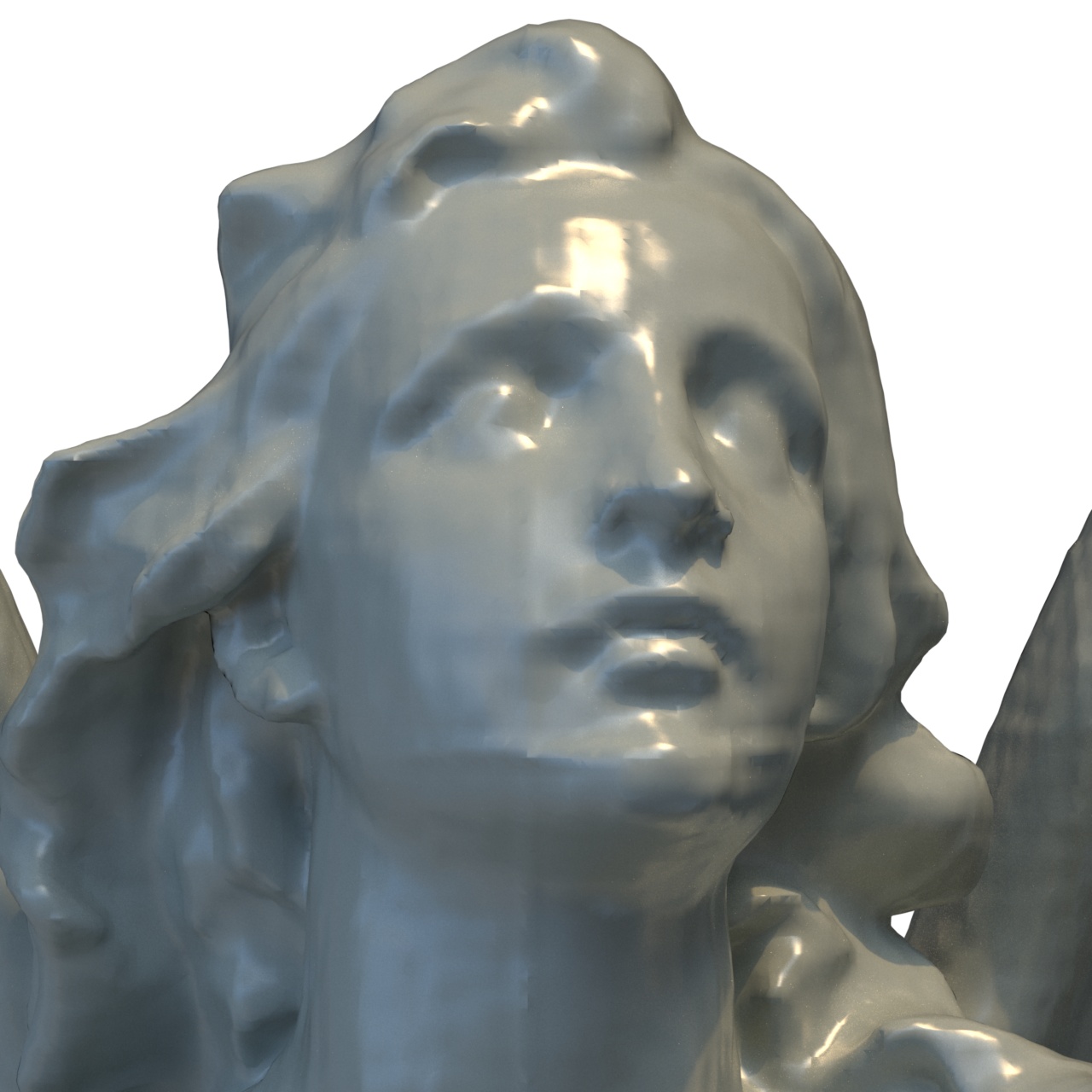}}\\[-0.5mm]\frame{\includegraphics[width=0.12\linewidth]{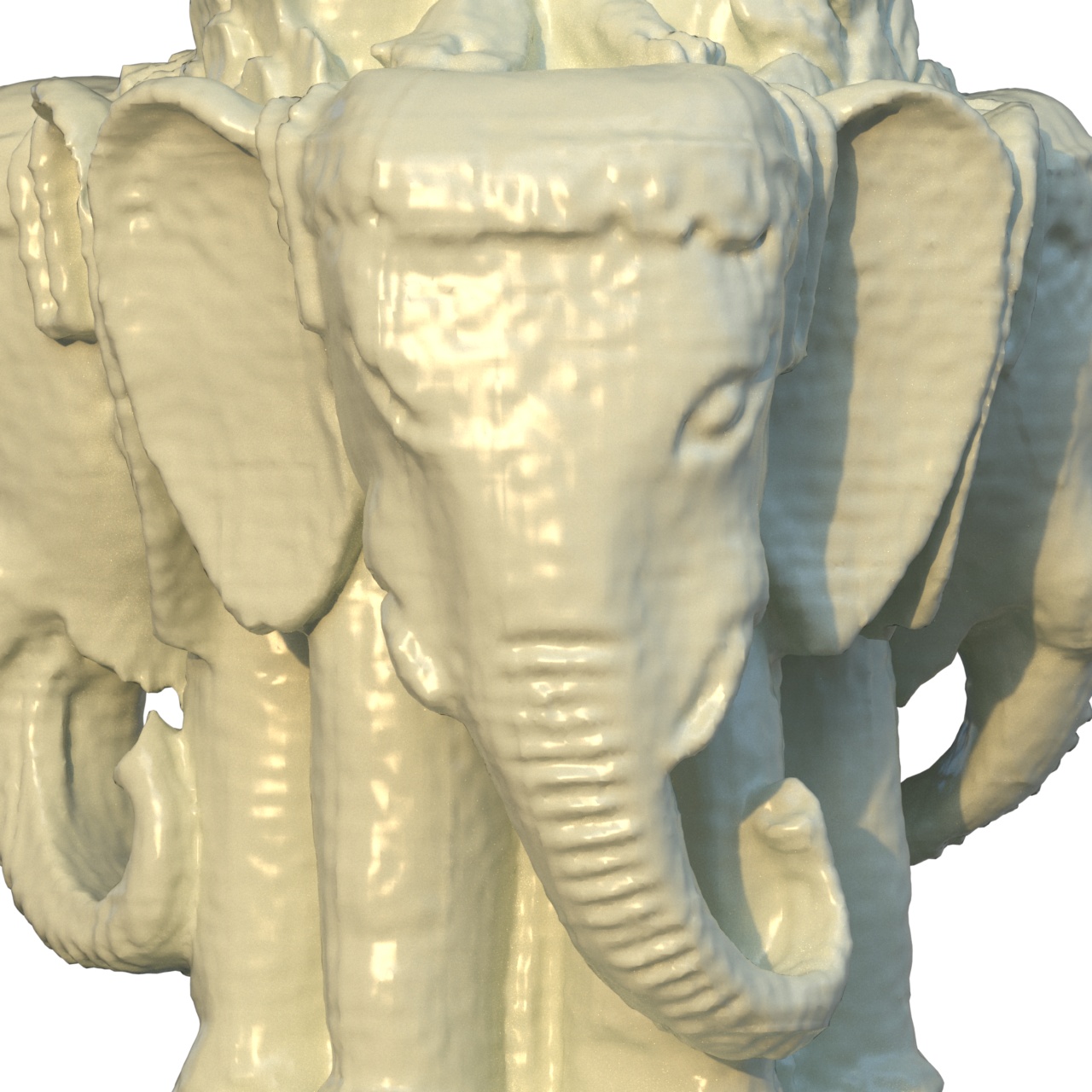}}} &
 	\makecell{\frame{\includegraphics[width=0.12\linewidth]{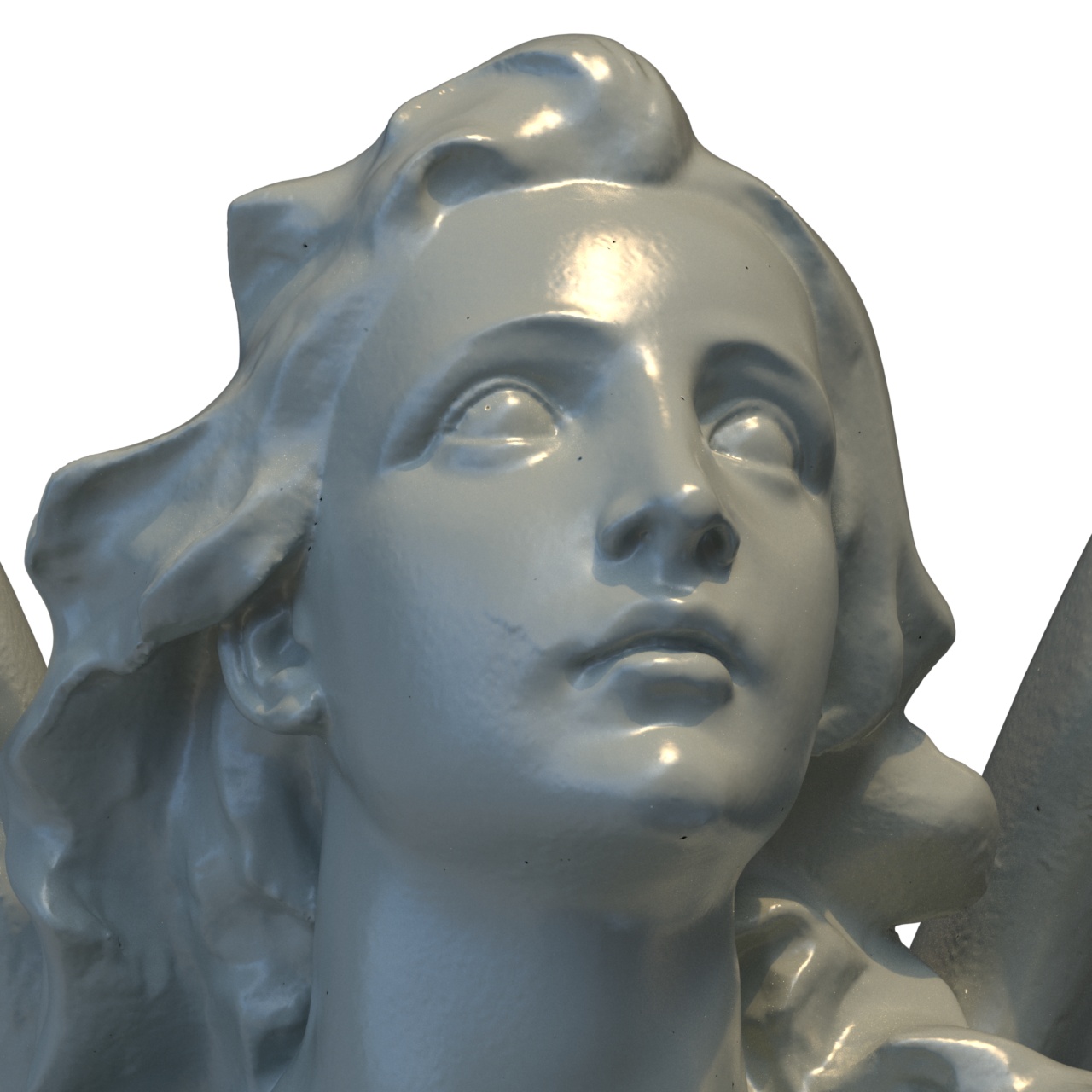}}\\[-0.5mm]\frame{\includegraphics[width=0.12\linewidth]{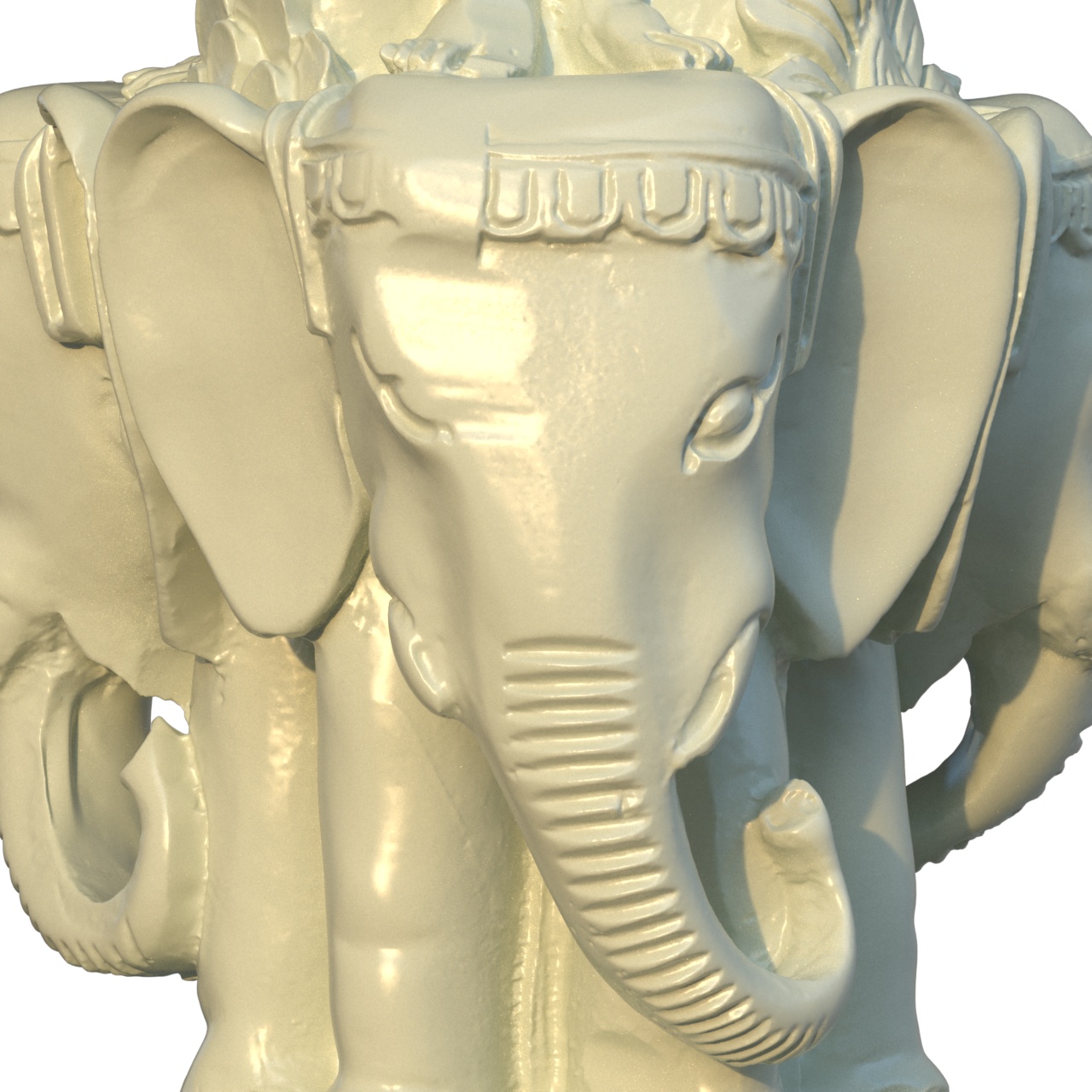}}} &
	\makecell{\includegraphics[width=0.169\linewidth]{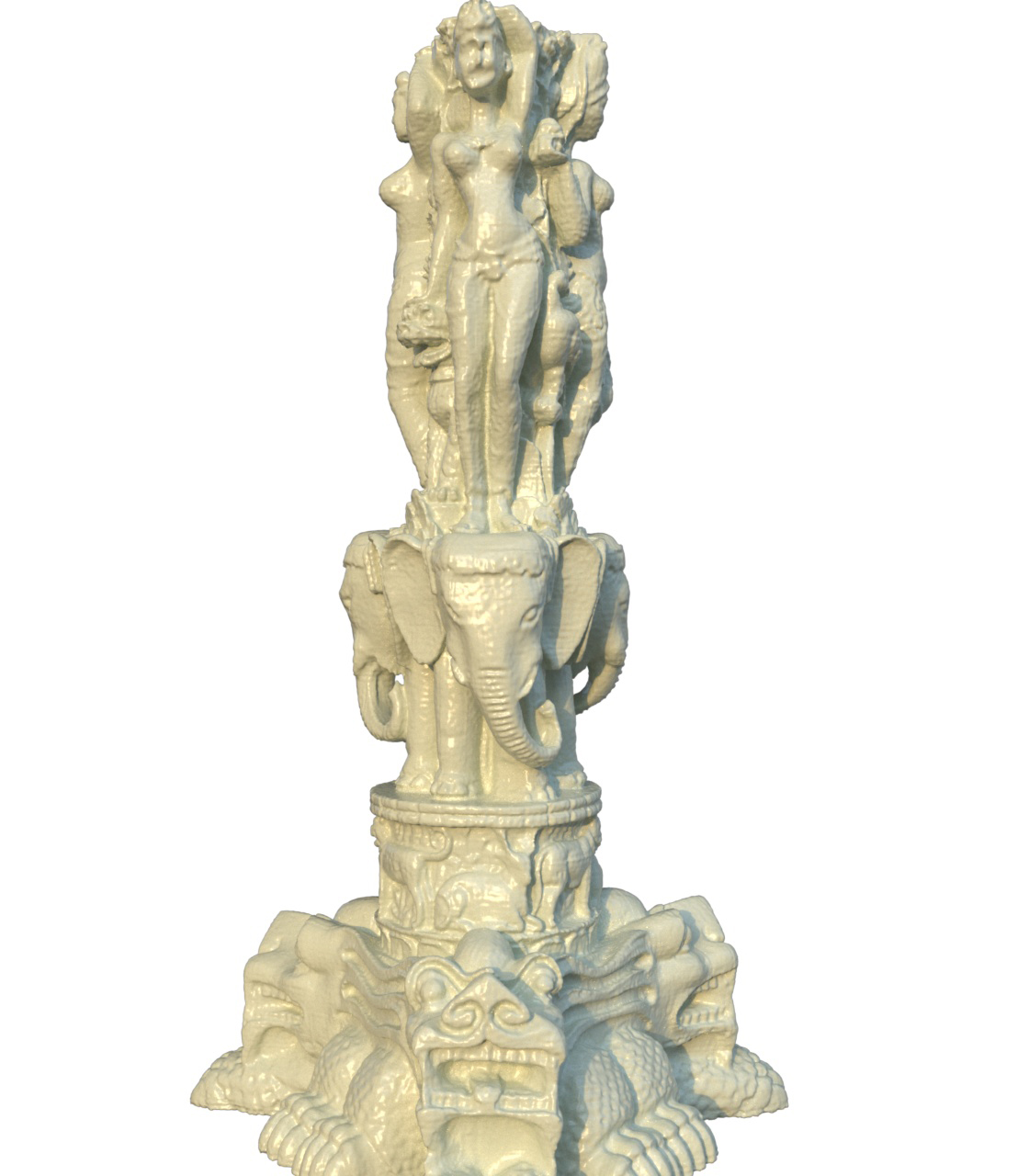}} \\[-5mm] &\\ [-3mm]

        \makecell{\hspace{-1mm}Lucy} & &&& &&\makecell{\hspace{-1mm}Statuette}  \\[-5mm] 

	\footnotesize \\[1.0mm]

	\makecell{\vspace{-8mm} \includegraphics[width=0.2\linewidth]{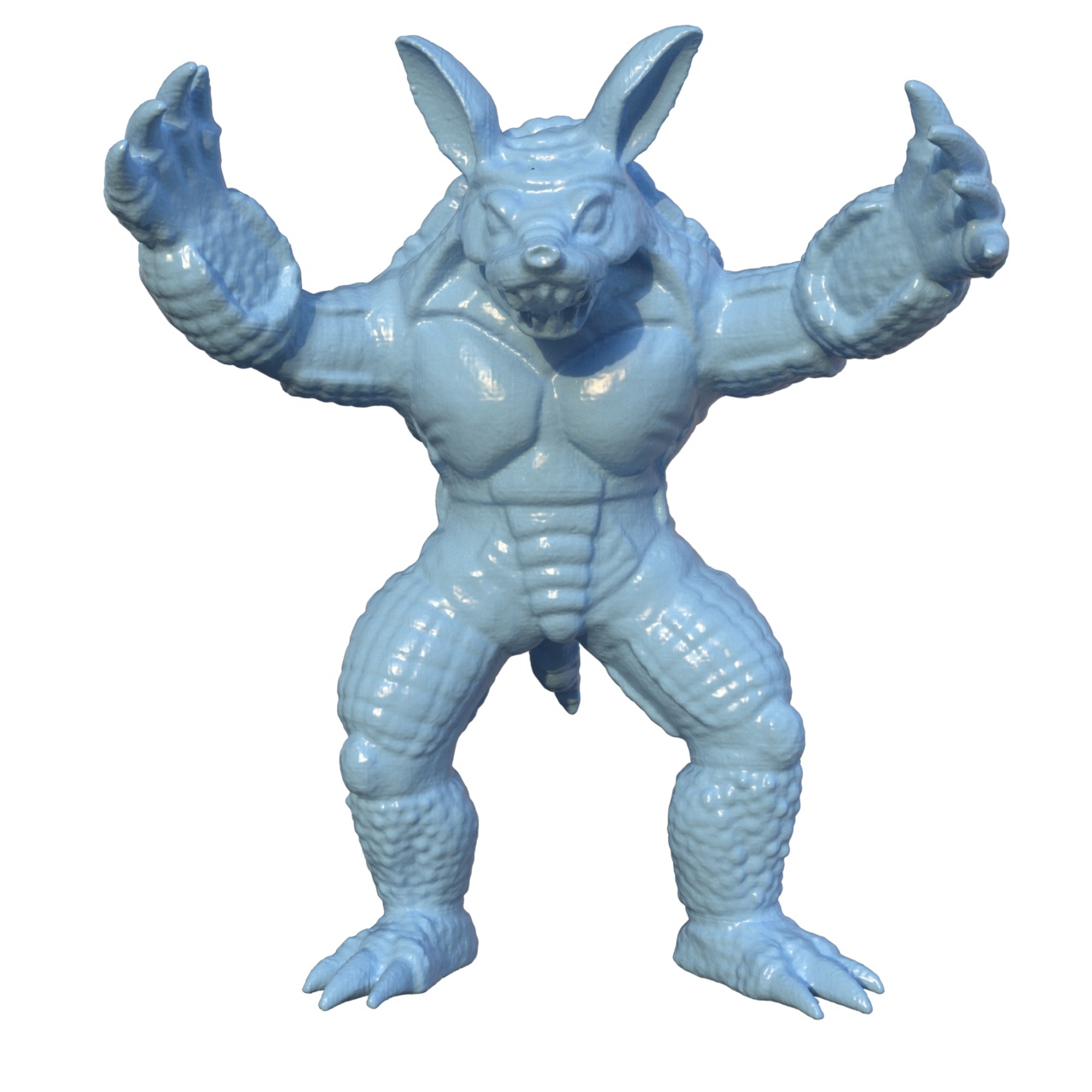}} &
	\makecell{\frame{\includegraphics[width=0.12\linewidth]{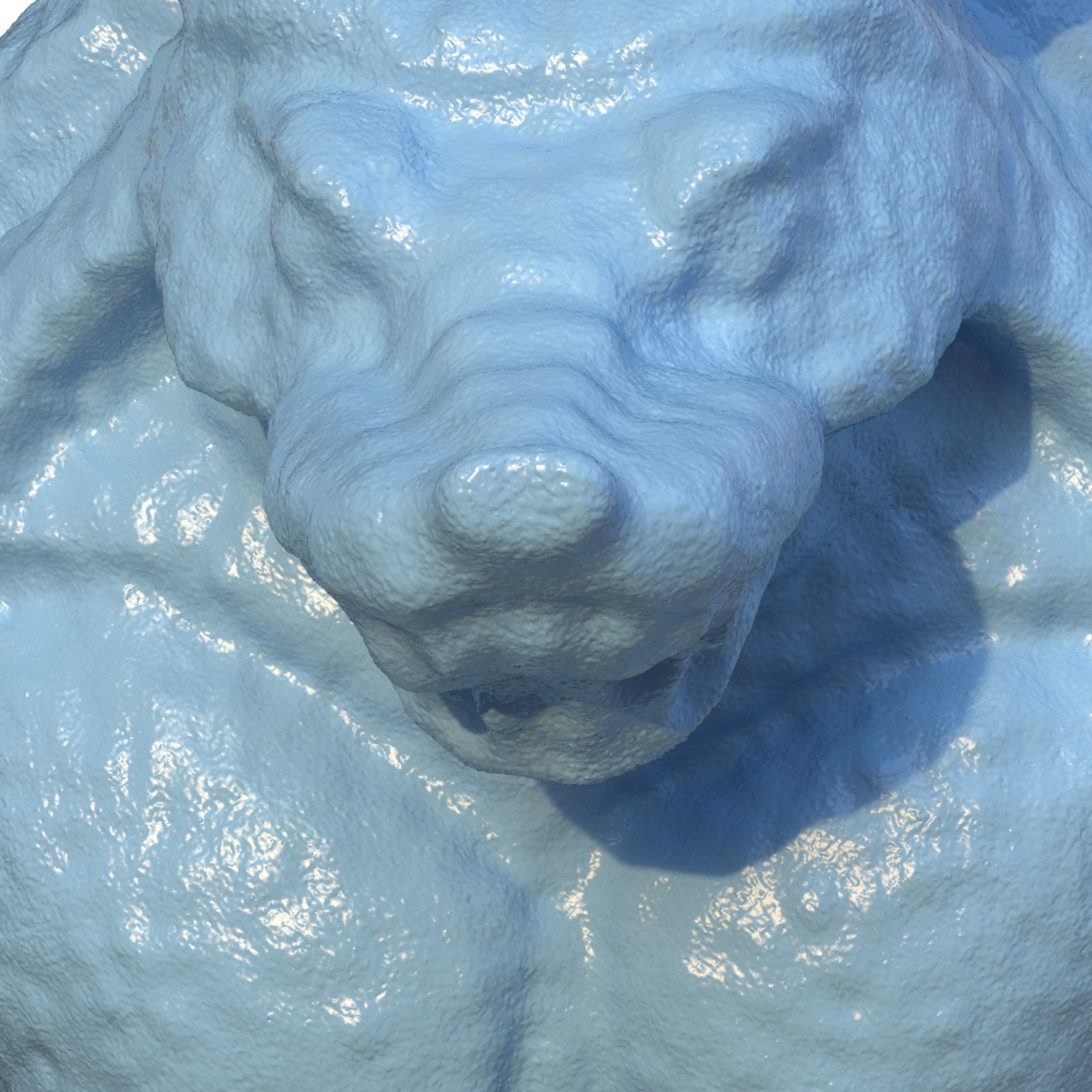}}\\[-0.5mm]\frame{\includegraphics[width=0.12\linewidth]{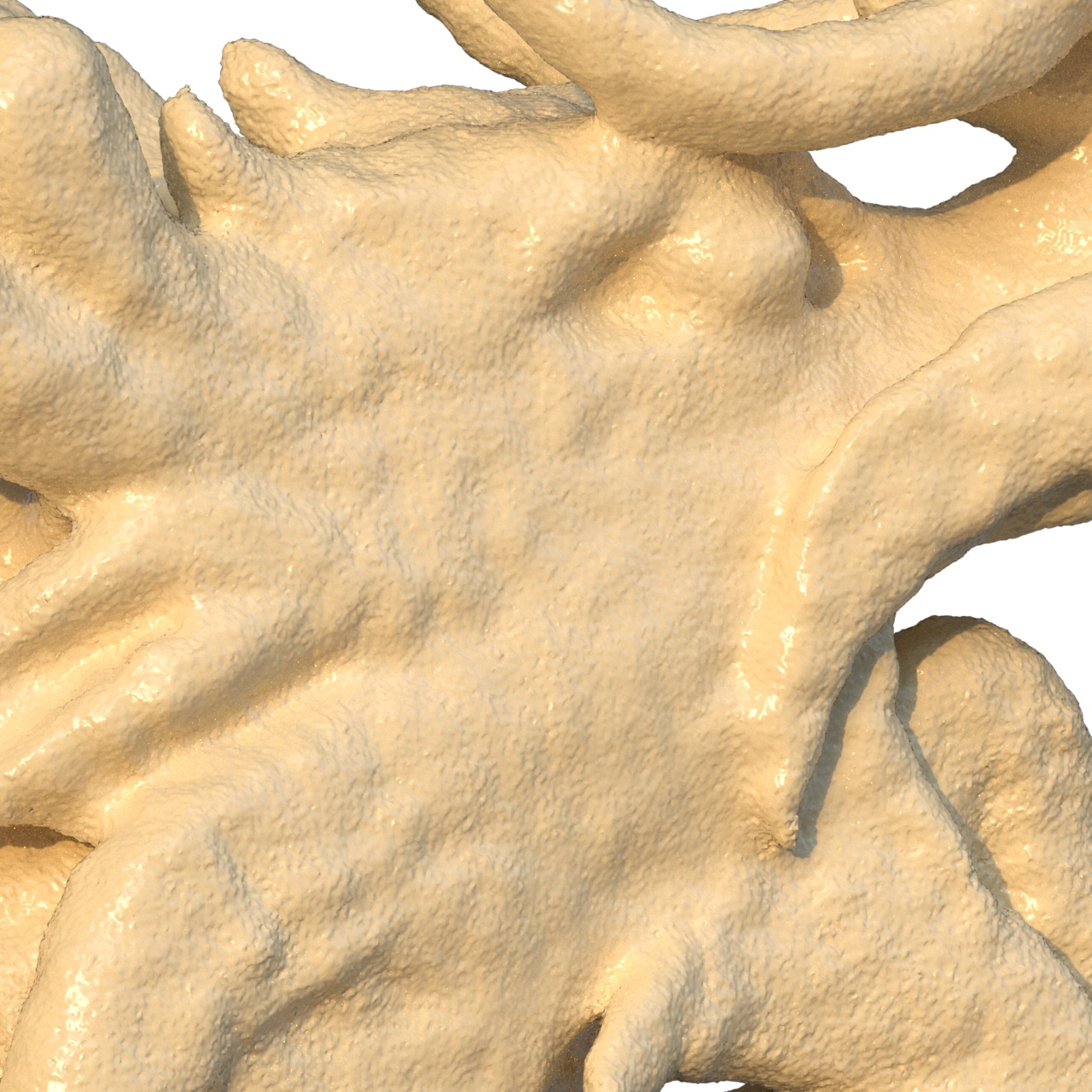}}} &
        \makecell{\frame{\includegraphics[width=0.12\linewidth]{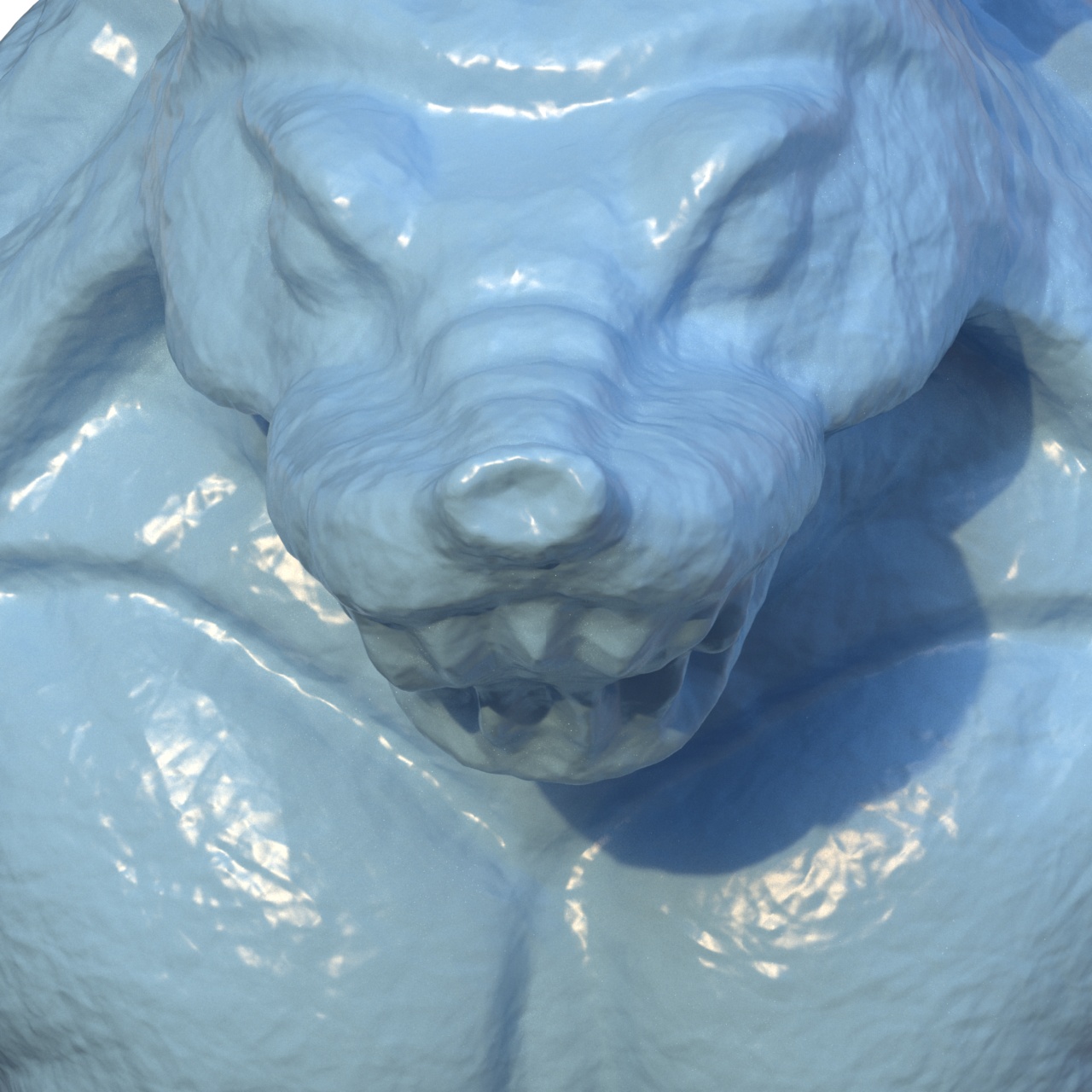}}\\[-0.5mm]\frame{\includegraphics[width=0.12\linewidth]{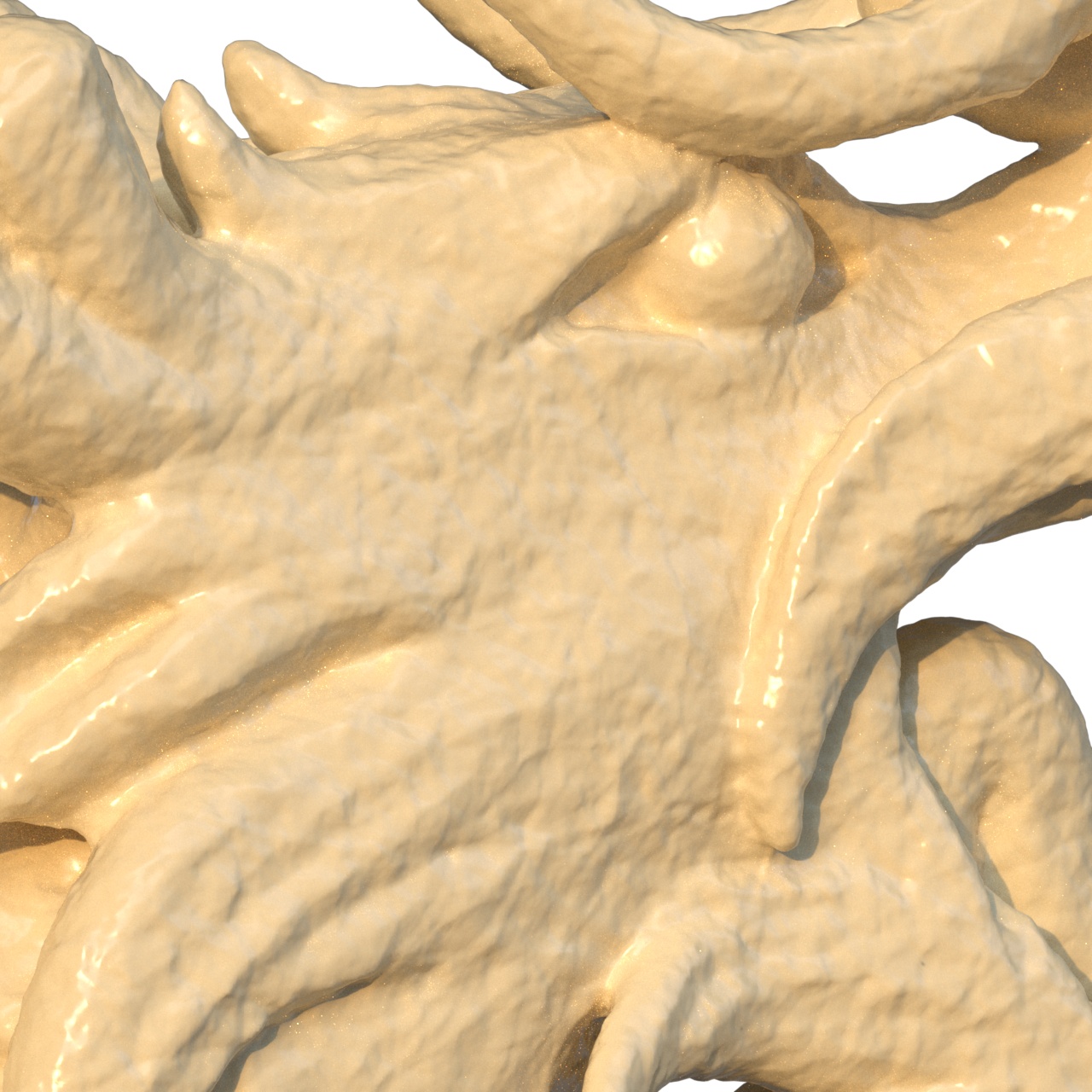}}} &
        \makecell{\frame{\includegraphics[width=0.12\linewidth]{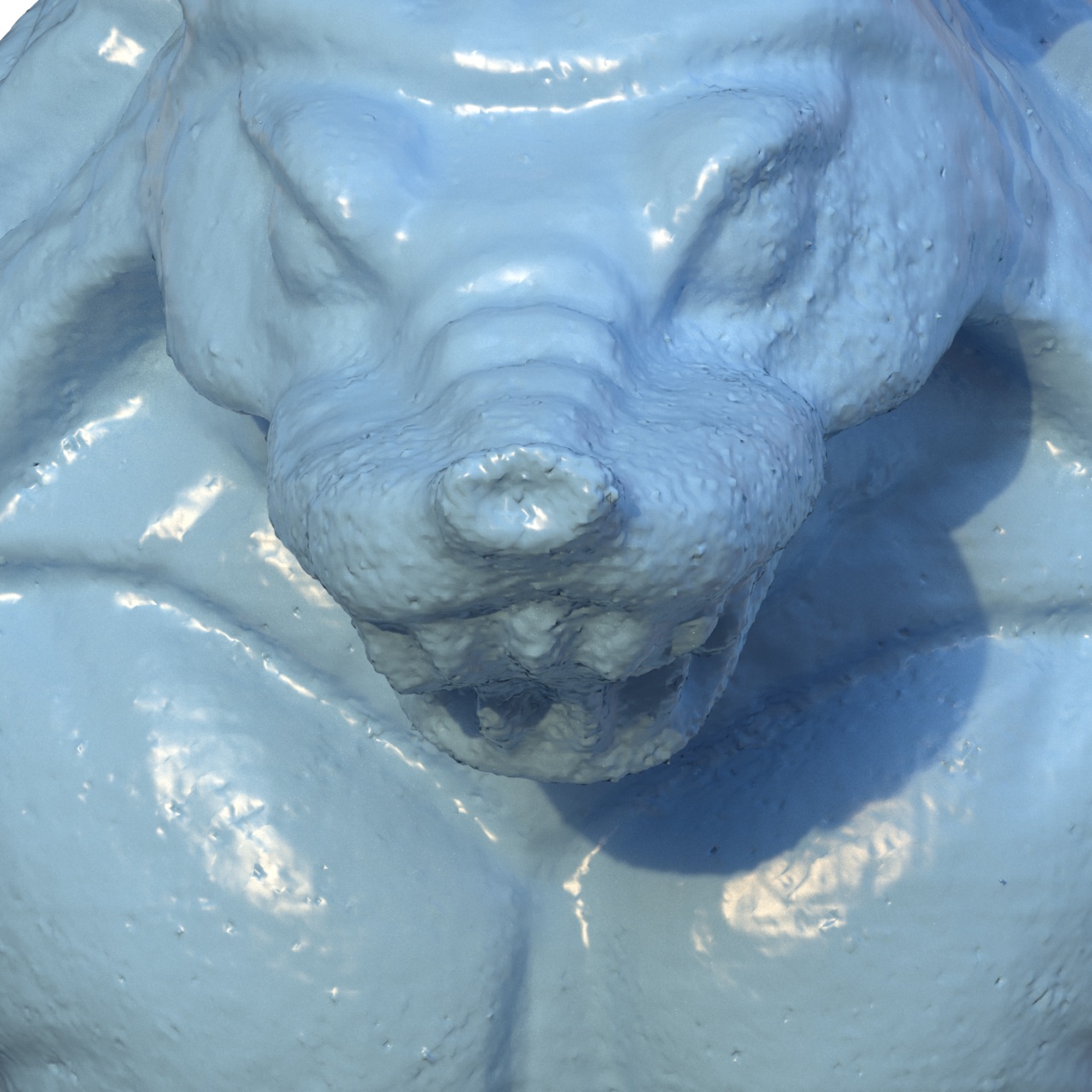}}\\[-0.5mm]\frame{\includegraphics[width=0.12\linewidth]{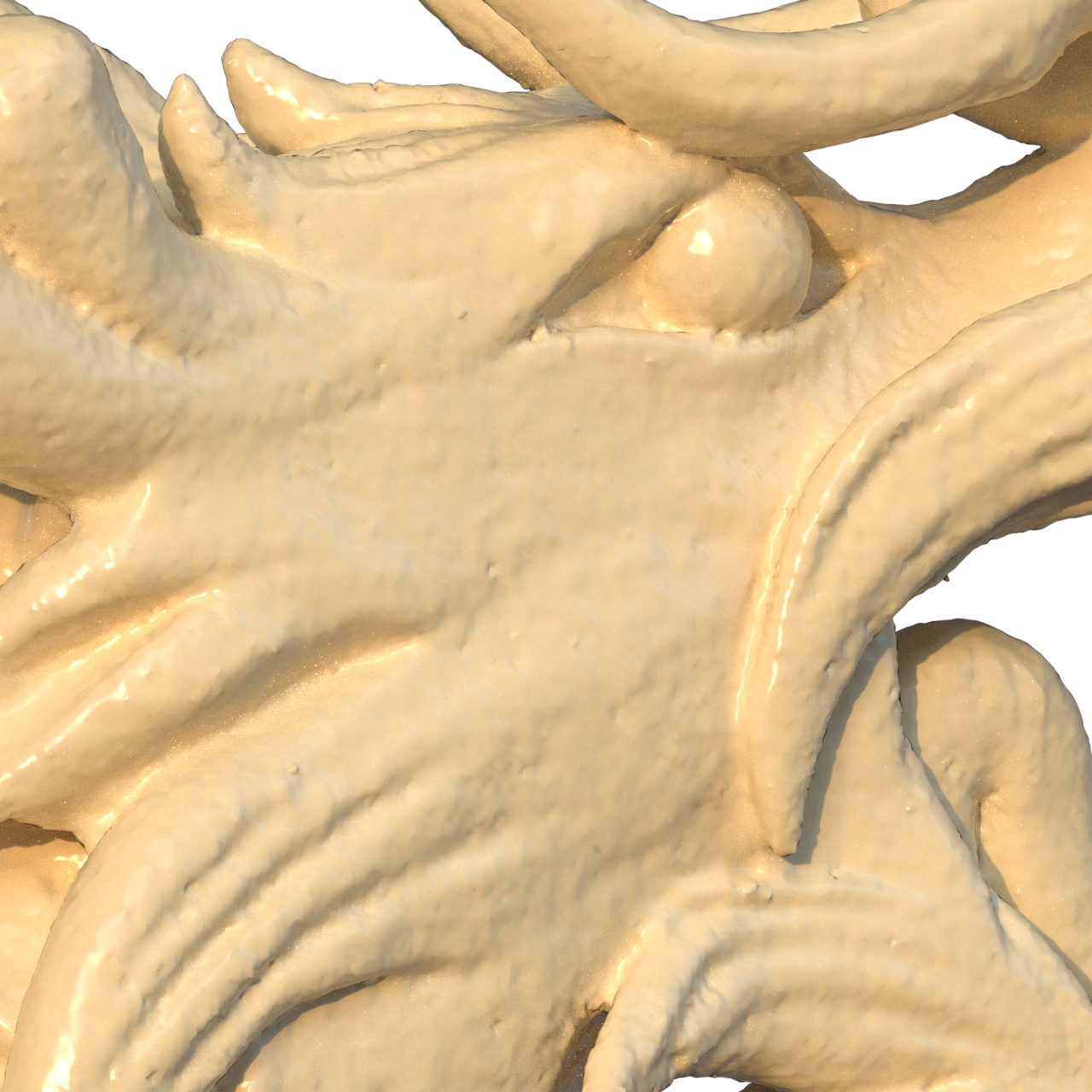}}} &
	\makecell{\frame{\includegraphics[width=0.12\linewidth]{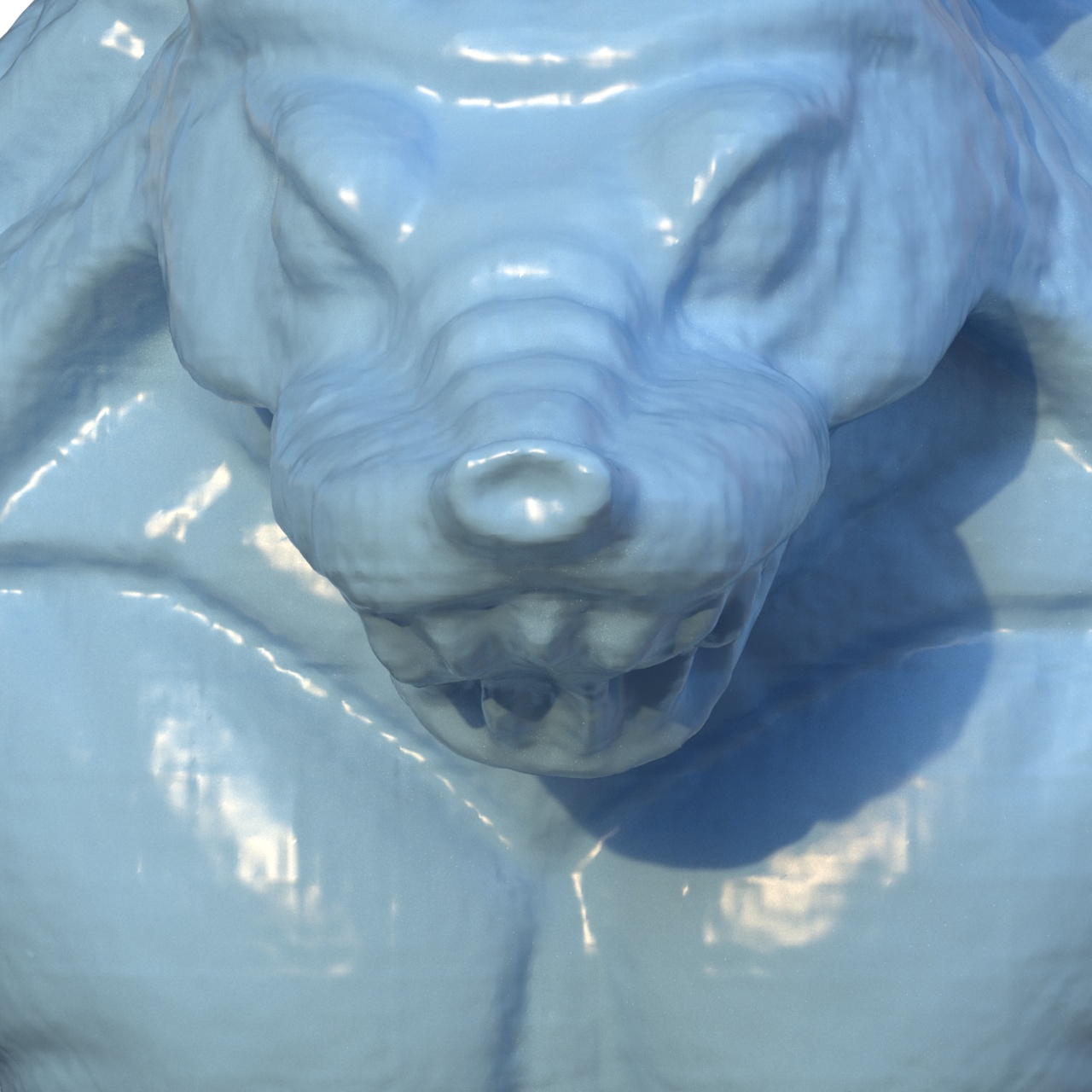}}\\[-0.5mm]\frame{\includegraphics[width=0.12\linewidth]{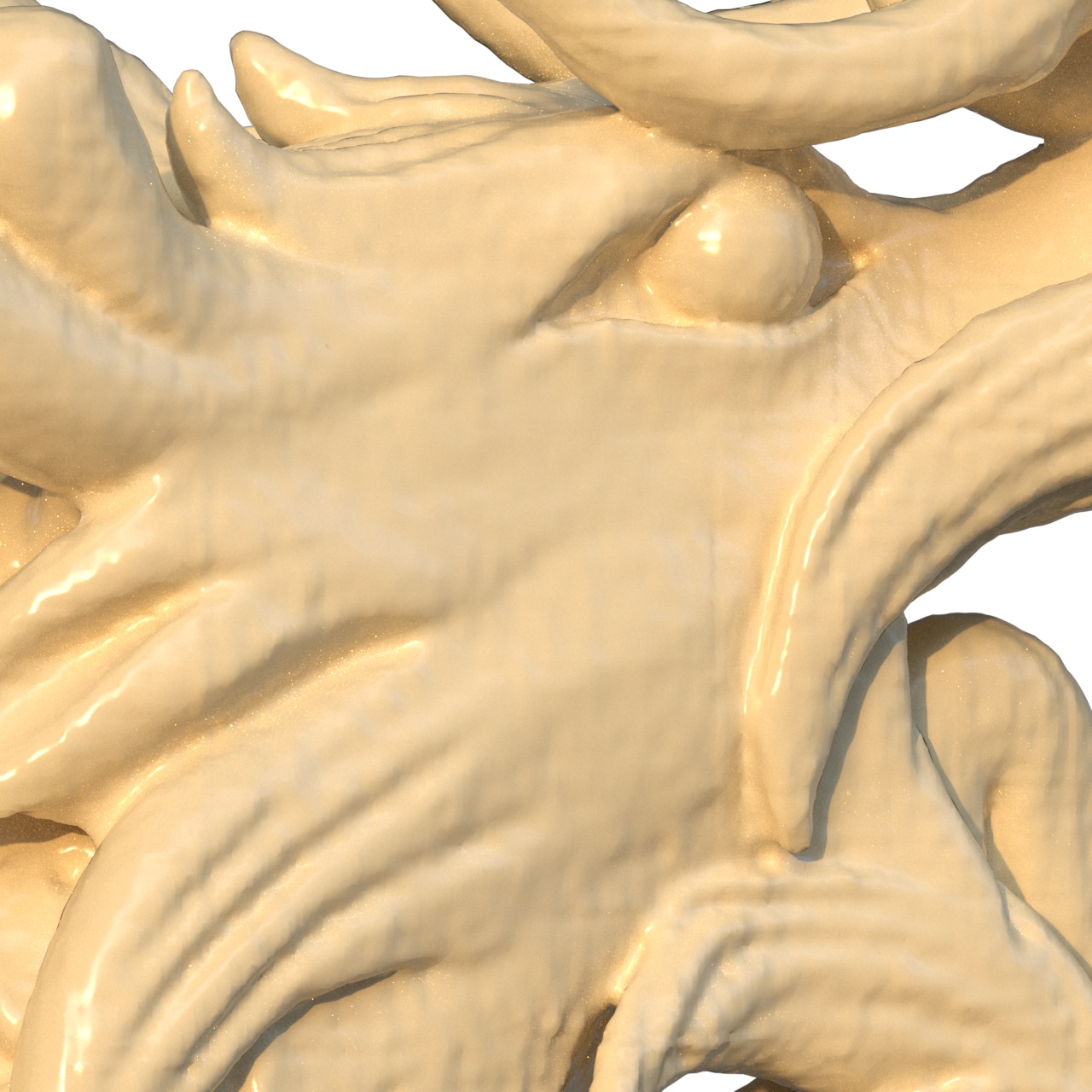}}} &
 	\makecell{\frame{\includegraphics[width=0.12\linewidth]{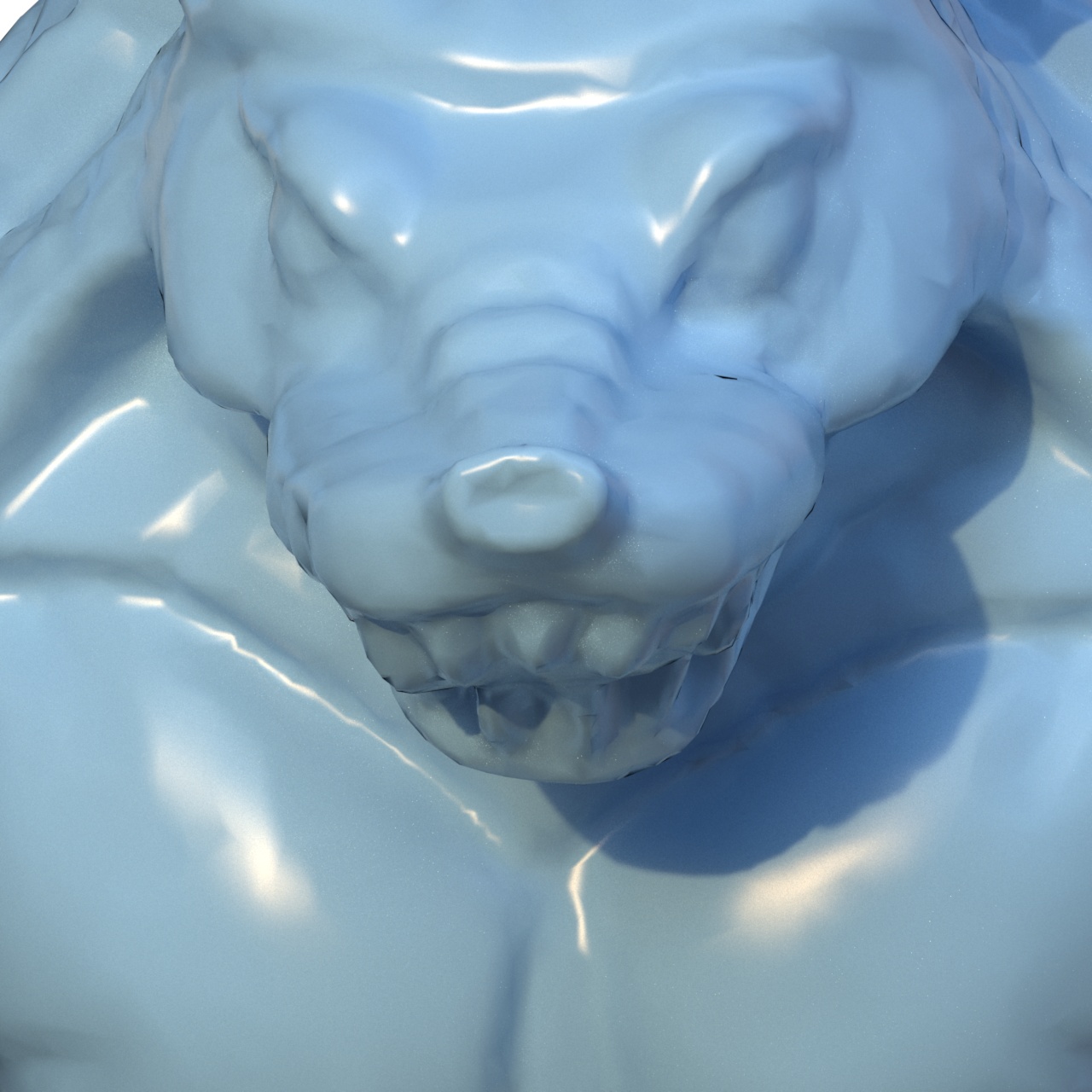}}\\[-0.5mm]\frame{\includegraphics[width=0.12\linewidth]{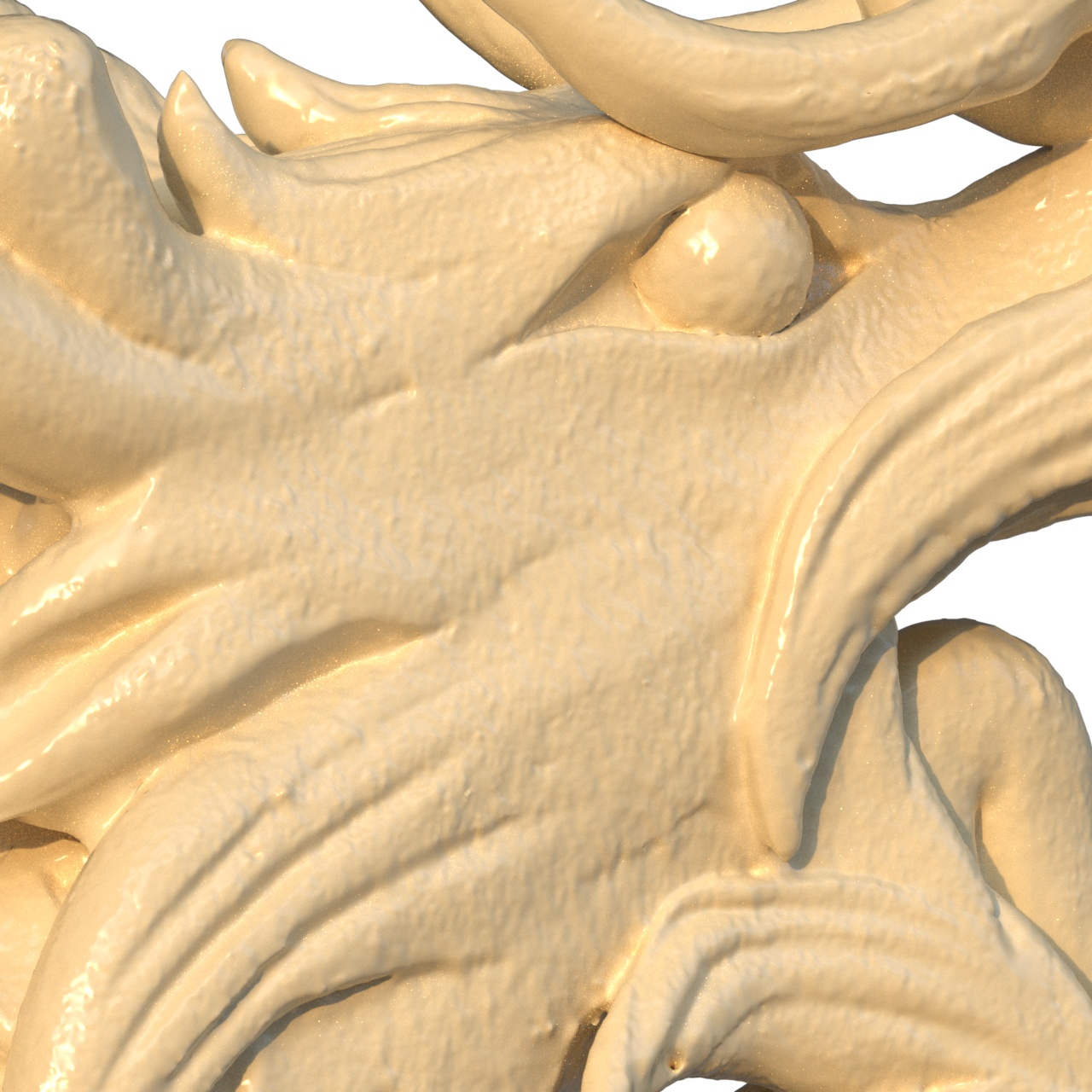}}} &
	\makecell{\vspace{-10mm}\includegraphics[width=0.18\linewidth]{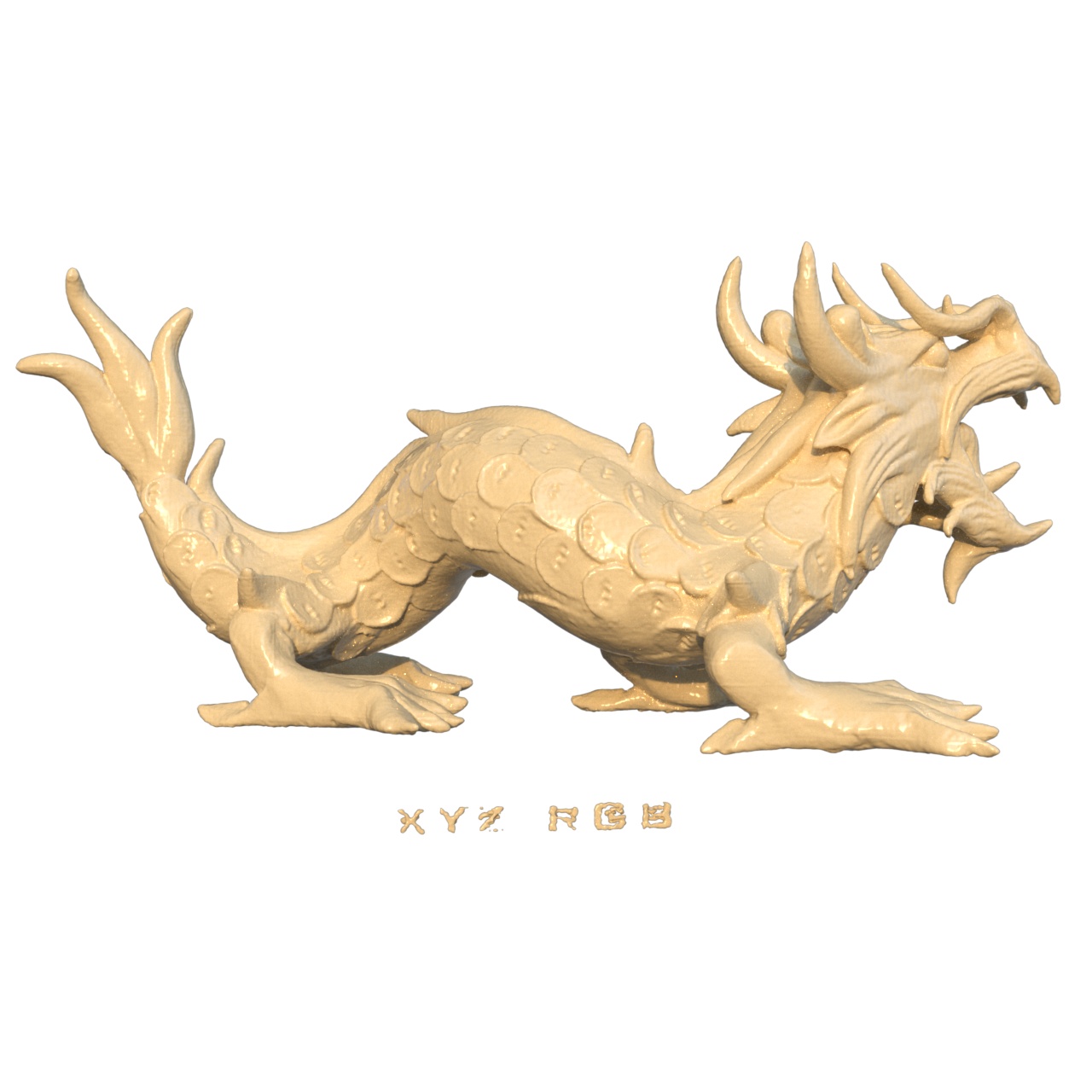}} \\
  
  \makecell{\hspace{2.0mm}Armadillo} & &&& &&\makecell{\hspace{0mm}Dragon}  \\
 [-5.0mm] &

	\footnotesize \SI{1.0}{\mega\nothing} (params) &
	\footnotesize \SI{1.0}{\mega\nothing}&  
	\footnotesize \SI{11.6}{\mega\nothing} &
	\footnotesize \SI{5.10}{\mega\nothing} & model size (\SI{}{\mega\nothing}) & 
	\footnotesize \\[-1.2mm] &
	\footnotesize $36$:$53$ (mm:ss) &
	\footnotesize $29$:$22$ & 
	\footnotesize $0$:$41$ &
	\footnotesize $0$:$31$ & speed (mm:ss) & 
	\footnotesize \\[-1.2mm] &
  	\footnotesize 2.414 &
	\footnotesize 1.561 & 
	\footnotesize 1.442 &
	\footnotesize 1.395 & CD ($1e-3$) &
        \footnotesize \\[-1.2mm] &
	\footnotesize $0.9169$ (gIoU) &
	\footnotesize $0.9663$ & 
	\footnotesize $0.9722$ &
	\footnotesize $0.9795$ & gIoU & 
	\footnotesize \\
  
\end{tabular}

  \vspace{-3mm}
  \caption{
    \textbf{Signed-Distance Field Reconstruction.} We reconstruct SDFs from \SI{8.0}{\mega\nothing} training points. We show qualitative visual comparisons on the top and quantitative comparisons on the bottom including the number of parameters, reconstruction time and gIoU. DiF-Grid and iNGP~\cite{Mueller2022TOG} are trained for $10k$ iterations, while SIREN~\cite{Sitzmann2020NIPS} and NeRF with Frequency Encodings~\cite{Tancik2020NEURIPS} are trained for $200k$ iterations.
  }
\label{fig:sdf_results}%
\end{figure*}

\figref{fig:sdf_results} shows a quantitative and qualitative comparison of all methods.
Our method leads to visually better results, it recovers high-frequency geometric details and contains less noise on smooth surfaces (\eg, the elephant face).
The high visual quality is also reflected by the highest gIoU value of all methods.
Meanwhile, our method also achieves the fastest reconstruction speed, while using less than half of the number of parameters used by CUDA-kernel enabled Instant-NGP, demonstrating the high accuracy, efficiency, and compactness of our factorized representation.


\boldparagraph{Radiance Field Reconstruction}
\label{sec:radiance_field_recon}
Radiance field reconstruction aims to recover the 3D density and radiance of each volume point from as multi-view RGB images. The geometry and appearance properties are updated via inverse volume rendering, as proposed in NeRF \cite{Mildenhall2020ECCV}. Recently, many encoding functions and advanced representations have been proposed that significantly improve reconstruction speed and quality, such as sparse voxel grids \cite{Keil2022CVPR}, hash tables\cite{Mueller2022TOG} and tensor decomposition\cite{Chen2022ECCV}. 

In \tabref{tab:radiance_field_score}, we quantitatively compare DiF-Grid with several state-of-the-art fast radiance field reconstruction methods (Plenoxel \cite{Keil2022CVPR}, DVGO \cite{Sun2022CVPR}, Instant-NGP \cite{Mueller2022TOG} and TensoRF-VM \cite{Chen2022ECCV}) on both synthetic \cite{Mildenhall2020ECCV} as well as real scenes (Tanks and Temple objects) \cite{Knapitsch2017SIGGRAPH}. 
Our method achieves high reconstruction quality, significantly outperforming NeRF, Plenoxels, and DVGO on both datasets, while being significantly more compact than Plenoxels and DVGO.
We also outperform Instant-NGP and are on par with TensoRF regarding reconstruction quality, while being highly compact with only $\SI{5.1}{\mega\nothing}$ parameters, less than one-third of TensoRF-VM and one-half of Instant-NGP. 
Our DiF-Grid also optimizes faster than TensoRF, at slightly over $10$ minutes, in addition to our superior compactness. 
Additionally, unlike Plenoxels and Instant-NGP which rely on their own CUDA framework for fast reconstruction, our implementation uses the standard PyTorch framework, making it easily extendable to other tasks.


In general, our model leads to state-of-the-art results on all three challenging benchmark tasks with both high accuracy and efficiency. 
Note that the baselines are mostly single-factor, utilizing either a local field (such as DVGO and Plenoxels) or a global field (such as Instant-NGP).
In contrast, our DiF model is a two-factor method, incorporating both local coefficient and global basis fields, hence resulting in better reconstruction quality and memory efficiency.

\setlength{\tabcolsep}{8pt}
\begin{table*}[t]
\centering

\begin{tabular}{lccllllll}
& &&\multicolumn{4}{c}{Synthetic-NeRF} & \multicolumn{2}{c}{\revised{Tanks and Temples}} \\
\cmidrule(lr){4-7} \cmidrule(lr){8-9} 
Method & BatchSize & Steps & Time $\downarrow$  & Size (\SI{}{\mega\nothing})$\downarrow$ & PSNR$\uparrow$ & SSIM$\uparrow$ & PSNR$\uparrow$ & SSIM$\uparrow$ \\
\hline

    NeRF~\cite{Mildenhall2020ECCV}      & 4096      & 300k  & $\sim$35h      & 01.25	\gold      & 31.01          & 0.947         & 25.78         & 0.864 \\
    Plenoxels~\cite{Keil2022CVPR}        & 5000      & 128k  & 11.4m \silve   & 194.5           & 31.71          & 0.958         & 27.43         & 0.906 \\
    DVGO~\cite{Sun2022CVPR}             & 5000      & 30k   & 15.0m          & 153.0           & 31.95 \bronze          & 0.957         & 28.41 \bronze & 0.911 \bronze \\
    Instant-NGP~\cite{Mueller2022TOG}   & 10k-85k   & 30k   & 03.9m \gold     & 11.64 \bronze   & 32.59 \silve  & 0.960 \bronze  & 27.09         & 0.905 \\
    TensoRF-VM~\cite{Chen2022ECCV}      & 4096      & 30k   & 17.4m          & 17.95           & 33.14 \gold    & 0.963 \gold   &  28.56 \silve & 0.920 \silve \\
    \hline
    \revised{DiF-Grid} (Ours)                 & 4096      & 30k   & 12.2m \bronze  & 05.10 \silve     & 33.14 \gold   & 0.961 \silve  & 29.00 \gold   & 0.938 \gold \\
\bottomrule
\end{tabular}
\vspace{1mm}
\caption{\textbf{Novel View Synthesis with Radiance Fields.} We compare our method to previous radiance field reconstruction methods on the Synthetic-NeRF\cite{Mildenhall2020ECCV} and Tanks and Temples \cite{Knapitsch2017SIGGRAPH} datasets. We report the scores reported in the original papers whenever available.  We also show average reconstruction time and model size for the Synthetic-NeRF dataset to compare the efficiency of the methods.}
\label{tab:radiance_field_score} 
\end{table*}
\setlength{\tabcolsep}{1.4pt}






\begin{figure}[t]
\begin{center}
    \includegraphics[width=0.95\linewidth]{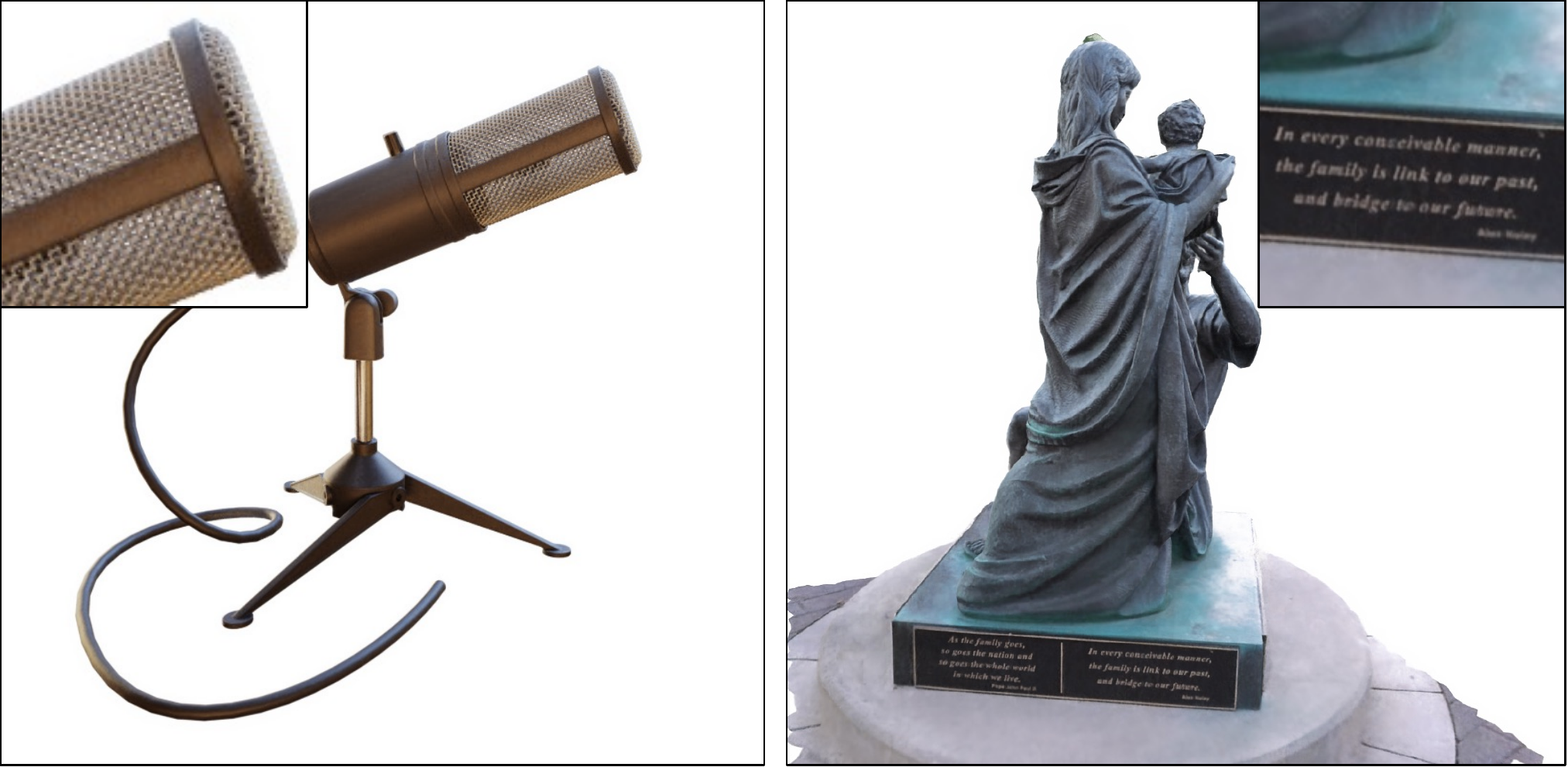}
\end{center}
\vspace{-3mm}
  \caption{\revised{\textbf{Radiance Field Reconstruction}. We evaluate our DiF using NeRF-Synthetic and Tanks and Temples datasets, our method is able to reconstruct high-quality surface details.}}
\label{fig:abalations_RF}
\end{figure}

\subsection{Generalization}
\label{sec:experiments_generalization}
Recent advanced neural representations such as NeRF, SIREN, ACORN, Plenoxels, Instant-NGP and TensoRF optimize each signal separately, lacking the ability to model multiple signals jointly or learning useful priors from multiple signals.
In contrast, our DiF representation not only enables accurate and efficient per-signal reconstruction (as demonstrated in \secref{sec:per-signal}) but it can also be applied to generalize across signals by simply sharing the basis field across signal instances.
We evaluate the benefits of basis sharing by conducting experiments on image regression from partial pixel observations and few-shot radiance field reconstruction.
For these experiments, instead of DiF-Grid, we adopt DiF-MLP-B (i.e., $(5)$ in the \tabref{tab:field_representation}) as our DiF representation, where we utilize a tensor grid to model the coefficient and $6$ tiny MLPs (two layers with $32$ neurons each) to model the basis.
We find that DiF-MLP-B performs better than DiF-Grid in the generalization setting, owing to the strong inductive smoothness bias of MLPs.

\boldparagraph{Image Regression from Sparse Observations}
Unlike the image regression experiments conducted in Sec.~\ref{sec:per-signal} which use all image pixels as observations during optimization, this experiment focuses on the scenario where only part of the pixels are used during optimization.
Without additional priors, a single-signal optimization easily overfits in this setting due to the sparse observations and the limited inductive bias, hence failing to recover the unseen pixels.

We use our DiF-MLP-B model to learn data priors by pre-training it on $800$ facial images from the FFHQ dataset\cite{Karras2018ARXIV} while sharing the MLP basis and projection function parameters.
The final image reconstruction task is conducted by optimizing the coefficient grids for each new test image.

\begin{figure}
  \small\sffamily
  \hspace{-1mm}
  
\setlength{\tabcolsep}{1pt}%
\renewcommand{\arraystretch}{1.1}%
\begin{tabular}{cccc}
	\makecell{\includegraphics[height=0.245\linewidth]{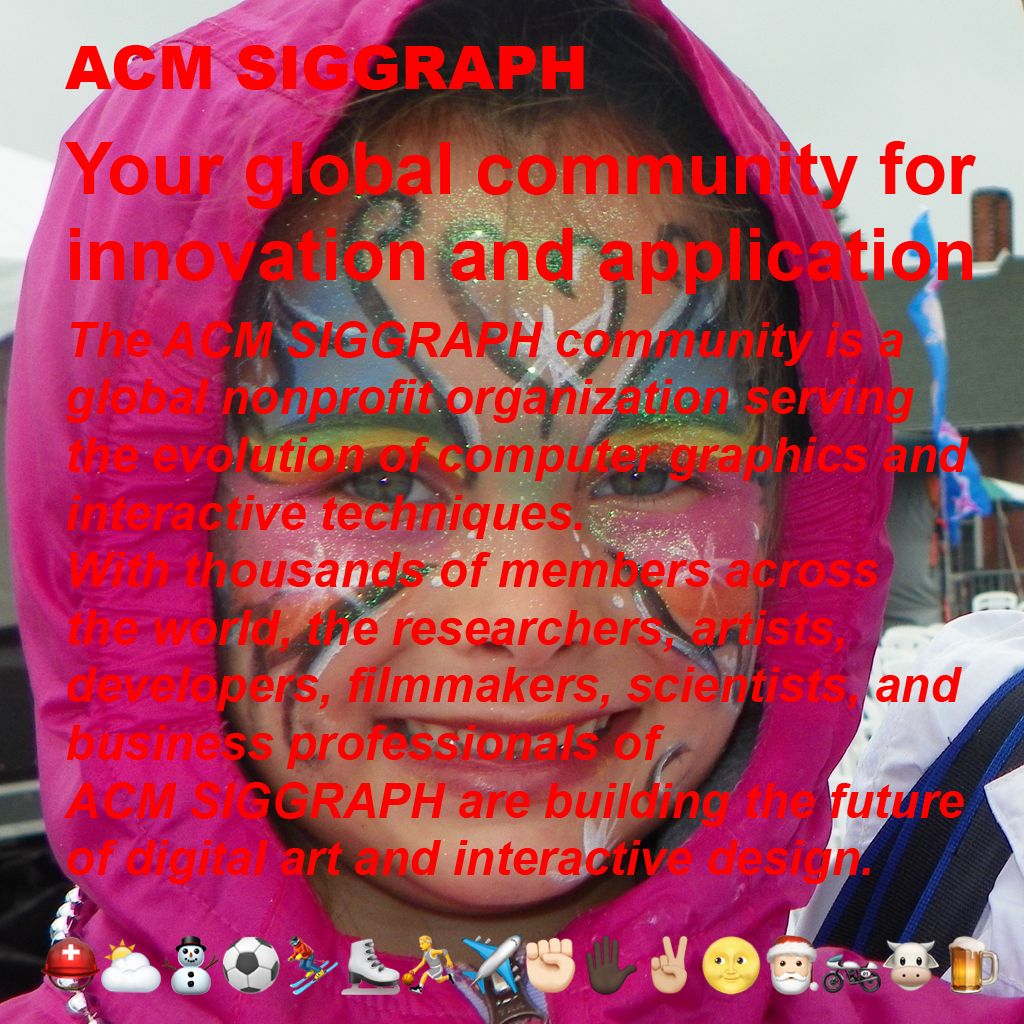}} &
	\makecell{\includegraphics[height=0.245\linewidth]{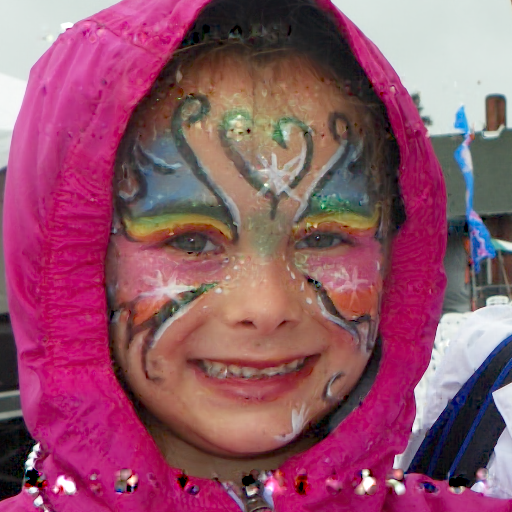}} &
	\makecell{\includegraphics[height=0.245\linewidth]{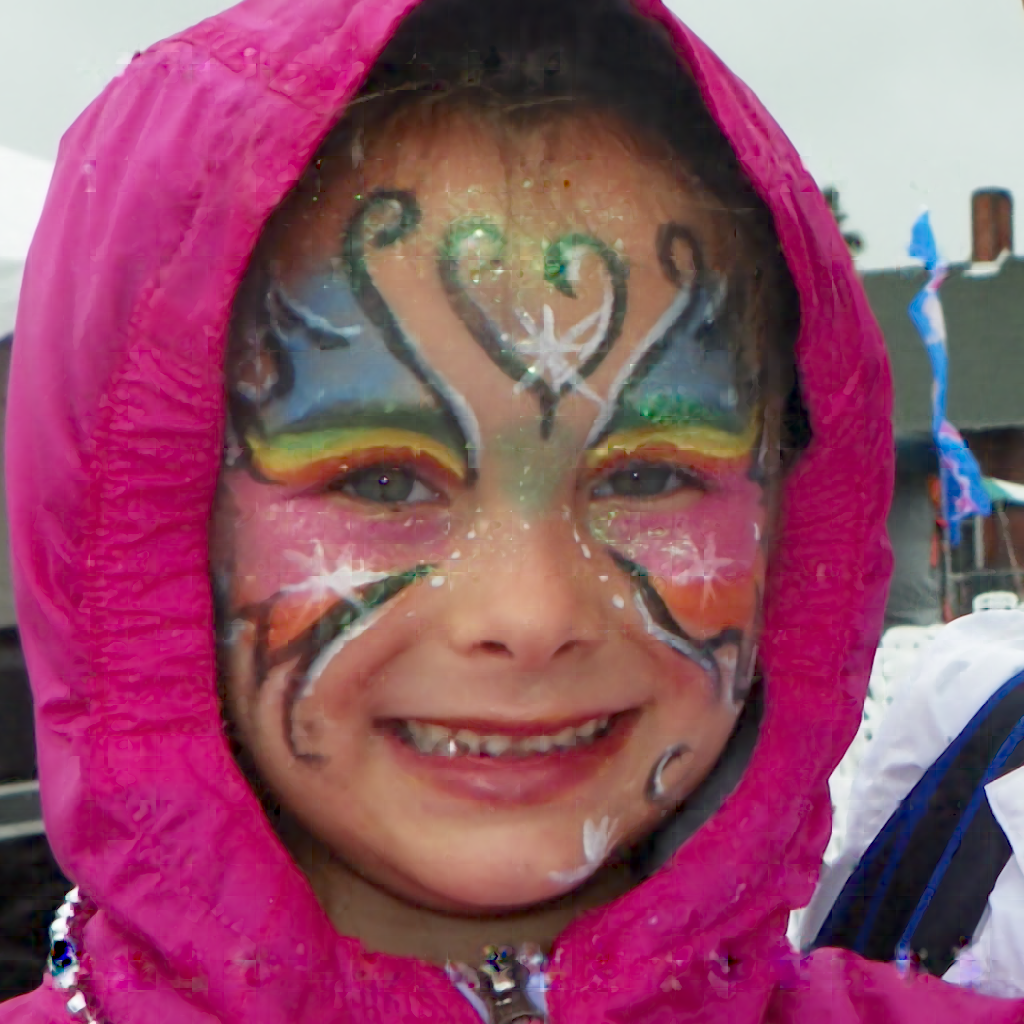}} &
	\makecell{\includegraphics[height=0.245\linewidth]{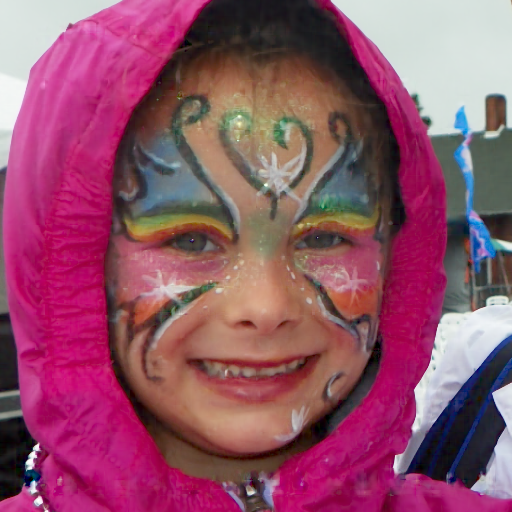}} \\[-1.2mm]
	\makecell{\includegraphics[height=0.245\linewidth]{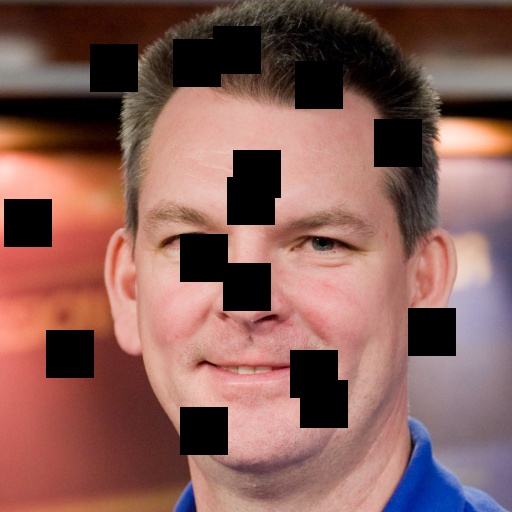}} &
	\makecell{\includegraphics[height=0.245\linewidth]{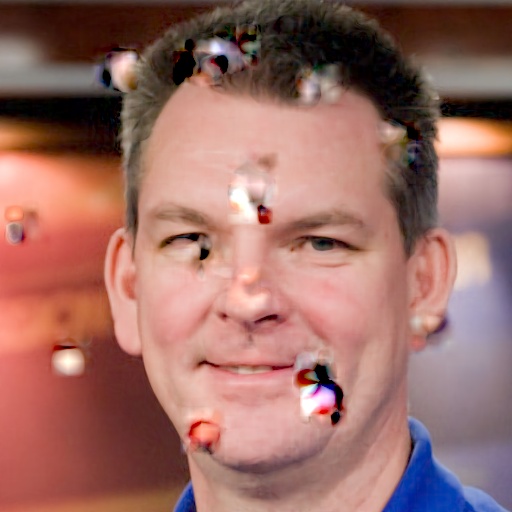}} &
	\makecell{\includegraphics[height=0.245\linewidth]{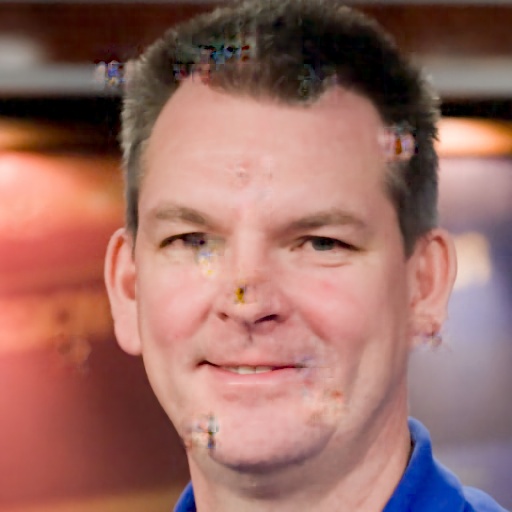}} &
	\makecell{\includegraphics[height=0.245\linewidth]{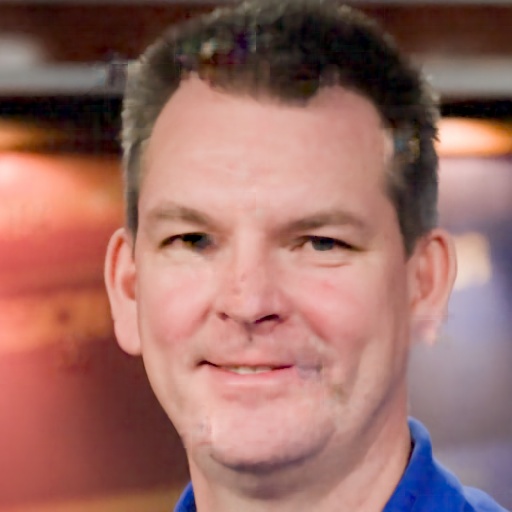}} \\[-1.2mm]
 	\makecell{\includegraphics[height=0.245\linewidth]{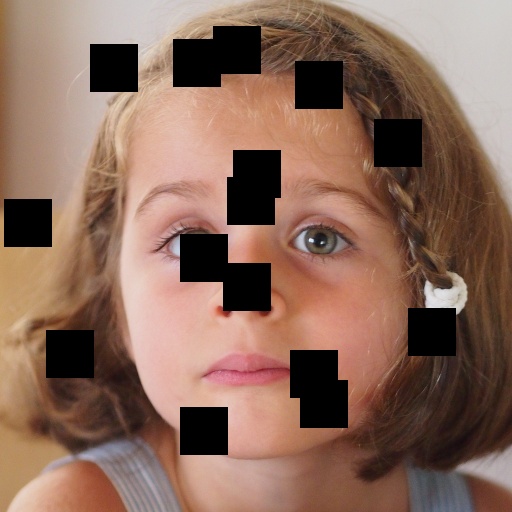}} &
	\makecell{\includegraphics[height=0.245\linewidth]{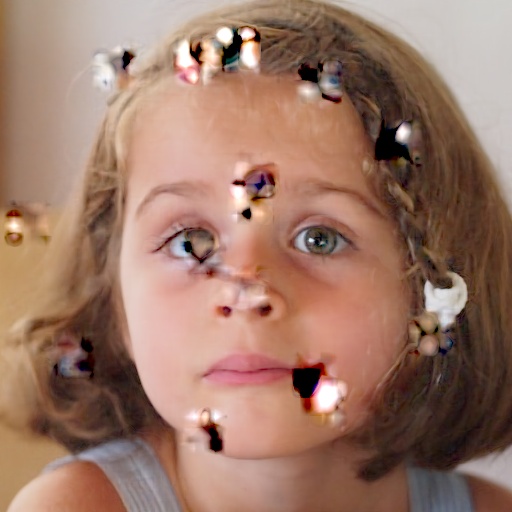}} &
	\makecell{\includegraphics[height=0.245\linewidth]{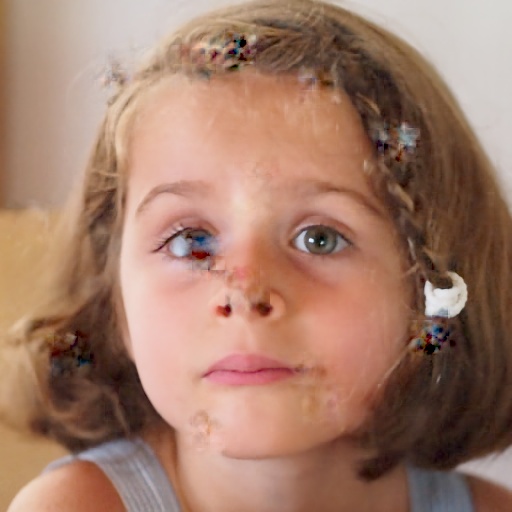}} &
	\makecell{\includegraphics[height=0.245\linewidth]{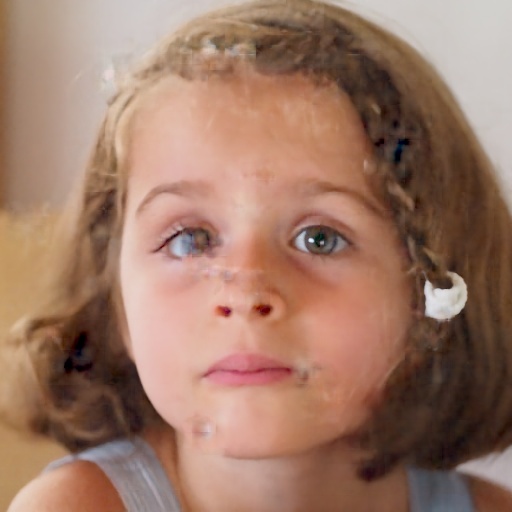}} \\[-1.2mm]
	\scriptsize Input&
	\scriptsize Instant-NGP&
	\scriptsize DiF-MLP-B&
	\scriptsize DiF-MLP-B$^*$
\end{tabular}

  \vspace{-4mm}
  \caption{%
    \textbf{Image Regression from Sparse Observations.} Results obtained by fitting each model to all unmasked pixels. We use randomly placed black squares as masks for the bottom two rows and an image of text and small icons as mask for the top row. The symbol $^*$ denotes pre-training of the basis factors using the FFHQ facial image set. Our pre-trained model (DiF-MLP-B$^*$) learns robust basis fields which lead to better reconstruction compared to the per-scene baselines Instant-NGP and DiF-MLP-B. 
  }
  \label{fig:img_inpainting}
\end{figure}

In \figref{fig:img_inpainting}, we show the image regression results on three different facial images with various masks and compare them to baseline methods that do not use any data priors, including Instant-NGP and our DiF-MLP-B without pre-training.
As expected, Instant-NGP can accurately approximate the training pixels but results in random noise in the untrained mask regions.
Interestingly, even without pre-training and priors from other images, our DiF-MLP-B is able to capture structural information to some extent within the same image being optimized; as shown in the eye region, the model can learn the pupil shape from the right eye and regress the left eye (masked during training) by reusing the learned structures in the shared basis functions. 
%
As shown on the right of \figref{fig:img_inpainting}, 
our DiF-MLP-B with pre-trained prior clearly achieves the best reconstruction quality with better structures and boundary smoothness compared to the baselines, demonstrating that our factorized DiF model allows for learning and transferring useful prior information from the training set.

\boldparagraph{Few-Shot Radiance Field Reconstruction}
\label{sec:few_shot}
Reconstructing radiance fields from few-shot input images with sparse viewpoints is highly challenging. 
Previous works address this by imposing sparsity assumptions \cite{Niemeyer2022CVPR, Kim2022} in per-scene optimization or training feed-forward networks \cite{Yu2021CVPR, Chen2021ICCV, Kulhanek2022ECCV} from datasets.
Here we consider $3$ and $5$ input views per scene and seek a novel solution that leverages data priors in pre-trained basis fields of our DiF model during the optimization task. It is worth-noting that the views are chosen in a quarter sphere, thus the overlapping region between views is quite limited.

\begin{figure}[t]
\begin{center}

\renewcommand{\arraystretch}{1.1}%
\begin{tabular}{cc}

    \makecell{\rotatebox{90}{\footnotesize MVSNeRF}} & \makecell{\includegraphics[width=0.95\linewidth]{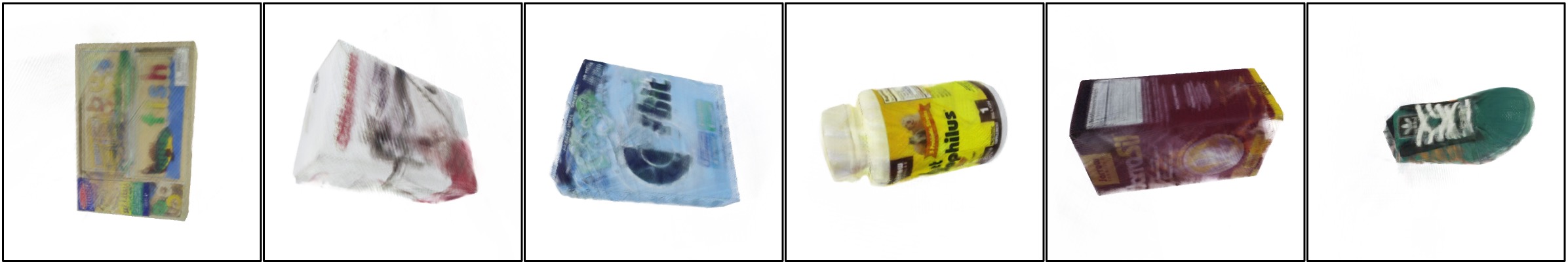}} \\
    \makecell{\rotatebox{90}{\footnotesize PixelNeRF}} & \makecell{\includegraphics[width=0.95\linewidth]{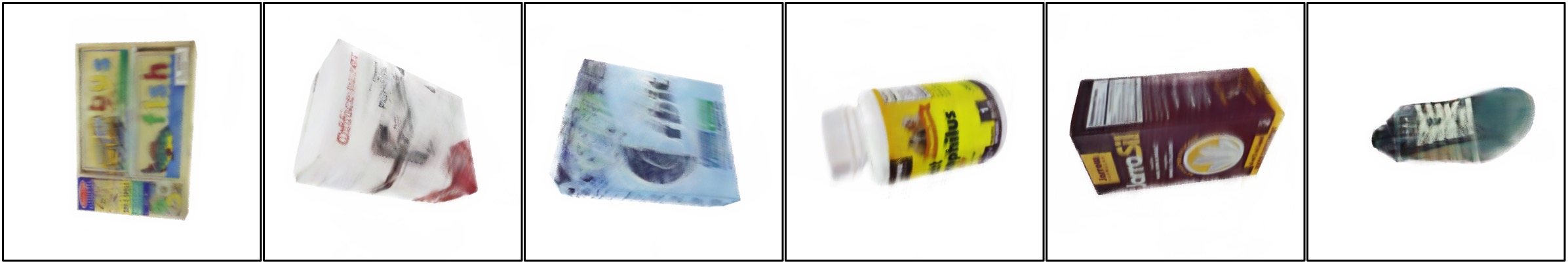}} \\
    \makecell{\rotatebox{90}{\footnotesize iNGP}} & \makecell{\includegraphics[width=0.95\linewidth]{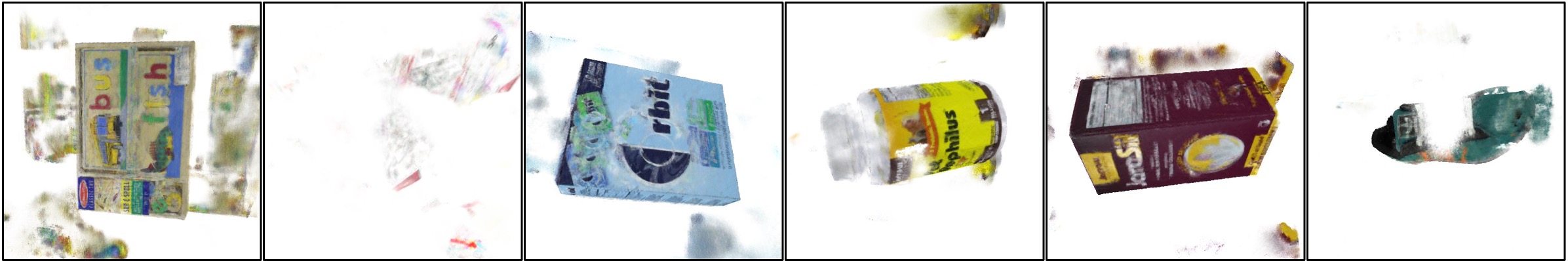}} \\
    \makecell{\rotatebox{90}{\footnotesize w/o prior}} & \makecell{\includegraphics[width=0.95\linewidth]{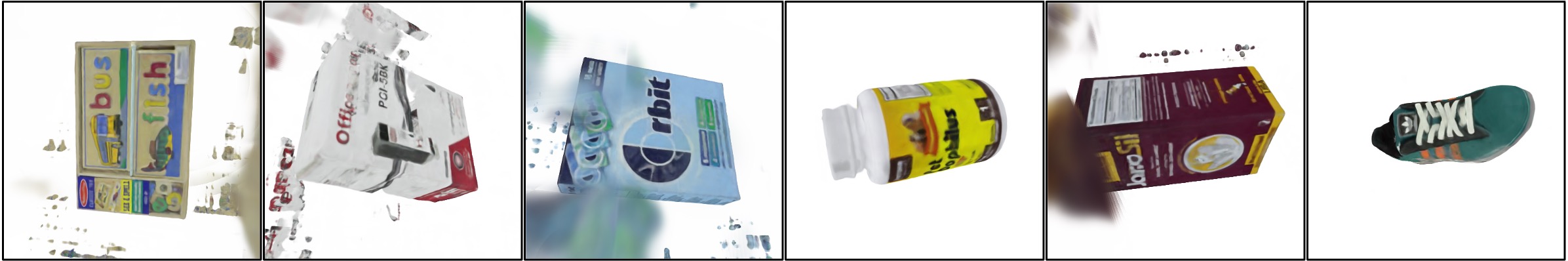}} \\
    \makecell{\rotatebox{90}{\footnotesize w/ prior}} & \makecell{\includegraphics[width=0.95\linewidth]{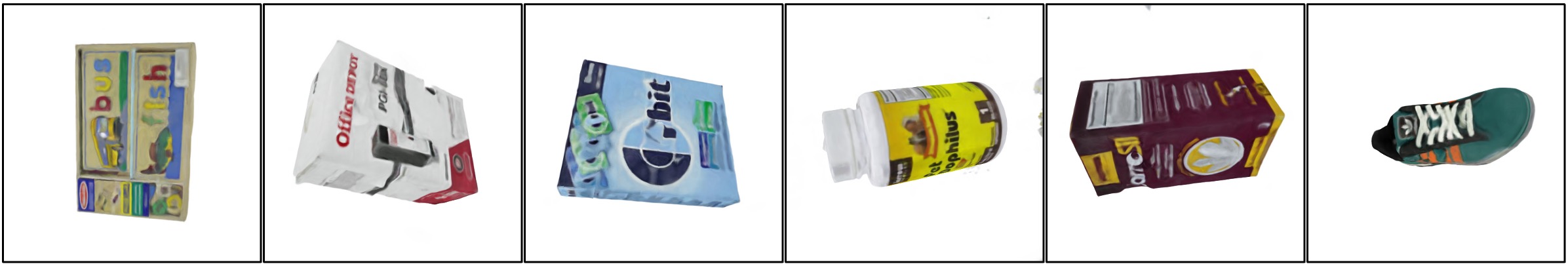}} \\
    \makecell{\rotatebox{90}{\footnotesize GT}} & \makecell{\includegraphics[width=0.95\linewidth]{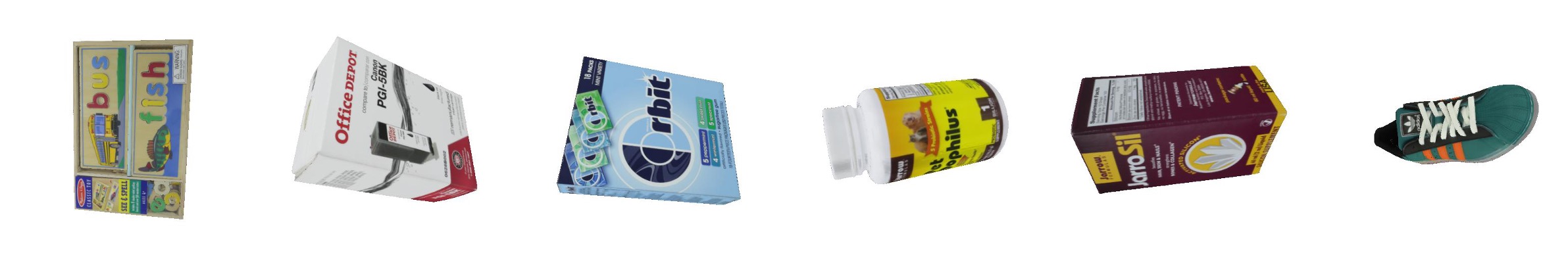}} \\
	
\end{tabular}



\end{center}
  \caption{\textbf{Radiance Fields from 5 Views.} We visualize novel view synthesis results of six test scenes, corresponding to the quantitative results in Tab.~\ref{table:few_shot_nerf}. We show our DiF-MLP-B model w/ and w/o pre-trained data priors (bottom two rows) and compare it to Instant-NGP, PixelNeRF and MVSNeRF (top three rows). Our model with pre-trained basis factors can effectively utilize the learned data priors, resulting in superior qualitative results with fewer outliers compared to single-scene models (iNGP and ours w/o priors), as well as sharper details compared to feed-forward models (PixelNeRF and MVSNeRF).}
\label{fig:few_shot}
\end{figure}

Specifically, we first train DiF models on $100$ Google Scanned Object scenes~\cite{downs2022google}, which contains $250$ views per scene. During cross-scene training, we maintain $100$ per-scene coefficients and share the basis $\bc$ and projection function $\cP$. After cross-scene training, we use the mean coefficient values of pre-trained coefficient fields as the initialization, while fixing the pre-trained functions ($\bc$ and $\cP$) and fine-tuning the coefficient field for new scenes with few-shot observations.
In this experiment, we compare results from both DiF-MLP-B and DiF-Grid with and without the pre-training.
We also compare with Instant-NGP and previous few-shot reconstruction methods, including PixelNeRF \cite{Yu2021CVPR} and MVSNeRF \cite{Chen2021ICCVb}, re-train with the same training set and test using the same 3 or 5 views.  
As shown in \tabref{table:few_shot_nerf} and \figref{fig:few_shot}, our pre-trained DiF representation with MLP basis provides strong regularization for few-shot reconstruction, resulting in fewer artifacts and better reconstruction quality than the single-scene optimization methods without data priors and previous few-shot reconstruction methods that also use pre-trained networks.
In particular, without any data priors, single-scene optimization methods (Instant-NGP and ours w/o prior) lead to a lot of outliers due to overfitting to the few-shot input images.
Previous methods like MVSNeRF and PixelNeRF achieve plausible reconstructions due to their learned feed-forward prediction which avoids per-scene optimization. However, they suffer from blurry artifacts.
Additionally, the strategy taken by PixelNeRF and MVSNeRF assumes a narrow baseline and learns correspondences across views for generalization via feature averaging or cost volume modeling which does not work as effectively in a wide baseline setup. On the other hand, by pre-training shared basis fields on multiple signals, our DiF model can learn useful data priors, enabling the reconstruction of novel signals from sparse observations via optimization.

\begin{table}[t]
\centering
\setlength{\tabcolsep}{3.0pt}
\begin{tabularx}{\linewidth}{llllll}
&  & \multicolumn{2}{c}{3 views} & \multicolumn{2}{c}{5 views} \\
\cmidrule(lr){3-4} \cmidrule(lr){5-6} 
Method & Time$\downarrow$ & PSNR$\uparrow$ & SSIM$\uparrow$ & PSNR$\uparrow$ & SSIM$\uparrow$  \\
\hline

iNGP         & 03:38 \silve & 14.74 & 0.776 & 20.79 & 0.860  \\
\revised{DiF-Grid}  & 13:39  & 18.13 & 0.805 & 20.83 & 0.847  \\
\revised{DiF-MLP-B} & 18:24 & 16.31 & 0.804 & 22.26 & 0.900 \bronze  \\
\hline
PixelNeRF   & 00:00 \gold & 21.37 \bronze  & 0.878 \bronze & 22.73  & 0.896 \\
MVSNeRF   & 00:00 \gold & 20.50  & 0.868 & 22.76 & 0.891  \\
\revised{PixelNeRF-ft}   & \revised{25:18} & \revised{22.21} \gold  & \revised{0.882} \silve & \revised{23.67} \bronze  & \revised{0.895} \\
\revised{MVSNeRF-ft}   & \revised{13:06} \bronze & \revised{18.51}  & \revised{0.864} & \revised{20.49} & \revised{0.887}  \\
\revised{DiF-Grid}$^*$  & 13:18 \silve & 20.77 & 0.871 & 25.41 \silve & 0.915 \silve \\
\revised{DiF-MLP-B}$^*$  & 18:44 & 21.96 \silve & 0.891 \gold & 26.91 \gold & 0.927 \gold  \\%

\bottomrule
\end{tabularx}
\caption{\textbf{Few-shot Radiance Field Reconstruction.} We show quantitative comparisons of few-shot radiance field reconstruction from 3 or 5 viewpoints regarding optimization time and novel view synthesis quality (PSNRs and SSIMs).  Results are averaged across 8 test scenes. The results of Instant-NGP and our DiF models are generated based on per-scene optimization, while DiF models with $^*$ use pre-trained basis factors across scenes. We train the feed-forward networks of PixelNeRF and MVSNeRF using the same dataset we learn our shared basis factors, and the results of PixelNeRF and MVSNeRF are generated from the networks via direct feed-forward inference. Our DiF-MLP-B$^*$ with pre-trained MLP basis factors leads to the best reconstruction quality. }
\label{table:few_shot_nerf}  
\end{table}
\setlength{\tabcolsep}{1.4pt}

\subsection{Influence of Design Choices in Factor Fields}
\label{sec:designs}
Factor Fields is a general framework that unifies many previous representations with different settings. In this section, we aim to analyze the properties of these variations and offer a comprehensive understanding of the components of the proposed representation framework. We conduct extensive evaluations on the four main components of our Factor Fields framework: factor number $N$, level number $L$, coordinate transformation function $\bgamma_i$, field representation $\bff_i$, and field connector $\circ$. 

We present a comprehensive assessment of the representations' capabilities in terms of efficiency, compactness, reconstruction quality, as well as generalizability, with a range of tasks including 2D image regression (with all pixels), and per-scene and across-scene 3D radiance field reconstruction. Note that, the settings in per-scene and across-scene radiance field reconstruction are the same as introduced in \secref{sec:radiance_field_recon} and \secref{sec:few_shot},  while for the 2D image regression task, we use the same model setting as in \secref{sec:radiance_field_recon} and test on $256$ high fidelity images at a resolution of $1024 \times 1024$ from the DIV2K dataset \cite{Agustsson2017CVPRW}. 
To enable meaningful comparisons, we evaluate the variations within the same code base and report their performance using the same number of iterations number, batch size, training point sampling strategy and pyramid frequencies.
Correspondingly, the results for Instant-NGP, EG3D, OccNet, NeRF, DVGO, TensoRF-VM/-CP in Tab.~\ref{tab:ablation} are based on our reimplementation of the original methods in our unified Factor Fields framework with the corresponding design parameters shown in the tables.

\boldparagraph{Factor Number $N$} As illustrated in \eqnref{eq:expansion_4}, the factor number $N$ refers to the number of factors used to represent the target signal.
We show comparisons between single-factor models (1,3,5) and two-factor models (2,4,6) in \tabref{tab:factor_number}.
%
Compared to the models (1) iNGP, (3) EG3D and (5) DiF-no-C which only use a single factor, the models (2) DiF-Hash-B, (4) TensoRF-VM and (6) DiF-Grid use the same factor as (1), (3), (5) respectively, but are extended to two-factor models with an additional factor, leading to 3dB and 0.35 dB PSNR improvement in image regression and 3D radiance field reconstruction tasks, while also increasing training time ($\sim 10\%$) and model size ($\sim 5\%$) as expected.
Despite the marginal computational overhead, introducing additional factors in the framework clearly leads to better reconstruction quality and represents the signal more effectively. The multiplication between factors allows the two factor fields to modulate each other's feature encoding and represent the entire signal more flexibly in a joint manner, alleviating the problem of feature collision in Instant-NGP and related problems of other single-factor models.
%
Additionally, as shown in \tabref{tab:radiance_field_score}, multi-factor modeling (e.g., $N>=2$) is able to  provide more compact modeling while maintaining a similar level of reconstruction quality. Moreover, it allows for generalization by partially sharing the fields across instances, such as the across-scene radiance field modeling in the few-shot radiance reconstruction task. 

\boldparagraph{Level Number $L$}
Our DiF model adopts multiple levels of transformations to achieve pyramid basis fields, similar to the usage of a set of sinusoidal positional encoding functions in NeRF~\cite{Mildenhall2020ECCV}.
We compare multi-level models (including DiF and NeRF) with their reduced single-level versions that only use a single transformation level in \tabref{tab:level_number}.
Note that Occupancy Networks (OccNet, row (1)) do not leverage positional encodings and can be seen as a single-level version of NeRF (row (2)) while the model with multi-level sinusoidal encoding functions (NeRF) leads to about 10dB PSNR performance boost for both 2D image and 3D reconstruction tasks. 
%
On the other hand, the single-level DiF models are also consistently worse than the corresponding multi-level models in terms of speed and reconstruction quality, despite the performance drops being not as severe as those in purely MLP-based representations.

\boldparagraph{Coordinate Transformation $\bgamma_i$}
\label{sec:transformation_function}
In \tabref{tab:coordiante_transformation}, we evaluate four coordinate transformation functions using our DiF representation. These transformation functions include sinusoidal, triangular, hashing and sawtooth. Their transformation curves are shown in \figref{fig:trans_func}. 
In general, in contrast to the random hashing function, the periodic transformation functions (2, 3, 4) allow for spatially coherent information sharing through repeated patterns, where neighboring points can share spatially adjacent features in the basis fields, hence preserving local connectivity. 
We observe that the periodic basis achieves clearly better performance in modeling dense signals (\eg, 2D images). 
For sparse signals such as 3D radiance fields, all four transformation functions achieve high reconstruction quality on par with previous state-of-the-art fast radiance field reconstruction approaches \cite{Sun2022CVPR, Mueller2022TOG, Chen2022ECCV}.

\boldparagraph{Field Representation $\bff_i$}
In \tabref{tab:field_representation}, we compare various functions for representing the factors in our framework (especially our DiF model) including MLPs, Vectors, 2D Maps and 3D Grids, encompassing most previous representations. 
%
Note that discrete feature grid functions (3D Grids, 2D Maps, and Vectors) generally lead to faster reconstruction than MLP functions (e.g. DiF-Grid is faster than DiF-MLP-Band DiF-MLP-C).
While all variants can lead to reasonable reconstruction quality for single-signal optimization, our DiF-Grid representation that uses grids for both factors achieves the best performance on the image regression and single-scene radiance field reconstruction tasks. 
On the other hand, the task of few-shot radiance field reconstruction benefits from basis functions that impose stronger regularization. Therefore, representations with stronger inductive biases (\eg, the Vectors in TensoRF-VM and MLPs in DiF-MLP-B) lead to better reconstruction quality compared to other variants.



\boldparagraph{Field Connector $\circ$} Another key design choice of our Factor Fields framework and DiF model is to adopt the element-wise product to connect multiple factors. Directly concatenating features from different components is an alternative choice and exercised in several previous works \cite{Mildenhall2020ECCV, Chan2022CVPR, Mueller2022TOG}. 
In \tabref{tab:concatenate}, we compare the performance of the element-wised product against the direct concatenation in three model variants. 
Note that the element-wise product consistently outperforms the concatenation operation in terms of reconstruction quality for all models on all applications, demonstrating the effectiveness of using the proposed product-based factorization framework.

\section{Conclusion and Future Work}
In this work, we present a novel unified framework for (neural) signal representations which factorizes a signal as a product of multiple factor fields.
We demonstrate that Factor Fields generalizes many previous neural field representations (like NeRF, Instant-NGP, DVGO, TensoRF) and enables new representation designs.
In particular, we propose a novel representation -- DiF -- with Dictionary factorization, as a new model in the Factor Fields family, which factorizes a signal into a localized coefficient field and a global basis field with periodic transformations.
We extensively evaluate our DiF model on three signal reconstruction tasks including 2D image regression, 3D SDF reconstruction, and radiance field reconstruction. 
We demonstrate that our DiF model leads to state-of-the-art reconstruction quality, better or on par with previous methods on all three tasks, while achieving faster reconstruction and more compact model sizes than most methods.
Our DiF model is able to generalize across scenes by learning shared basis field factors from multiple signals, allowing us to reconstruct new signals from sparse observations. 
We show that, using such pre-trained basis factors, our method enables high-quality few-shot radiance field reconstruction from only 3 or 5 views,  outperforming previous methods like PixelNeRF and MVSNeRF in the sparse view / wide baseline setting.
In general, our framework takes a step towards a generic neural representation with high accuracy and efficiency.
We believe that the flexibility of our framework will help to inspire future research on efficient signal representations, exploring the potential of multi-factor representations or novel coordinate transformations.








\begin{table*}[t]
\begin{center}
    
\begin{small}
\setlength{\tabcolsep}{1.0pt}
\begin{subtable}{\textwidth}\centering

\vspace{-2mm}

\begin{tabular}{llllllllll}

& & \multicolumn{3}{c}{\textbf{Design}} & \multicolumn{5}{c}{\textbf{Performance}} \\
\cmidrule(lr){3-5} \cmidrule(lr){6-10} 
& Name & N & $\bff_i(\bx)$ & $\bgamma_i(\bx)$ & Time & Size & $2$D Images & Radiance Field & Few-shot RF \\
\hline
    (1) & \textbf{iNGP$^*$} & $1$ & Vectors & Hashing($\bx$)  & 00:45/\textbf{12:02}/\emptyMed & 0.92/6.80/\emptySmall & 34.73/0.906 \bronze  & 32.56/0.958 \bronze  & \emptySmall/\emptySmall\\
    (2) & \textbf{DiF-Hash-B} & $2$ & Vectors; $3$D grids & Hashing($\bx$)   & 00:55/13:10/\textbf{04:45} & 1.09/\textbf{4.37}/\textbf{3.28} & 37.53/0.949 \silve & 32.80/0.960 \silve & 26.62/0.919 \silve\\
    (3) & \textbf{EG3D$^*$} & 1 & $2$D Maps & Orthog$_{2D}$($\bx$) & \emptyMed/14:17/\emptyMed    & \emptySmall/4.54/\emptySmall & \emptySmall/\emptySmall & 30.01/0.935 &  \emptySmall/\emptySmall\\
    (4) & \textbf{TensoRF-VM$^*$}  & $2$ & $2$D Maps; Vectors & Orthog$_{1,2D}$($\bx$) & \emptyMed/16:20/13:06    & \emptySmall/4.55/\textbf{4.93} & \emptySmall/\emptySmall & 30.47/0.940 & 26.79/0.908 \gold \\
    (5) & \textbf{DiF-no-C} & $1$ & $3$D Grids &   Sawtooth($\bx$)   & \textbf{00:41}/12:55/\emptyMed & \textbf{0.82}/4.55/\emptySmall  & 22.28/0.479 & 31.10/0.947 & \emptySmall/\emptySmall\\
    (6) & \textbf{DiF-Grid} & $2$ & 3D Grids; 3D Grids & Sawtooth($\bx$)   & 01:13/12:10/11:35   & 0.87/5.10/7.32 & 39.51/0.963 \gold & 33.14/0.961 \gold & 25.41/0.915 \bronze \\

\end{tabular}
\caption{\textbf{Design Study on Number of Factors $N$. }}
\label{tab:factor_number}  
\end{subtable}

\vspace{-1.5mm}

\setlength{\tabcolsep}{0.5pt}
\begin{subtable}{\textwidth}\centering

\begin{tabular}{lllllllllll}
& & \multicolumn{3}{c}{\textbf{Design}} & \multicolumn{5}{c}{\textbf{Performance}} \\
\cmidrule(lr){3-5} \cmidrule(lr){6-10} 
& Name & N & $\bff_i(\bx)$ & $\bgamma_i(\bx)$ & Time & Size & $2$D Images & Radiance Field & Few-shot RF \\
\hline
    (1) & \textbf{OccNet$^*$} & $1$ & $\bx$  &   $\bx$   & 02:17/100:23/\emptySmall & \textbf{0.38}/\textbf{0.43}/\emptySmall & 13.90/0.437 & 20.60/0.849 &  \emptySmall/\emptySmall\\
    (2) & \textbf{NeRF$^*$} & $1$ & $\bx$ & Sinusoidal($\bx$) & 02:18/65:50/\emptyMed & 0.39/0.44/\emptySmall & 28.99/0.816 & 27.81/0.919 &  \emptySmall/\emptySmall\\
    (3) & \textbf{DiF-Hash-B} & $2$ & Vectors; $3$D Grids & Hashing($\bx$)   & 00:55/13:10/4:45 & 1.09/4.37/\textbf{3.28} & 37.53/0.949 \bronze & 32.80/0.960 \silve & 26.62/0.919 \gold\\
    (4) & \textbf{DiF-Hash-B-SL} & $2$ & Vectors; $3$D Grids & Hashing($\bx$) &  \textbf{00:41}/13:21/\textbf{03:03} & 0.89/4.54/4.59 & 30.97/0.891 & 31.11/0.941 \bronze & 24.13/0.881 \bronze\\
    (5) & \textbf{DiF-Grid} & $2$ & 3D Grids; 3D Grids & Sawtooth($\bx$)   & 01:13/\textbf{12:10}/11:35   & 0.99/5.10/7.32 & 39.51/0.963 \gold & 33.14/0.961 \gold & 25.41/0.915 \silve\\
    (6) & \textbf{DiF-Grid-SL} & $2$ & 3D Grids; 3D Grids & Sawtooth($\bx$)    & 00:49/22:35/13:12 & 0.98/5.11/7.25 & 38.73/0.973 \silve & 31.08/0.942 & 23.88/0.882\\
    
\end{tabular}
\caption{\textbf{Design Study on Pyramid Levels $L$.}}
\label{tab:level_number}  
\end{subtable}

\vspace{-1.5mm}
\setlength{\tabcolsep}{1.25pt}
\begin{subtable}{\textwidth}\centering
\begin{tabular}{lllllllllll}
& & \multicolumn{3}{c}{\textbf{Design}} & \multicolumn{5}{c}{\textbf{Performance}} \\
\cmidrule(lr){3-5} \cmidrule(lr){6-10} 
& Name & N & $\bff_i(\bx)$ & $\bgamma_i(\bx)$ & Time & Size & $2$D Images & Radiance Field & Few-shot RF \\
\hline
    (1) & \textbf{DiF-Hash-B} & $2$ & Vectors; $3$D Grids & Hashing($\bx$)   & \textbf{00:55}/13:10/4:45 & 1.09/\textbf{4.37}/\textbf{3.28} & 37.53/0.949 & 32.80/0.960  & 26.62/0.919 \gold\\
    (2) & \textbf{DiF-Sin-B} & $2$ & 3D Grids; 3D Grids& Sinusoidal($\bx$)    & \textbf{00:55}/12:19/\textbf{08:28} & \textbf{0.99}/5.10/7.32  & 38.21/0.953 \bronze    & 32.85/0.961 \bronze & 25.43/0.908 \silve\\
    (3) & \textbf{DiF-Tri-B} & 2 & 3D Grids; 3D Grids& Triangular($\bx$)    & \textbf{00:55}/12:49/08:44  & \textbf{0.99}/5.10/7.32  & 39.38/0.962 \silve   & 32.95/0.960 \silve & 24.78/0.904 \\
    (4) & \textbf{DiF-Grid} & $2$ & 3D Grids; 3D Grids & Sawtooth($\bx$)   & 01:13/\textbf{12:10}/11:35   & \textbf{0.99}/5.10/7.32 & 39.51/0.963 \gold & 33.14/0.961 \gold & 25.41/0.915 \bronze\\
  
\end{tabular}
\caption{\textbf{Design Study on Coordinate Transformations $\bgamma_i$.}}
\label{tab:coordiante_transformation}  
\end{subtable}

\vspace{-1.5mm}
\setlength{\tabcolsep}{1.03pt}
\begin{subtable}{\textwidth}\centering
\centering
\begin{tabular}{lllllllllll}
& & \multicolumn{3}{c}{\textbf{Design}} & \multicolumn{5}{c}{\textbf{Performance}} \\
\cmidrule(lr){3-5} \cmidrule(lr){6-10} 
& Name & N & $\bff_i(\bx)$ & $\bgamma_i(\bx)$ & Time & Size & $2$D Images & Radiance Field & Few-shot RF \\
\hline

    (1) & \textbf{TensoRF-VM$^*$}  & $2$ & $2$D Maps; Vectors & Orthog$_{1,2D}$($\bx$) & \emptyMed/16:20/13:06    & \emptySmall/4.55/4.93 & \emptySmall/\emptySmall & 30.47/0.940 & 26.79/0.908 \silve \\
    (2) & \textbf{DiF-Grid} & $2$ & 3D Grids; 3D Grids & Sawtooth($\bx$)   & 01:13/\textbf{12:10}/11:35   & 0.99/5.10/7.32 & 39.51/0.963 \gold & 33.14/0.961 \gold & 25.41/0.915\\
    (3) & \textbf{DiF-DCT} & $2$ & 3D Grids; 3D Grids & Sawtooth($\bx$) & 00:53/\emptyMed/\emptyMed & \textbf{0.18}/\emptySmall/\emptySmall  & 23.16/0.606 & \emptySmall/\emptySmall &  \emptySmall/\emptySmall \\
    (4) & \textbf{DiF-Hash-B} & $2$ & Vectors; $3$D Grids & Hashing($\bx$)   & 00:55/13:10/10.15 & 1.09/4.37/3.28 & 37.53/0.949 \silve & 32.80/0.960 \silve & 26.53/0.924 \bronze \\
    (5) & \textbf{DiF-MLP-B} & $2$ & MLP; 3D Grids & Sawtooth($\bx$) &  01:24/18:18/18:23 & \textbf{0.18}/0.62/2.53 & 28.76/0.819 & 29.62/0.932 & 26.91/0.927 \gold \\
    (6) & \textbf{DiF-MLP-C} & $2$ & 3D Grid; MLP  & Sawtooth($\bx$) & 01:13/13:38/\textbf{08:23} & 0.87/4.54/4.86 & 34.72/0.910 \bronze & 32.57/0.956 \bronze & 23.54/0.875\\
    (7) &\textbf{TensoRF-CP$^*$} & $3$ & Vectors$\times$3 & Orthog$_{1D}$($\bx$)  & \textbf{00:43}/28:05/12:42  & 0.39/\textbf{0.29}/\textbf{0.29} & 33.79/0.899 & 31.14/0.944 & 22.62/0.867\\
    
\end{tabular}
\caption{\textbf{Design Study on Field Representations $\bff_i$.}}
\label{tab:field_representation}  
\end{subtable}

\vspace{-1.5mm}
\setlength{\tabcolsep}{0.45pt}
\begin{subtable}{\textwidth}\centering
\begin{tabular}{lllllllllll}
& & \multicolumn{3}{c}{\textbf{Design}} & \multicolumn{5}{c}{\textbf{Performance}} \\
\cmidrule(lr){3-5} \cmidrule(lr){6-10} 
& Name & N & $\bff_i(\bx)$ & $\bgamma_i(\bx)$ & Time & Size & $2$D Images & Radiance Field & Few-shot RF \\
\hline
    (1) &\textbf{TensoRF-CP$^*$} & $3$ & Vectors$\times$3 & Orthog$_{1D}$($\bx$)  & \textbf{00:43}/\textbf{28:05}/10:11  & 0.39/0.29/0.29 & 33.79/0.899 \gold & 31.14/0.944 \gold & 23.19/0.879 \gold\\
    (2) &\textbf{TensoRF-CP$^*$-Cat} & $3$ & vectors$\times$3 & Orthog$_{1D}$($\bx$)  & 00:47/39:05/\textbf{09:47}  & 0.40/0.31/0.31 & 25.67/0.683 \silve & 26.75/0.905 \silve & 21.43/0.856 \silve\\
    \hline
    (3) & \textbf{TensoRF-VM$^*$}  & $2$ & $2$D Maps; vectors & Orthog$_{1,2D}$($\bx$) & \emptyMed/\textbf{16:20}/12:52    & \emptySmall/4.55/4.93 & \emptySmall/\emptySmall & 30.47/0.940 \gold & 26.99/0.911 \gold \\
    (4) & \textbf{TensoRF-VM$^*$-Cat}  & $2$ & $2$D Maps; vectors & Orthog$_{1,2D}$($\bx$) & \emptyMed/18:35/\textbf{07:02}    & \emptySmall/4.56/4.94 & \emptySmall/\emptySmall & 29.86/0.939 \silve & 24.67/0.885 \silve \\
    \hline
    (5) & \textbf{DiF-Grid} & $2$ & 3D Grids; 3D Grids & Sawtooth($\bx$)   & 01:13/12:10/11:35   & 0.99/5.10/7.32 & 39.51/0.963 \gold & 33.14/0.961 \gold & 25.41/0.915 \gold\\
    (6) & \textbf{DiF-Grid-Cat} & $2$ & 3D Grids; 3D Grids & Sawtooth($\bx$)   & \textbf{00:51}/\textbf{11:35}/\textbf{06:47}   & 1.00/5.10/7.32 & 37.76/0.946 \silve & 32.95/0.960 \silve & 24.71/0.894 \silve\\
  
\end{tabular}
\caption{\textbf{Performance comparison on element-wise product $\circ$ vs. concatenation.}}
\label{tab:concatenate}  
\end{subtable}

\end{small}
\end{center}
\caption{\textbf{Influence of Design Choices in the Factor Fields Framework.} The comparison is done using the same code base and hyperparameter configuration including number of levels, frequency of each level, \etc. Prior methods represented in our framework are labeled with $^*$ due to minor differences with respect to the original publications. Runtime and model size are reported separately for 2D Images / Radiance Fields / Few-shot Radiance Fields.\label{tab:ablation}}
\end{table*}

{\small
\bibliographystyle{ieee_fullname}
\bibliography{bibliography_short,bibliography,bibliography_custom}
}

\end{document}